\documentclass[pdflatex,sn-vancouver-num]{sn-jnl}

\usepackage{cmap}                %
\usepackage{lmodern}             %
\usepackage[T1]{fontenc}         %
\ifdefined\pdfinterwordspaceon\pdfinterwordspaceon\fi

\usepackage{amsmath,amssymb}     %
\usepackage{xcolor}              %
\usepackage{graphicx}
\usepackage{booktabs}            %
\usepackage{tikz}
\usetikzlibrary{positioning, arrows.meta, calc, fit, backgrounds}
\let\orcidlogo\undefined
\usepackage{orcidlink}           %

\hypersetup{bookmarksdepth=2}

\begin{document}

\title{ITU-DESC: a missingness-aware machine-learning dementia risk score in the UK Biobank cohort
}

\author*[1]{\fnm{Abdullah} \sur{Kavakl\i{}}\orcidlink{0009-0008-5959-249X}}\email{kavakli23@itu.edu.tr}
\author[1]{\fnm{Faik Boray} \sur{Tek}\orcidlink{0000-0002-8649-6013}}

\affil*[1]{\orgdiv{Department of Artificial Intelligence and Data Engineering}, \orgname{Istanbul Technical University}, \orgaddress{\city{Istanbul}, \country{Turkey}}}

\abstract{    
    \paragraph*{Background} Existing dementia risk scores rely on fixed, pre-selected features and linear models, or ignore bias from informative missingness in cohort data.
    Here we introduce the Istanbul Technical University Dementia Risk Score (ITU-DESC), an explainable machine-learning score for incident all-cause dementia that combines broad feature selection across roughly 8,000 candidate features with explicit control of missingness bias.

    \paragraph*{Results} We studied approximately 500,000 UK Biobank participants with a median follow-up of 16.6 years.
    We withheld missingness patterns from the model using unconditional random hot-deck imputation. We stabilized feature selection by repeating BorutaSHAP-based feature selection with ten random-imputation seeds and selecting predictors chosen in at least eight of the ten seeds, which yielded 38 predictors. We trained a CatBoost gradient-boosted decision-tree model and evaluated it on a holdout test set of 50,041 participants. For all-cause incident dementia, ITU-DESC achieved an average precision of 0.138 (95\% confidence interval 0.123--0.156) and an area under the receiver operating characteristic curve (ROC AUC) of 0.853 (0.844--0.863), significantly outperforming other risk scores, including the strongest comparator, the UK Biobank Dementia Risk Prediction model (0.118 and 0.841). We identified dementia risk predictors absent from established scores, including leukocyte telomere length and liver enzymes.

    \paragraph*{Conclusions} ITU-DESC is the first dementia risk score to combine broad feature selection with explicit control of missingness bias.}%


\maketitle

\section{Background}
Dementia affects approximately 57 million people worldwide, with nearly 10 million new cases each year~\cite{who_dementia_2025}. Survival after diagnosis is short: average life expectancy ranged from 8.0 years for women diagnosed at age 65 to 2.2 years for men diagnosed at age 85~\cite{bmj_dementia_prognosis_2025}. The 2024 Lancet Commission estimated that almost half of dementia cases could theoretically be prevented or delayed by addressing 14 modifiable risk factors across the life course. These factors span education in early life; sensory, vascular, metabolic, mental-health and behavioral risks in midlife; and social, environmental and sensory risks in later life, including social isolation, air pollution and vision loss~\cite{lancet2024}. Because prevention works best when it starts early and reaches those who need it most~\cite{lancet2024}, it depends on identifying individuals at high risk before dementia develops.

In recent years, several statistical risk scores have been developed to estimate future dementia risk. Most are based on linear modeling frameworks, including Cox proportional hazards, logistic regression or competing-risk regression, and use predefined sets of clinical and demographic predictors. Established examples include the Australian National University Alzheimer's Disease Risk Index (ANU-ADRI)~\cite{ANU_ADRI}; the Cardiovascular Risk Factors, Aging, and Incidence of Dementia (CAIDE) Risk Score~\cite{CAIDE}; the Dementia Risk Score (DRS)~\cite{DRS}; and the UK Biobank Dementia Risk Score (UKBDRS)~\cite{Anaturk2023}. These scores share predictors such as age, hypertension, stroke, parental history of dementia and high cholesterol, but differ in their target populations, input variables and model assumptions. Although these scores are explainable and practical, their fixed feature sets and linear assumptions limit their ability to capture non-linear associations and interactions between predictors. In the Norwegian HUNT cohort, several such scores did not outperform age and education alone in predicting dementia or cognitive decline~\cite{stubs2025hunt,stubs2026hunt}.

Machine-learning approaches relax these constraints and have also been applied to dementia risk. You et al.~\cite{YOU2022101665} developed the UK Biobank Dementia Risk Prediction model (UKB-DRP) using LightGBM, a gradient-boosting framework based on decision trees~\cite{lightgbm}. They were among the first to derive a dementia risk score from UK Biobank data with machine learning. The model identified several important predictors, including age, \textit{APOE} $\varepsilon$4 status, cognitive-test performance, number of medications, reaction time, peak expiratory flow, mother's age at death, long-standing illness and mean corpuscular volume. In large cohort datasets, however, machine learning introduces a problem: prediction may depend not only on recorded values but also on whether a measurement is available.

Missing data in UK Biobank are more than a preprocessing problem because the probability that a value is missing depends on participant characteristics, assessment procedures or health status. In the UKBDRS development sample, for example, participants excluded under complete-case analysis differed in age, sex and education from those retained~\cite{Anaturk2023,CODE_anaturk_include_vs_exclude_2023}. Because these characteristics are themselves dementia risk factors, missingness is related to the outcome and therefore informative for prediction~\cite{josse2024consistency}, whether or not it satisfies the assumption of missing at random~\cite{little2019statistical,structured_missingness}. Machine-learning models can therefore discriminate partly from the presence or absence of measurements rather than from the measured biological or clinical information. For example, if a blood test is unavailable more often among participants who later develop dementia, the model can learn this recording pattern as a proxy for future dementia risk. Such patterns may be specific to the cohort and its assessment procedures. In the present cohort, missingness in the predictors of the final score was itself associated with incident dementia (Section~\ref{subsec:missingness-results}).

Relying on such patterns introduces leakage-like bias, whereby predictive performance partly reflects how data were collected rather than associations expected to persist in the target population~\cite{leakage}. No approach to handling missing data is neutral in this setting. Algorithms with native missing-value handling can exploit informative missingness for prediction. Constant imputation using the mean, median or mode attains prediction-consistency under a missing-at-random mechanism. It does so because a sufficiently flexible learner detects the imputed value and can therefore treat missingness as a predictor~\cite{josse2024consistency}. Under informative missingness, the predictor it recovers relies on the recording pattern, the behavior we set out to avoid.

No existing dementia risk score combines broad machine-learning-based feature selection across UK Biobank variables with explicit control of this bias. We therefore developed the Istanbul Technical University Dementia Risk Score (ITU-DESC), an explainable machine-learning score for incident dementia. We treated missing-value handling as a design choice rather than a preprocessing step (Section~\ref{subsec:nan-value-data-leakage}). Starting from a large set of candidate baseline variables, we let the algorithm identify predictors under a missingness-handling procedure designed to limit the use of measurement availability as predictive information. We repeated feature selection across random-imputation seeds and kept only consistently selected predictors, making the final set robust to imputation variability.

In approximately 500{,}000 UK Biobank participants with variable follow-up, we developed and evaluated ITU-DESC for all-cause dementia and Alzheimer's disease. We assessed prediction across 5-year, 10-year and all-incident horizons, and compared ITU-DESC with ANU-ADRI~\cite{ANU_ADRI}, CAIDE~\cite{CAIDE}, DRS~\cite{DRS}, UKBDRS~\cite{Anaturk2023} and UKB-DRP~\cite{YOU2022101665}. Because these scores come from different cohorts or data partitions, with different outcome definitions and missing-data procedures, their published performance estimates are not directly comparable. We therefore applied to every comparator the same missing-data policy, outcome definitions and isotonic calibration procedure used for ITU-DESC. We evaluated all models on the same holdout test set, which no stage of feature selection or model fitting had seen.

\section{Methods}
\subsection{Dataset}
\label{subsec:dataset}
\begin{figure}[htb]
    \centering
\begin{tikzpicture}[
    font=\fontfamily{phv}\fontsize{7}{8.4}\selectfont,
    node distance=1.25cm and 1.1cm,
    box/.style={
        draw,
        rectangle,
        rounded corners=2pt,
        align=center,
        text width=4.5cm,
        minimum height=0.9cm,
        inner sep=5pt
    },
    smallbox/.style={
        draw,
        rectangle,
        rounded corners=2pt,
        align=left,
        text width=2.75cm,
        minimum height=1.45cm,
        inner sep=6pt
    },
    exclude/.style={
        draw,
        rectangle,
        rounded corners=2pt,
        align=left,
        text width=4.5cm,
        minimum height=1.2cm,
        inner sep=5pt,
        font=\fontfamily{phv}\fontsize{7}{8.4}\selectfont
    },
    arrow/.style={
        -{Stealth[length=2mm]},
        line width=0.35pt
    }
]

\node[box] (start) {UK Biobank participants \\ (n = 501,936)};
\node[box, below=1.9cm of start] (eligible) {Participants eligible\\for analysis \\ (n = 500,410)};

\coordinate (midpoint) at ($(start.south)!0.5!(eligible.north)$);
\node[exclude, right=0.9cm of midpoint] (exclude) {
    Excluded participants with:\\
    dementia at baseline (n = 230)\\
    no follow-up (n = 1,296)
};

\node[box, below=1.55cm of eligible, xshift=-3.15cm] (dev)
{Development set \\ (n = 450,369)};

\node[box, below=1.55cm of eligible, xshift=3.15cm] (test)
{Holdout test set \\ (n = 50,041)};

\node[box, below=1.55cm of dev, xshift=0.6cm] (control)
{No incident dementia\\[0.2em]
\begin{tabular}{@{}r@{\hspace{1.2em}}r@{}}
Dev. & Test \\
440,127 & 48,903 \\
\end{tabular}};

\node[box, below=1.55cm of test, xshift=-0.6cm] (incident)
{Incident outcomes\\during follow-up};

\node[smallbox, below=1.35cm of incident, xshift=-1.85cm] (ad) {
\textbf{Alzheimer's disease}\\[0.2em]
\begin{tabular}{@{}l@{\hspace{0.5em}}r@{\hspace{0.6em}}r@{}}
 & Dev. & Test \\
5-year & 231 & 26 \\
10-year & 1,506 & 167 \\
Overall & 4,609 & 513 \\
\end{tabular}
};

\node[smallbox, below=1.35cm of incident, xshift=1.85cm] (dem) {
\textbf{Dementia}\\[0.2em]
\begin{tabular}{@{}l@{\hspace{0.5em}}r@{\hspace{0.6em}}r@{}}
 & Dev. & Test \\
5-year & 681 & 75 \\
10-year & 3,802 & 421 \\
Overall & 10,242 & 1,138 \\
\end{tabular}
};

\draw[arrow] (start.south) -- (eligible.north);
\draw[arrow] (midpoint) -- (exclude.west);

\coordinate (split1) at ($(eligible.south) + (0,-0.55cm)$);
\draw (eligible.south) -- (split1);
\draw[arrow] (split1) -| (dev.north);
\draw[arrow] (split1) -| (test.north);

\coordinate (merge) at ($(dev.south)!0.5!(test.south) + (0,-0.55cm)$);
\draw (dev.south) |- (merge);
\draw (test.south) |- (merge);
\draw[arrow] (merge) -| (control.north);
\draw[arrow] (merge) -| (incident.north);

\coordinate (split2) at ($(incident.south) + (0,-0.45cm)$);
\draw (incident.south) -- (split2);
\draw[arrow] (split2) -| (ad.north);
\draw[arrow] (split2) -| (dem.north);

\end{tikzpicture}
    \caption{Participant selection for the present study from the UK Biobank cohort. The upper boxes show the number of participants remaining after exclusions and the split of the eligible cohort into a 90\% development set and a 10\% holdout test set. The lower boxes report counts separately for the development (Dev.) and holdout test sides of the split. For each side, they give the number of participants without incident dementia during follow-up. They also report the numbers of incident dementia and Alzheimer's disease cases within the 5-year, 10-year and full follow-up periods.}
    \label{fig:ukbiobank-selection-process}
\end{figure}
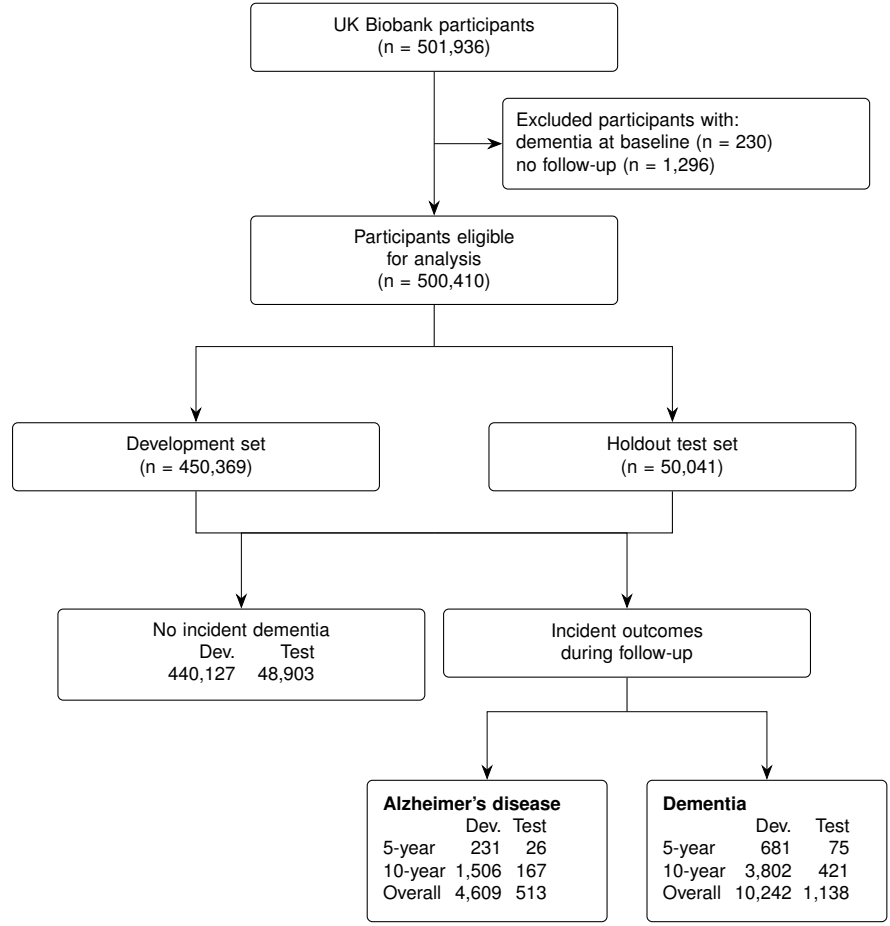

UK Biobank is a population-based longitudinal cohort study comprising more than 500{,}000 participants aged 40--69 years at recruitment~\cite{sudlow2015ukbiobank}. Baseline assessments were conducted between 2006 and 2010 at 22 assessment centers across the United Kingdom, with ongoing follow-up and repeat assessments in a subset of participants. Figure~\ref{fig:ukbiobank-selection-process} shows the participant selection process for the present study. Of 501{,}936 participants, we excluded 230 with prevalent dementia at baseline and 1{,}296 with no follow-up data, leaving 500{,}410 participants eligible for analysis. The median follow-up was 16.6 years (interquartile range, 15.8--17.4 years). We then divided the eligible cohort into a 90\% development set of 450{,}369 participants and a 10\% holdout test set of 50{,}041 participants (Section~\ref{subsec:split}).

\subsection{Outcome definition}
\label{subsec:outcome}
Dementia includes several clinical subtypes, of which Alzheimer's disease (AD) is the most common. Other subtypes include vascular dementia, frontotemporal dementia and dementia with Lewy bodies. To assess prediction across time horizons, we evaluated 5-year, 10-year and all-incident outcomes for both all-cause dementia and AD. Outcomes were defined using algorithmically derived UK Biobank variables based on self-reported diagnoses and ICD-10 codes~\cite{ukb_alg_outcome_main_2022}: Field 42018 for all-cause dementia and Field 42020 for AD. Incident outcomes were defined relative to each participant's baseline assessment date, corresponding to the first assessment visit (Field 53). Participants with dementia onset before baseline were classified as having prevalent dementia and excluded from the analysis.

\subsection{Development and holdout test split}
\label{subsec:split}
We reserved a holdout test set before feature selection, model development or any other modeling decisions, ensuring that test-set information did not influence model construction or performance estimation. Using a single stratified split, we assigned 10\% of the eligible cohort (50{,}041 participants) to the holdout test set and the remaining 90\% (450{,}369 participants) to the development set. The split was stratified on the six binary outcome indicators and sex. Given the large cohort size, we did not search over many candidate splits to optimize covariate balance; instead, we accepted the first candidate split that met prespecified acceptance criteria on outcome balance and on age, assessment center and ethnicity, covariates not included in the stratification scheme. The first candidate seed met all criteria. The stratification scheme, acceptance criteria and full split diagnostics are reported in Additional file 1: Supplementary Methods (Tables~S1 and~S2).

All subsequent preprocessing decisions, imputation donor pools, feature selection, model fitting and calibration were based exclusively on the development set. The holdout test set was used only for final model evaluation. Within the development set, we also used five-fold cross-validation for two purposes: screening candidate features (Section~\ref{subsec:feature-selection}) and generating out-of-fold predictions for fitting the calibration mappings (Section~\ref{sec:cross-validation}). For both applications, imputation donor pools were constructed separately within each training fold (Section~\ref{subsec:nan-value-data-leakage}).

\subsection{Methodological challenges in dementia risk prediction}
\label{subsec:challenges}
Dementia risk prediction presents several methodological challenges arising from the disease course, cohort structure and characteristics of the data.

First, dementia risk increases markedly with age, and some participants classified as non-cases during the observed follow-up may develop dementia later. This introduces label uncertainty, particularly for outcomes with long latency, because participants considered dementia-free at the censoring date may subsequently receive a diagnosis.

Second, participants with established dementia are difficult to characterize reliably at baseline. Cognitive impairment may reduce the accuracy or completeness of self-reported information and may also affect participation in detailed assessments. For risk prediction, data collected before clinical onset and linked to subsequent diagnosis are therefore more informative than data collected after disease onset. Accordingly, participants with prevalent dementia at baseline were excluded from the analysis.

Third, incident dementia was rare in the analyzed UK Biobank cohort. The positive class represented 2.27\% of participants, resulting in substantial class imbalance. This imbalance can reduce the stability of model estimation and make evaluation metrics sensitive to the small number of outcome-positive cases. The observed incidence of 2.27\% also represents a lower bound on eventual lifetime dementia risk because some participants classified as non-cases during follow-up may be diagnosed after the current censoring date.

Fourth, missing data are common in UK Biobank. Missing values may be imputed, handled natively by algorithms that support missing inputs or addressed through complete-case analysis~\cite{CCA}. Each approach has limitations, particularly when missingness is informative or associated with the outcome. Our approach to missing data and the associated risk of leakage-like bias are described in Section~\ref{subsec:nan-value-data-leakage}.

Fifth, UK Biobank includes repeated measurements for some variables across multiple assessment visits. All participants attended the baseline assessment, but only a subset returned for repeat visits, and not all variables were measured at each visit~\cite{visit_times}. Although repeated assessments can provide longitudinal information, their use would introduce inconsistencies in both follow-up time and feature availability across participants. Later visits were available for only a subset of participants, contained fewer variables, and could reflect changes in participant characteristics with age. To define a common prediction time point and feature set, we therefore used only data from the baseline assessment to predict incident dementia.

Finally, the dataset contained a large number of variables that required preprocessing and computationally demanding feature selection. Many variables had to be excluded to prevent data leakage, and subsequent feature selection was computationally demanding and required careful validation. After preprocessing, 7{,}977 candidate features remained, creating a prediction problem subject to the curse of dimensionality~\cite{koutroumbas2008pattern}. We therefore applied a two-stage feature-selection pipeline, as described in Section~\ref{subsec:feature-selection}.

\subsection{Feature preprocessing and leakage prevention}
\label{subsec:feature-preprocessing}
The extract contained 32{,}612 columns spanning healthcare records; questionnaire responses; physical measurements; and environmental, demographic, lifestyle and self-reported health data. Restricting the data to the baseline assessment and collapsing each field to a single typed column reduced this number to 11{,}926. Removing dementia- and outcome-linked fields (explained below) then left 11{,}340 columns. Multiple-choice fields, for which all selected responses were stored in a single delimited cell, were expanded into separate binary indicators for each observed response code, increasing the dataset to 37{,}693 columns. For participants with no recorded response to the original field, all corresponding indicators were set to missing rather than zero. Thus, zero indicates that an option was not selected, whereas missing indicates that the question was not answered. Applying the a priori exclusions described below reduced this set to 7{,}977 candidate features, which formed the input to all subsequent feature-selection procedures. Column counts at each stage are reported in Additional file 1: Table~S3, and the composition of the resulting candidate feature set in Additional file 1: Table~S5.

We excluded variables for three main reasons. First, we removed variables that could introduce leakage, including dementia subtype fields; certain fields whose values are recorded after the baseline assessment, namely Hospital Episode Statistics inpatient records (Category 2000), which may include dementia diagnoses, and COVID-19 fields; and fields recording the year of a diagnosis or measurement.

Second, we excluded variables that primarily reflected UK Biobank data-collection procedures rather than participant characteristics. These included assay and genotyping quality-control flags, bulk-data pointers and assessment-center timestamps, because such signals are unlikely to be present beyond the study setting.

Third, we excluded variables whose recorded values did not adequately represent the baseline characteristic of interest. Environmental exposure variables are assigned according to place of residence and may change during follow-up, making a single baseline measurement an unstable representation of longer-term exposure. Year-of-diagnosis fields for various health outcomes posed an additional problem beyond leakage: these values are missing for participants who never experienced the corresponding event. Because all retained variables were imputed (Section~\ref{subsec:nan-value-data-leakage}), retaining these fields would have assigned diagnosis years to participants who had never received those diagnoses.

All exclusions were specified at the field level before model fitting, and no exclusion decision was based on an observed association with the outcome or on model performance. In total, 3{,}452 UK Biobank entries were excluded. Most exclusions were due to outcome or temporal leakage (2{,}692 entries), followed by study-administration and timing (243) and COVID-19 (201). The remaining exclusions comprised employment-history variables, assay and genotyping quality-control fields, self-reported mental-health variables, environmental-exposure fields, bulk-data pointers, derived accelerometry measures, subjective health items and variables redundant with retained fields. Additional file 1: Table~S4 reports the group-level breakdown, and the main groups are described in detail under ``Column construction and a priori exclusions'' in Supplementary Methods. The individual excluded entries and the columns of the candidate feature set are listed in Additional file 2.

Beyond these exclusions, we combined parental survival status and longevity into one derived variable for each parent. Participants whose parent had died were assigned that parent's age at death, whereas participants whose parent was still alive were assigned a sentinel value below the observed age range. Combining these variables reflects the structure of the UK Biobank data, in which a missing age at death generally indicates that the parent was still alive rather than that the value was unrecorded~\cite{mothers_age_at_death_ukbiobank,fathers_age_at_death_ukbiobank}.

\subsection{Missing data and imputation}
\label{subsec:nan-value-data-leakage}
In UK Biobank, whether a measurement is available depends on participant characteristics, assessment procedures and health status.  Such missingness is structured rather than incidental~\cite{structured_missingness}. Mitra et al.\ describe a common clinical example: a test is ordered when disease is suspected, so whether a value is observed depends on the very status the test would reveal, and the resulting missingness pattern is itself unobservable~\cite{structured_missingness}. Under Rubin's taxonomy, such data may be missing not at random (MNAR)~\cite{rubin1976inference,little2019statistical}. Even in the absence of a causal relationship, missingness can carry predictive information about the outcome, and a supervised learner will exploit that association as a shortcut.

Tree-based learners can take that shortcut directly. Rather than requiring complete inputs, they route missing values inside the splitting procedure itself. One general strategy, missing incorporated in attribute (MIA), treats missingness as a category; eXtreme Gradient Boosting (XGBoost)~\cite{xgb} and scikit-learn's histogram gradient boosting implement this approach~\cite{josse2024consistency}, while LightGBM and CatBoost provide comparable native handling of missing values. Josse et al.\ recommend MIA because it can accommodate both informative and non-informative missingness~\cite{josse2024consistency}. We nevertheless did not use native missing-value handling in the primary analysis because our aim was to identify predictors from their recorded values rather than from cohort-specific patterns of measurement availability. Although native handling can increase observed discrimination, gains driven by such cohort-specific patterns may not generalize to clinical settings, where measurements are obtained under different procedures.

Deterministic single-value imputation raised a related concern. Mean or median imputation concentrates additional mass at a single value of a continuous variable, whereas mode imputation inflates one category; such deterministic approaches therefore alter the distribution of the imputed variable~\cite{hasler2016balanced}. Constant imputation can nevertheless yield consistent predictions with a sufficiently flexible learner. This guarantee holds under a missing-at-random mechanism, and only when the imputed value is identifiable. Identifiability lets the learner detect that value and use missingness as a predictor. For the coded categorical variables in our candidate set, it requires encoding missingness as an explicit category~\cite{josse2024consistency}. That is the information we sought to withhold from the model. We rejected complete-case analysis for a different reason: excluding every participant with a missing value in any required predictor can substantially reduce the sample in a high-dimensional feature set and, depending on the missingness mechanism, produce a selected analysis population~\cite{josse2024consistency,CCA}. Multiple imputation by chained equations (MICE)~\cite{mice_imputation} and random-forest methods such as missForest~\cite{missforest} use information from the remaining variables and may therefore reconstruct missing values more accurately than draws from a variable's marginal distribution. We did not use these methods because they condition on multivariable patterns that may include the cohort-specific measurement structure we aimed to suppress, and because each imputation model would need to be refitted within every feature-selection fold.

We therefore used unconditional, or marginal, random hot-deck imputation~\cite{chaput2015joint}. For each variable, every missing entry was replaced by an independent draw, with replacement, from the values observed for that variable in the corresponding training partition. Because hot-deck imputation draws only from observed values, imputed entries remain within the observed support of the variable, while random donor selection preserves the marginal distribution of the donor pool in expectation~\cite{andridge2010hotdeck,kim2004fractional,hasler2016balanced}. Random donor sampling also meant that the imputed data varied across random seeds; Section~\ref{subsec:feature-selection} describes how this variability was handled. Donor pools consisted of all observed values for a variable in the relevant training partition and were not stratified by auxiliary variables. Donors were drawn exclusively from the development data. During five-fold cross-validation within the development set (Section~\ref{subsec:split}), a separate donor pool was constructed from each training fold and used to impute both that fold and its validation fold. For the final model, donor pools were constructed from the full development set. The holdout test set therefore never contributed to the imputation rule applied to it, consistent with the principle that preprocessing fitted during model development should be applied unchanged to unseen data~\cite{josse2024consistency}. UK Biobank also encodes responses such as ``Do not know'' and ``Prefer not to answer'' as ordinary field values, which a gradient-boosting model could otherwise treat as substantive responses. We therefore recoded these responses as missing before imputation (Additional file 1: Supplementary Methods, ``Missing-value handling'').

For benchmarking, we applied the same missing-data policy to all models. UKB-DRP was reimplemented using random imputation instead of LightGBM's native missing-value handling, and missing components of UKBDRS, CAIDE, DRS and ANU-ADRI were likewise randomly imputed rather than treated according to their original specifications. For UKBDRS, which was originally evaluated using complete-case analysis, this represented a departure from its published implementation. Draws were keyed by participant identifier rather than row position, so that each score received the same imputed value for a given participant and variable. The risk-score comparison therefore used identical imputed data rather than only a common imputation policy. The harmonized policy allowed all models to be evaluated in the same participants and prevented differences in missing-data handling from confounding the comparison. To quantify the effect of withholding native missing-value handling, we additionally evaluated UKB-DRP with LightGBM's native handling on the same holdout test set (Section~\ref{subsec:comparison-existing-risk-scores}).

\subsection{Feature filtering and selection}
\label{subsec:feature-selection}
\begin{figure}[htbp]
    \centering
\begin{tikzpicture}[
    font=\fontfamily{phv}\fontsize{7}{8.4}\selectfont,
    node distance=0.42cm and 1.0cm,
    process/.style={
        draw=black!70,
        rectangle,
        rounded corners=2pt,
        fill=black!3,
        align=center,
        text width=5.5cm,
        minimum height=0.82cm,
        inner sep=5pt
    },
    side/.style={
        draw=black!70,
        rectangle,
        rounded corners=2pt,
        fill=black!1,
        align=center,
        text width=3.7cm,
        minimum height=0.82cm,
        inner sep=5pt
    },
    decision/.style={
        draw=black!70,
        rectangle,
        rounded corners=2pt,
        fill=black!6,
        align=center,
        text width=4.8cm,
        minimum height=0.82cm,
        inner sep=5pt
    },
    arrow/.style={
        -{Stealth[length=2mm]},
        line width=0.35pt
    },
    heldout/.style={
        -{Stealth[length=2mm]},
        line width=0.35pt,
        dashed
    },
    seedloop/.style={
        draw=black!55,
        dashed,
        rounded corners=3pt,
        inner sep=8pt
    }
]

\node[process] (features) {Preprocessed feature matrix\\(\textasciitilde8k features)};
\node[decision, below=of features] (split) {Stratified 90/10 split};
\node[process, below=of split] (dev) {Development set\\(90\%, n = 450,369)};
\node[side, right=of dev] (test) {Holdout test set\\(10\%, n = 50,041)};

\node[process, below=0.7cm of dev] (shap) {Random imputation (seed \textit{s})\\+ 5-fold SHAP screen};
\node[process, below=of shap] (zero) {Remove features with zero SHAP importance\\across folds};
\node[process, below=of zero] (reduced) {Seed-specific reduced set\\(4{,}554--4{,}635 features)};
\node[process, below=of reduced] (boruta) {BorutaSHAP wrapper selection\\on that seed's reduced set};

\begin{scope}[on background layer]
    \node[seedloop, fit=(shap)(zero)(reduced)(boruta)] (loop) {};
\end{scope}
\node[anchor=south, rotate=90, font=\fontfamily{phv}\fontsize{7}{8.4}\itshape\selectfont, inner sep=3.5pt]
    at (loop.west) {Repeated in full for each seed \textit{s} = 1, \ldots, 10};

\node[process, below=0.7cm of boruta] (perseed) {10 seed-specific feature sets\\(40--59 features each)};
\node[decision, below=of perseed] (cutoff) {Consensus cutoff:\\selected in at least 8 of 10 seeds};
\node[process, below=of cutoff] (selected) {ITU-DESC feature set\\(38 features)};
\node[process, below=of selected] (fit) {Fit on the development set};
\node[side, below=of fit] (evaluation) {Single evaluation on\\the holdout test set};

\draw[arrow] (features) -- (split);
\draw[arrow] (split) -- (dev);
\draw[arrow] (split.east) -| (test.north);
\draw[arrow] (dev) -- (shap);
\draw[arrow] (shap) -- (zero);
\draw[arrow] (zero) -- (reduced);
\draw[arrow] (reduced) -- (boruta);
\draw[arrow] (boruta) -- (perseed);
\draw[arrow] (perseed) -- (cutoff);
\draw[arrow] (cutoff) -- (selected);
\draw[arrow] (selected) -- (fit);
\draw[arrow] (fit) -- (evaluation);
\draw[heldout] (test.east) -- ++(1.0cm,0) |- (evaluation.east);

\end{tikzpicture}
    \caption{Feature-selection pipeline. A stratified 90/10 split holds 10\% of participants out as the test set. Within the 90\% development set, the boxed two-stage pipeline consists of a five-fold SHAP screen that removes features with zero importance, followed by BorutaSHAP wrapper selection. The pipeline is repeated in full for each of the ten random-imputation seeds. Both stages therefore run ten times, and each seed carries its own screened feature set into its own wrapper run. Features selected in at least 8 of the 10 seeds form the ITU-DESC feature set. The model is then fitted on the development set and evaluated once on the holdout test set, which the selection stages never see.}
    \label{fig:ukbiobank-flow}
\end{figure}
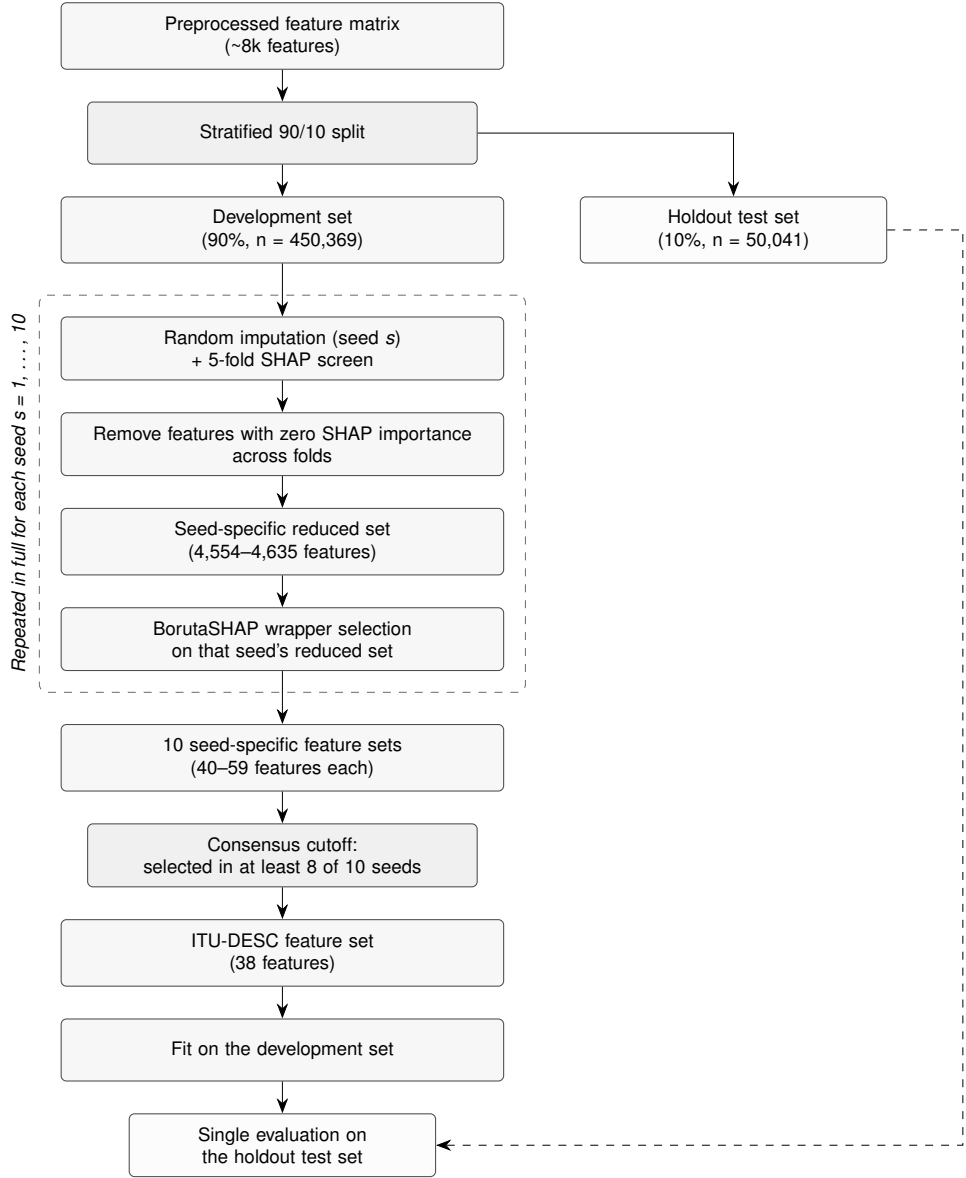

Feature selection started from the 7{,}977 candidate features (Section~\ref{subsec:feature-preprocessing}) and ran entirely within the development set (Fig.~\ref{fig:ukbiobank-flow}). To allow broad feature discovery, we considered all feasible baseline variables, including those without established associations with dementia. Selection proceeded in two stages: screening based on SHapley Additive exPlanations (SHAP)~\cite{SHAP}, followed by wrapper selection. SHAP values decompose an individual prediction into additive contributions from each feature, providing a signed attribution for every feature on the model output scale. Global feature importance was summarized as the mean absolute SHAP value across participants. Both stages used XGBoost as the underlying learner. Because random imputation made both stages stochastic, the complete pipeline was repeated for each of ten random-imputation seeds, as described below.

For initial screening, we computed SHAP values on each validation fold of the five-fold cross-validation within the development set (Section~\ref{subsec:split}), imputing each fold's training and validation partitions under the corresponding seed. SHAP values were then aggregated across the folds, and features with zero SHAP importance in every fold were removed. This step eliminated features with no observed contribution before the more computationally intensive wrapper-based selection stage. The ten seed-specific screened sets were similar in size, containing between 4{,}554 and 4{,}635 variables.

After reducing the feature set to a tractable size, we applied BorutaSHAP~\cite{borutashap}, a wrapper-based feature-selection algorithm extending the Boruta approach~\cite{boruta}. BorutaSHAP combines a tree-based learning model, here the same XGBoost learner, with SHAP-based feature importance to identify relevant predictors. Briefly, the algorithm augments the dataset with shuffled copies of the original variables, termed shadow features, and compares the importance of each real feature with a shadow-feature threshold. Features were retained when they exceeded this threshold significantly more often than expected by chance across BorutaSHAP iterations; features that failed this test were rejected.

Because random imputation makes both selection stages stochastic, a single pass reports the features favored by one imputation draw rather than the features the data support. Drawing values at random adds imputation variance to every downstream estimate~\cite{hasler2016balanced}, and feature selection inherits it. We therefore repeated the pipeline independently for ten random-imputation seeds, screen and wrapper alike, so that each wrapper run saw both a different imputation draw and the screened feature set produced under that draw. The ten runs selected between 40 and 59 features each, and their selections did not coincide. We then ranked features by how many seeds selected them and defined nested consensus sets: Set $k$ contained the features selected in at least $k$ of the ten seeds, so Set 10 was the intersection of all ten lists and Set 1 their union. We adopted Set 8, the 38 features selected in at least eight of ten seeds, as the fixed ITU-DESC specification. This cutoff keeps only predictors that survive most imputation draws, while avoiding the stricter intersection, which would discard nine features that eight or nine seeds agreed on.

\subsection{Model development and validation}
\label{sec:cross-validation}
We used gradient-boosted decision trees because they can capture non-linear associations and interactions and perform well in high-dimensional, imbalanced classification settings. XGBoost~\cite{xgb} was used throughout the feature-selection pipeline, whereas the final ITU-DESC model was trained with CatBoost~\cite{catboost}. We did not apply resampling, class weighting or other class-imbalance corrections.
All stages, from SHAP screening and BorutaSHAP to the final CatBoost fit, used the same training label: the binary all-incident, all-cause dementia outcome defined in Section~\ref{subsec:outcome}, assessed over each participant's full available follow-up rather than at a fixed time horizon. Software and package versions are reported under ``Software and implementation'' in Additional file 1: Supplementary Methods (Table~S6).

Within the development set, we cross-fitted the model across five stratified folds (Section~\ref{subsec:split}) so that each participant received an out-of-fold prediction from a model that had not been trained on that participant. The pooled out-of-fold predictions were used both to fit the isotonic calibration mappings subsequently applied to the holdout-test predictions and to generate internal development diagnostics. A separate calibration mapping was fitted for each outcome and prediction horizon, allowing the same base model to be recalibrated for all-cause dementia and Alzheimer's disease at the 5-year, 10-year and all-incident horizons. The base model itself was trained only once, using the all-cause dementia label. We then refitted the model on the full development set and evaluated it once on the holdout test set. Because the holdout test participants were not used in the split procedure, imputation donor pools, SHAP screening, BorutaSHAP selection or calibration fitting, the holdout-test results provide the primary performance estimates, whereas the development cross-fitting results are reported as internal diagnostics.

\subsection{Evaluation and statistical analysis}
We formulated prediction as a binary classification task. The aim was to compare discrimination between ITU-DESC and established risk scores under a common outcome definition, data split and set of performance metrics, rather than to estimate absolute time-to-event risk.

Because incident dementia was rare, average precision (AP) was chosen a priori as the primary discrimination metric. In imbalanced settings, precision--recall summaries are more informative than ROC-based summaries because they focus on performance in the outcome-positive class~\cite{prauc}. We used AP rather than the trapezoidal area under the precision--recall curve because AP summarizes precision at observed changes in recall without linear interpolation, thereby avoiding optimistic interpolation between points on stepwise precision--recall curves~\cite{scikitap}. We reported ROC AUC as a complementary metric. Both metrics were computed from isotonic-calibrated predicted risks using the calibration procedure described below, which was applied identically to all models. Because the isotonic mapping is monotonic, it preserves the ordering of predicted risks except where ties are introduced. Corresponding results based on uncalibrated predictions are reported in Additional file 1: Supplementary Note, ``Uncalibrated discrimination''.

Pairwise differences in AP were evaluated using two-sided paired permutation tests on holdout-test predictions (2{,}000 permutations). Outcome labels were held fixed, and model scores were randomly swapped within participants. Differences in ROC AUC were assessed using two-sided paired DeLong tests on matched holdout-test predictions. Uncertainty was quantified using two-sided 95\% confidence intervals (CIs), reported for each model's AP and ROC AUC and for each difference from ITU-DESC. Because average precision has no analytic variance estimator, AP CIs were obtained by paired subject-level bootstrap resampling of the holdout test set (10{,}000 resamples; percentile method). The same resampled participants were used for all models within each bootstrap replicate, so differences remained paired. ROC AUC CIs, for both individual models and pairwise differences, were derived from the DeLong variance estimator, ensuring that each interval and its corresponding $P$ value were based on the same variance estimate. These intervals quantify sampling variability within the holdout test set under the single development/test split, but do not capture variability arising from model training or from the split itself.

Calibration is particularly important when predictions are intended to support clinical decision-making, because systematically overestimated or underestimated risks may mislead clinical decisions~\cite{vancalster2019_calibration}. We therefore recalibrated each model's predictions post hoc using isotonic regression. Isotonic regression fits a monotonically non-decreasing, piecewise-constant mapping from raw scores to calibrated probabilities and, unlike Platt scaling, assumes only monotonicity~\cite{zadrozny2002_transforming_classifier_scores, niculescu2005_predicting_good_probabilities}. This flexibility increases the risk of overfitting when the calibration sample or the number of outcome events is limited~\cite{niculescu2005_predicting_good_probabilities}, although the large numbers of participants and outcome events in the present study reduced this risk. Because all scores were recalibrated on UK Biobank data, the benchmark reflects UK Biobank-specific recalibrations rather than performance on the models' original probability scales. Calibration was assessed for the all-incident outcome, our primary endpoint, whereas discrimination, which depends only on the ranking of predicted risks, was additionally evaluated at the 5-year and 10-year horizons.

As a sensitivity analysis, we re-evaluated all scores on the holdout test set with death without dementia treated as a competing event (Additional file 1: Supplementary Note, ``Competing risk of death''). Because the latest baseline visit occurred more than 15 years before the administrative censoring date and participants lost to follow-up had been excluded, no participant was censored before the 5-, 10- or 15-year horizons. The binary horizon labels therefore coincide with the observed cumulative-incidence status at these time points. Under this design, we compared time-dependent AUCs that counted participants who died as controls or excluded them, concordance indices with deaths censored or retained as non-events, cause-specific Cox and Fine--Gray hazard ratios and the Aalen--Johansen cumulative incidence of dementia across ITU-DESC risk deciles.

Beyond discrimination and calibration, we interpreted the fitted model using SHAP values, the same attribution method used for feature screening in Section~\ref{subsec:feature-selection}. These attributions were computed post hoc, after the feature set and final model had been fixed, and therefore describe how the fitted model uses its predictors rather than contributing to model construction. Unless otherwise stated, SHAP values were computed on the holdout test set using the model refitted on the full development set. Global feature importance was summarized as the mean absolute SHAP value across holdout-test participants, whereas individual predictions were examined using local SHAP explanations for representative cases. The cumulative-discrimination analysis was the only exception: predictors were ranked by mean absolute SHAP value computed on the development set, ensuring that their ordering was determined without reference to the holdout test data.

Baseline characteristics of the 38 predictors were compared between participants with incident dementia ($n = 11{,}380$) or incident Alzheimer's disease ($n = 5{,}122$) and those free of incident dementia ($n = 489{,}030$) using the standardized mean difference (SMD), computed on observed values, with an absolute SMD of at least 0.1 considered meaningful. Univariate associations of each predictor with incident dementia and with incident Alzheimer's disease were summarized as odds ratios with 95\% confidence intervals from logistic regression, with continuous predictors z-standardized so that each odds ratio corresponds to a one-SD increase (Additional file 1: Supplementary Note, ``Baseline characteristics and univariate associations'').

Finally, we tested post hoc whether missingness in the final predictors was itself associated with the outcome in this cohort (Section~\ref{subsec:nan-value-data-leakage}). For each of the 37 ITU-DESC predictors with any missing values, we estimated the association between its missingness and incident dementia in the full study cohort using logistic regression, unadjusted and adjusted for age, sex, education, deprivation, recruitment year, assessment center and ethnicity (Additional file 1: Supplementary Note, ``Missingness as a predictive signal'').

\subsection{Use of generative artificial intelligence}
Generative artificial intelligence (AI) tools were used to assist with code implementation and refinement, grammar correction, and improvements to manuscript readability and clarity. The authors reviewed all AI-assisted outputs, checked the code and resulting analyses, and revised the text. The authors approved the final manuscript and take full responsibility for the code, analyses, text and interpretation of the findings.

\section{Results}

\subsection{Baseline characteristics}
\label{subsec:baseline-characteristics}
Age separated the groups most sharply: participants who later developed dementia were about eight years older at baseline than those who remained dementia-free (median 65.9 versus 58.1 years; SMD 1.19; Additional file 1: Table~S7). Maternal survival came next, with 87.3\% of future dementia cases reporting that their mother had died, against 59.2\% of the remainder (SMD $-0.67$). \textit{APOE} $\varepsilon$4 carriage, the Alzheimer's disease polygenic risk score and the number of medications taken followed, each near an SMD of 0.56. Most other selected predictors showed modest-to-moderate univariate differences, suggesting that prediction depended on their combined multivariable contribution rather than on uniformly large standalone effects. An extended baseline table reporting every response category and the number of participants with a recorded value in each group is provided as Additional file 1: Table~S8, and univariate associations of each predictor with both outcomes as Table~S9.

\subsection{Comparison with existing risk scores}
\label{subsec:comparison-existing-risk-scores}
\begin{figure}[htbp]
    \centering
    \includegraphics{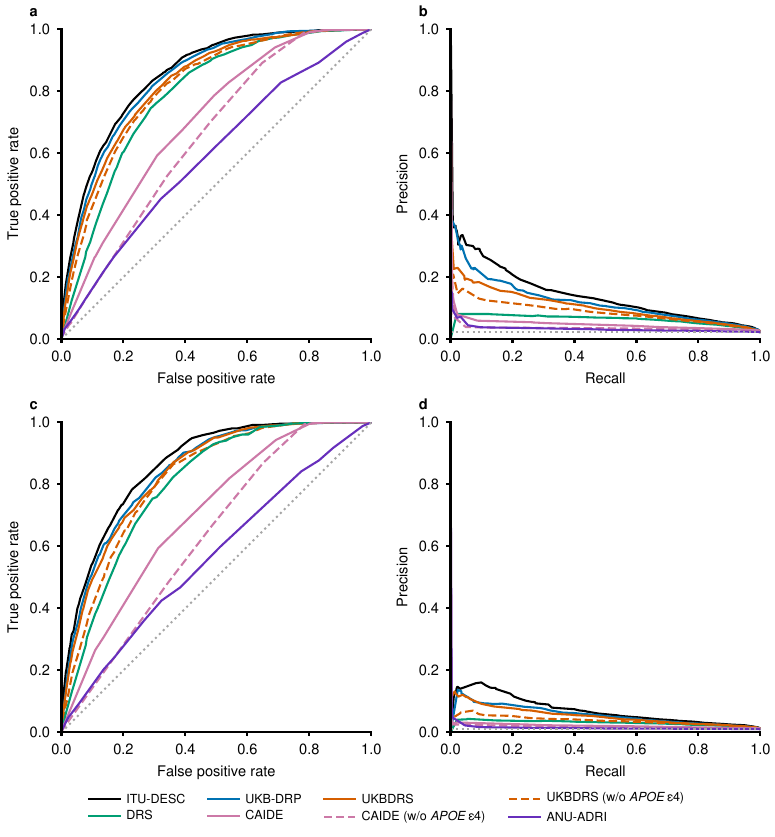}
    \caption{Comparison of ITU-DESC with established dementia risk scores. All models were evaluated on the holdout test set using calibrated predictions. \textbf{a}, Receiver operating characteristic curves for all-incident all-cause dementia ($1{,}138$ incident cases). \textbf{b}, Precision--recall curves for all-incident all-cause dementia. \textbf{c}, Receiver operating characteristic curves for all-incident Alzheimer's disease ($513$ incident cases). \textbf{d}, Precision--recall curves for all-incident Alzheimer's disease. The dotted line marks chance discrimination in \textbf{a} and \textbf{c}, and the outcome prevalence in \textbf{b} and \textbf{d}, which equals the average precision of a non-informative model. Discrimination estimates and pairwise comparisons are reported in Table~\ref{tab:risk_scores_table} for dementia and Additional file 1: Table~S12 for Alzheimer's disease.}
    \label{fig:benchmark_comparison}
\end{figure}

We compared ITU-DESC with UKB-DRP~\cite{YOU2022101665}, UKBDRS~\cite{Anaturk2023}, DRS~\cite{DRS}, CAIDE~\cite{CAIDE} and ANU-ADRI~\cite{ANU_ADRI} using isotonic-calibrated predictions from the holdout test set (Table~\ref{tab:risk_scores_table}). All risk scores were evaluated under the same random-imputation policy used for ITU-DESC, allowing every model to score the same 50{,}041 holdout-test participants without restricting any comparison to a complete-case subset. ITU-DESC showed the highest discrimination, with an AP of 0.138 (95\% CI, 0.123 to 0.156) and a ROC AUC of 0.853 (0.844 to 0.863), exceeding all comparators on both metrics. UKB-DRP ranked second, with an AP of 0.118 (0.105 to 0.133) and a ROC AUC of 0.841 (0.831 to 0.851). The differences between ITU-DESC and UKB-DRP were 0.021 (0.008 to 0.033) for AP and 0.013 (0.006 to 0.019) for ROC AUC, the smallest margins among the comparators; both confidence intervals excluded zero. Applying random imputation rather than each model's native missing-value handling also prevented models from gaining discrimination through informative missingness. Consistent with this effect, UKB-DRP achieved higher discrimination with native LightGBM missing-value handling than under random imputation (AP, 0.127 versus 0.118; ROC AUC, 0.847 versus 0.841), both estimated from isotonic-calibrated holdout-test predictions. Figure~\ref{fig:benchmark_comparison} shows the ROC and precision--recall curves for all-cause dementia in panels a and b, and for Alzheimer's disease in panels c and d. Additional file 1: Fig.~S1 and Tables~S10 and~S11 report the corresponding results using uncalibrated predictions.

\begin{table}[htbp]
\setlength{\tabcolsep}{5pt}
\caption{Discrimination of ITU-DESC and established dementia risk scores for all-incident dementia}
\label{tab:risk_scores_table}
\centering
\begin{tabular}{lccc}
\toprule
\textbf{Model} & \textbf{AP} (95\% CI) & \textbf{$\Delta$AP} (95\% CI) & \textbf{$P$} \\
\midrule
ITU-DESC (our model) & 0.138 (0.123, 0.156) & --- & --- \\
UKB-DRP~\cite{YOU2022101665} & 0.118 (0.105, 0.133) & 0.021 (0.008, 0.033) & $<0.01$ \\
UKBDRS~\cite{Anaturk2023} & 0.102 (0.092, 0.114) & 0.036 (0.023, 0.050) & $<0.001$ \\
UKBDRS (w/o \textit{APOE} $\varepsilon$4) & 0.084 (0.076, 0.093) & 0.055 (0.041, 0.070) & $<0.001$ \\
DRS~\cite{DRS} & 0.064 (0.058, 0.069) & 0.075 (0.061, 0.090) & $<0.001$ \\
CAIDE~\cite{CAIDE} & 0.044 (0.040, 0.048) & 0.094 (0.081, 0.110) & $<0.001$ \\
CAIDE (w/o \textit{APOE} $\varepsilon$4) & 0.035 (0.032, 0.038) & 0.104 (0.090, 0.120) & $<0.001$ \\
ANU-ADRI~\cite{ANU_ADRI} & 0.032 (0.029, 0.036) & 0.106 (0.092, 0.123) & $<0.001$ \\
\addlinespace
\midrule
\textbf{Model} & \textbf{ROC AUC} (95\% CI) & \textbf{$\Delta$ROC AUC} (95\% CI) & \textbf{$P$} \\
\midrule
ITU-DESC (our model) & 0.853 (0.844, 0.863) & --- & --- \\
UKB-DRP~\cite{YOU2022101665} & 0.841 (0.831, 0.851) & 0.013 (0.006, 0.019) & $<0.001$ \\
UKBDRS~\cite{Anaturk2023} & 0.825 (0.814, 0.836) & 0.028 (0.021, 0.035) & $<0.001$ \\
UKBDRS (w/o \textit{APOE} $\varepsilon$4) & 0.812 (0.801, 0.823) & 0.041 (0.033, 0.049) & $<0.001$ \\
DRS~\cite{DRS} & 0.789 (0.778, 0.800) & 0.065 (0.056, 0.074) & $<0.001$ \\
CAIDE~\cite{CAIDE} & 0.704 (0.691, 0.717) & 0.149 (0.136, 0.162) & $<0.001$ \\
CAIDE (w/o \textit{APOE} $\varepsilon$4) & 0.654 (0.641, 0.666) & 0.200 (0.186, 0.214) & $<0.001$ \\
ANU-ADRI~\cite{ANU_ADRI} & 0.592 (0.575, 0.608) & 0.262 (0.243, 0.280) & $<0.001$ \\
\bottomrule
\end{tabular}
\footnotetext{All models were evaluated on the same holdout test set ($n = 50{,}041$; $1{,}138$ incident dementia cases), under identical outcome definitions, using isotonic-calibrated predictions. The upper panel reports average precision and the lower panel reports ROC AUC for the same models. Differences are ITU-DESC minus the comparator and were calculated from unrounded estimates. ROC AUC estimates, confidence intervals, differences and two-sided $P$ values were derived using DeLong's method. AP confidence intervals were obtained by paired subject-level bootstrap resampling (10{,}000 resamples; percentile method), whereas AP $P$ values were obtained using two-sided paired permutation tests (2{,}000 permutations). AP, average precision; CI, confidence interval; ROC AUC, area under the receiver operating characteristic curve.}
\end{table}

For Alzheimer's disease, ITU-DESC again showed the highest discrimination (Additional file 1: Table~S12), with an AP of 0.070 (95\% CI, 0.059 to 0.085) and a ROC AUC of 0.863 (0.850 to 0.876). UKB-DRP ranked second, with an AP of 0.056 (0.047 to 0.067) and a ROC AUC of 0.844 (0.829 to 0.858); the differences from ITU-DESC were 0.014 (0.005 to 0.025) for AP and 0.019 (0.010 to 0.028) for ROC AUC. All scores ranked in the same order as for all-cause dementia, and ITU-DESC exceeded every score on both metrics ($P<0.01$ for every comparison). AP fell for every model relative to all-cause dementia, consistent with the lower prevalence of Alzheimer's disease (513 versus 1{,}138 cases). ROC AUC, which does not depend on prevalence, rose slightly for ITU-DESC (0.863 versus 0.853).

\subsection{Discrimination across time horizons}
\begin{figure}[htbp]
    \centering
    \includegraphics{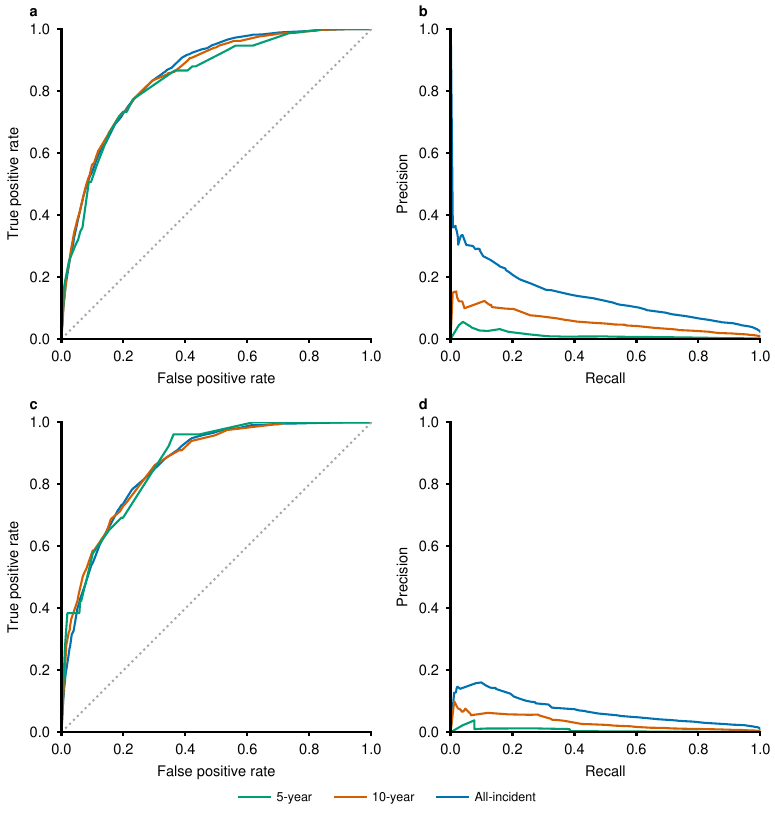}
    \caption{Discrimination of ITU-DESC across the 5-year, 10-year and all-incident horizons. A single fitted model underlies all three horizons, evaluated on the holdout test set using calibrated predictions. \textbf{a}, Receiver operating characteristic curves for all-cause dementia. \textbf{b}, Precision--recall curves for all-cause dementia. \textbf{c}, Receiver operating characteristic curves for Alzheimer's disease. \textbf{d}, Precision--recall curves for Alzheimer's disease. The dotted line marks chance discrimination in \textbf{a} and \textbf{c}. Discrimination values are reported in Additional file 1: Table~S13, and the case counts underlying each horizon in Fig.~\ref{fig:ukbiobank-selection-process}.}
    \label{fig:temporal_performance}
\end{figure}

ITU-DESC produces a single risk score for each participant, which we evaluated on the holdout test set at the 5-year, 10-year and all-incident horizons. The base model was fitted once and assigned each participant the same raw score at every horizon; the evaluations differed only in the outcome label used for each time window. Differences in discrimination across horizons therefore reflect differences in the outcome definition rather than changes in the underlying model. Each horizon was assigned a separate isotonic calibration mapping, so the calibrated risks differed in scale and in the ties introduced by calibration, while preserving the underlying ordering of the raw scores (Fig.~\ref{fig:temporal_performance}; Additional file 1: Table~S13). For all-cause dementia, ROC AUC increased modestly with longer follow-up, from 0.837 at 5 years to 0.850 at 10 years and 0.853 for the all-incident outcome. AP increased more substantially, from 0.012 to 0.056 and 0.138 across the same horizons.

For Alzheimer's disease, ROC AUC was similar across horizons: 0.868 at 5 years, 0.866 at 10 years and 0.863 for the all-incident outcome. AP increased from 0.007 at 5 years to 0.030 at 10 years and 0.070 for all-incident Alzheimer's disease. Thus, ranking performance changed little as the outcome window widened, whereas AP increased with outcome prevalence, which rose from 0.15\% to 0.84\% to 2.27\% for dementia and from 0.05\% to 0.33\% to 1.03\% for Alzheimer's disease (Fig.~\ref{fig:ukbiobank-selection-process} reports the corresponding case counts). Because the AP of a non-informative model equals the outcome prevalence, the increase in AP across horizons is consistent with the accumulation of cases rather than improved ranking performance. Median follow-up was 16.6 years (interquartile range, 15.8--17.4), and every participant had at least 15.1 years of potential follow-up; no participant was therefore censored before the 5- or 10-year horizons. At these horizons, the binary labels coincide with the observed cumulative-incidence status, with death before dementia treated as a non-event (Additional file 1: Table~S20). A sensitivity analysis treating death as a competing event (Additional file 1: Supplementary Note, ``Competing risk of death''; Table~S22) did not alter the ranking of the scores at any horizon. Corresponding analyses based on uncalibrated predictions are shown in Additional file 1: Table~S14 and Fig.~S2.

\subsection{Calibration}
\begin{figure}[htbp]
    \centering
    \includegraphics{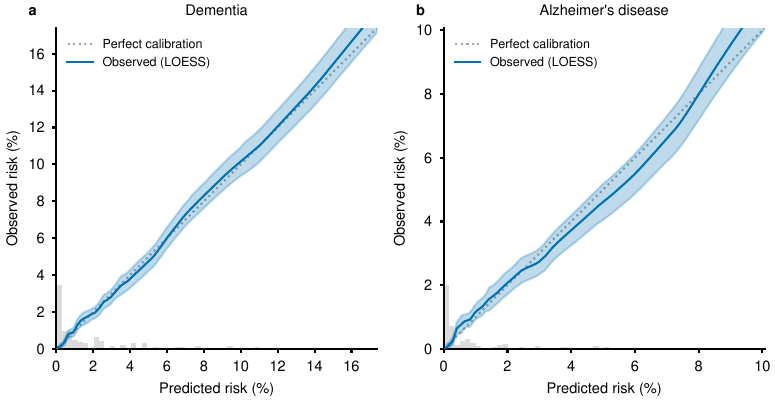}
    \caption{Flexible calibration of ITU-DESC for the all-incident outcomes. Both panels use the holdout test set. \textbf{a}, All-cause dementia. \textbf{b}, Alzheimer's disease. Observed risk, estimated by locally estimated scatterplot smoothing (LOESS), is plotted against predicted risk with pointwise 95\% confidence bands, calculated as the smoothed estimate plus or minus 1.96 standard errors of the local observed proportion. Histograms along the bottom of each panel show the distribution of predicted risks. Calibration slope, mean predicted risk against observed incidence and Brier score are given in Additional file 1: Table~S23.}
    \label{fig:calibration}
\end{figure}

We assessed ITU-DESC calibration for the all-incident outcomes of all-cause dementia and Alzheimer's disease. For each outcome, we plotted a flexible calibration curve against the line of perfect calibration (Fig.~\ref{fig:calibration}). Calibration curves were estimated using locally estimated scatterplot smoothing (LOESS) of observed incidence against predicted risk. Pointwise 95\% confidence bands were calculated as the smoothed estimate plus or minus 1.96 standard errors. At each predicted-risk value the standard error was that of the local observed proportion and therefore depended on how many participants contributed to the fit at that point. For both outcomes, the curves closely followed the diagonal, indicating close agreement between predicted and observed risks. Uncertainty increased at the upper end of the predicted-risk distribution, where fewer participants were represented, as shown by the accompanying risk-distribution histograms. This region was also sparsely populated when the isotonic calibration mappings were fitted, making calibrated estimates in the upper-risk tail more susceptible to overfitting.

Weak-calibration summaries were consistent with this pattern. For both outcomes the calibration slope was close to one and the calibration intercept close to zero, and mean predicted risk closely matched observed incidence ($2.27\%$ versus $2.27\%$ for dementia; $1.02\%$ versus $1.03\%$ for AD). Calibration-in-the-large is reported on both the probability and log-odds scales, which agree in indicating no systematic over- or underestimation. In the competing-risk sensitivity analysis, the mean calibrated 10-year ITU-DESC risk tracked the Aalen--Johansen cumulative incidence of dementia in every risk decile (Additional file 1: Table~S21 and Fig.~S3). Because incident outcomes were rare, the raw Brier scores were close to their no-information baselines; we therefore also report the scaled Brier score, which indicated a modest but consistent improvement over a non-informative model. Full calibration statistics, with 95\% CIs and metric definitions, are reported in Additional file 1: Supplementary Note (Table~S23).

\subsection{Feature importance}
\begin{figure}[htbp]
    \centering
    \includegraphics{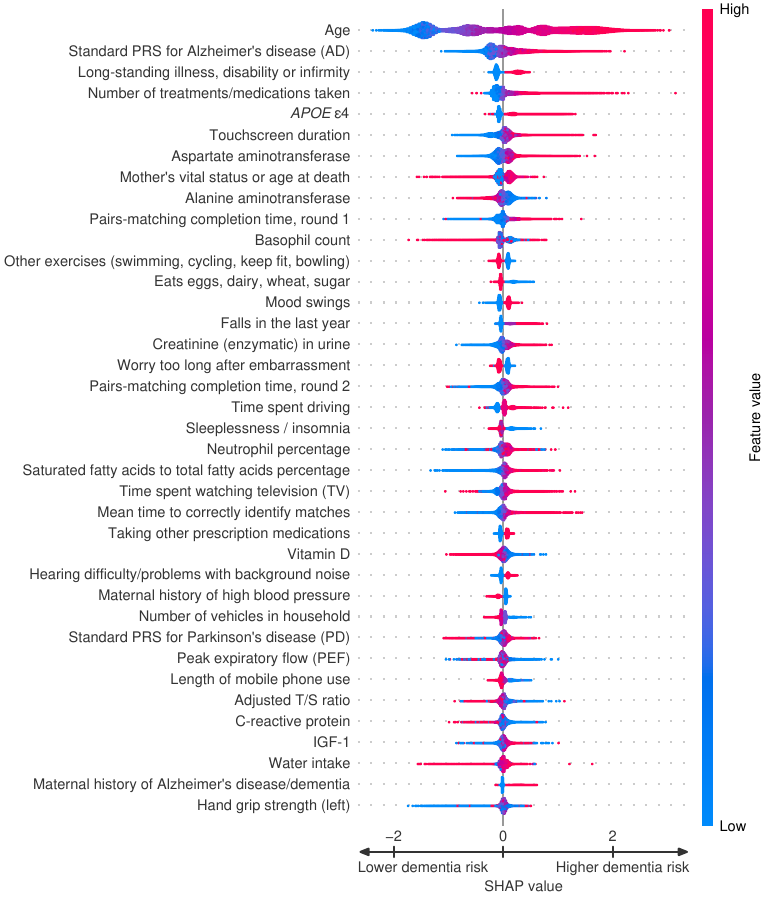}
    \caption{SHAP summary plot for ITU-DESC for all-incident dementia. Each point represents one participant and is plotted according to the feature's SHAP value, corresponding to its signed contribution to the model output on the log-odds scale. Points are colored by feature value. Features are ordered from top to bottom by mean absolute SHAP value. SHAP values were computed on the holdout test set ($n = 50{,}041$), using the model refit on the full development set.}
    \label{fig:mean_shap}
\end{figure}

The SHAP summary plot (Fig.~\ref{fig:mean_shap}) shows each feature's signed contribution to ITU-DESC predictions, with color indicating the corresponding feature value. Features are ranked by mean absolute SHAP value computed on the holdout test set. The complete list of selected features is provided in Additional file 1: Supplementary Note (Table~S15).

\begin{figure}
    \centering
    \includegraphics{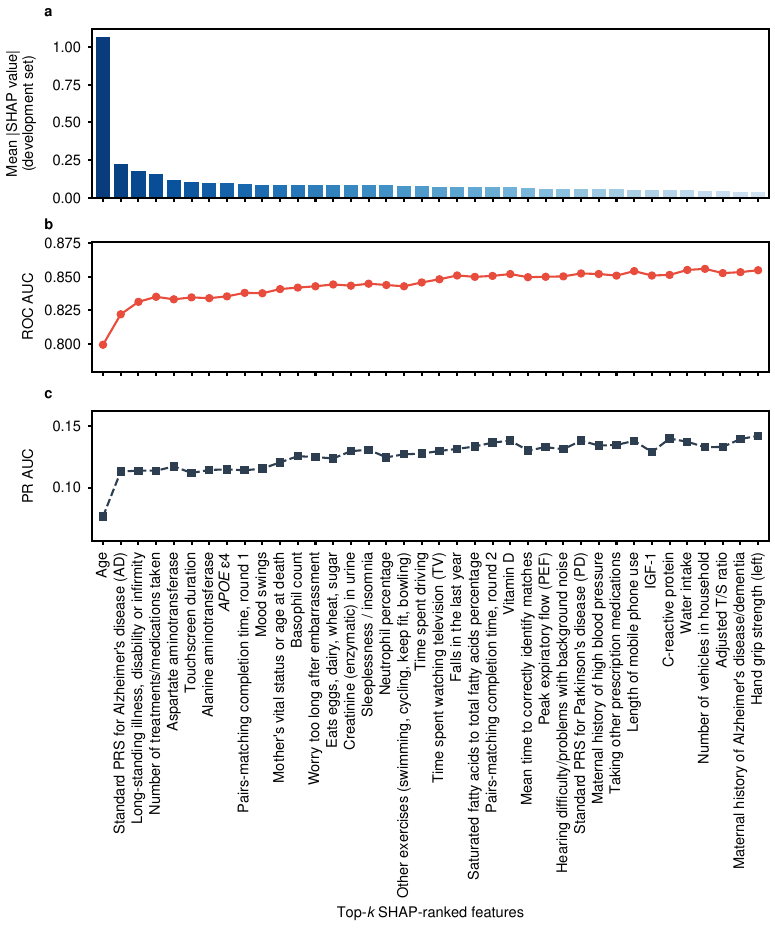}
    \caption{Cumulative discrimination of ITU-DESC as predictors are added in SHAP-rank order for all-incident dementia. The x-axis lists the top-$k$ features ranked by mean absolute SHAP value computed on the 90\% development set. \textbf{a}, Development-set SHAP importance of each feature. \textbf{b}, ROC AUC when the model is refit on the development set using the top-$k$ features and evaluated on the holdout test set ($n = 50{,}041$). \textbf{c}, Precision--recall AUC (average precision) for the same refits. Each point is a single holdout-test evaluation.}
    \label{fig:cumulative_auc}
\end{figure}

After identifying the final set of 38 predictors, we examined whether predictive performance was concentrated in a small subset of variables or distributed across the selected feature set. We evaluated cumulative discrimination by adding predictors in SHAP-rank order (Fig.~\ref{fig:cumulative_auc}), adapting an incremental feature-addition approach from You et al.~\cite{YOU2022101665}. Here the ranking was derived from mean absolute SHAP values computed on the development set rather than the holdout-test values reported above, so the ordering in which features enter never uses test data. For each top-$k$ feature set, the model was refit on the development set and evaluated on the holdout test set, and ROC AUC and AP were recorded. Age alone gave a ROC AUC of 0.799 and an AP of 0.076; the full 38-feature model reached 0.854 and 0.142. ROC AUC peaked at 0.856 with 35 features and stayed within 0.007 of that peak from 21 features onward, so the first 21 predictors carried nearly all of the ranking performance. These values come from uncalibrated scores and from a separate refit at each top-$k$; the calibrated ITU-DESC results are in Table~\ref{tab:risk_scores_table} (0.853; 0.138) and the uncalibrated ones in Additional file 1: Table~S10 (0.854; 0.142).

To illustrate individual-level model predictions, we also examined local SHAP explanations for representative true-positive, true-negative, false-positive and false-negative predictions. These examples show how different combinations of feature contributions led to correct and incorrect classifications (Additional file 1: Figs.~S5, S6, S7 and~S8; Supplementary Note).

Because random imputation removes predictive information carried by missingness, the ranking in Fig.~\ref{fig:mean_shap} could depend on the imputation strategy. However, refitting the model using CatBoost's native handling of missing values produced the same features at the top and bottom of the ranking, indicating that the ordering interpreted here was robust to the choice of missing-data strategy (Additional file 1: Fig.~S4; Supplementary Note).

\subsection{Selected predictors}

Age was the most important predictor in ITU-DESC (Fig.~\ref{fig:mean_shap}), consistent with age being the dominant risk factor for dementia. This variable refers to the participant's age at the initial UK Biobank assessment visit.

Family history contributed three predictors. The model retained maternal history of Alzheimer's disease or dementia and maternal history of high blood pressure, but no paternal counterpart. It also retained the combined maternal survival and age-at-death variable described in Section~\ref{subsec:feature-preprocessing}.

Several selected predictors captured general health status or neuropsychiatric symptoms: long-standing illness, disability or infirmity; number of treatments or medications taken; taking other prescription medications; falls in the last year; sleeplessness or insomnia; mood swings; worrying too long after embarrassment; and hearing difficulty or problems with background noise.

Blood-based and urinary biomarkers also contributed to prediction. These included aspartate aminotransferase (AST), alanine aminotransferase (ALT), neutrophil percentage, basophil count, C-reactive protein, insulin-like growth factor 1 (IGF-1), vitamin D, saturated fatty acids as a percentage of total fatty acids and enzymatic creatinine in urine.

The model retained leukocyte telomere length as the telomere-to-single-copy-gene ratio adjusted for assay-related technical parameters~\cite{ukbiobank_field22191,codd2022telomere}. Genetic predictors included \textit{APOE} $\varepsilon$4 allele copy number~\cite{bycroft2018ukbiobank}, the standard polygenic risk score (PRS) for Alzheimer's disease and the standard PRS for Parkinson's disease~\cite{thompson2024ukbprs}.

Cognitive, sensorimotor and task-performance measures were also selected. These included touchscreen duration, pairs-matching completion time for rounds 1 and 2, mean time to correctly identify matches, left hand grip strength and peak expiratory flow (PEF). You et al.~\cite{YOU2022101665} used the summed first and second rounds of UK Biobank Field 400~\cite{ukbiobank_field400}; our pipeline instead retained the individual rounds.

Lifestyle and socioeconomic variables were retained as well. Behavioral predictors included time spent watching television; time spent driving; types of physical activity in the last 4 weeks; water intake; consumption of eggs, dairy, wheat or sugar; and length of mobile phone use. For physical activity, the retained category corresponded to other exercises, such as swimming, cycling, keep fit or bowling (Additional file 1: Supplementary Methods, ``Feature sets and related data fields''). The model also retained the number of vehicles in the household as a socioeconomic predictor.

Together, these variables formed Set 8, the 38 features selected in at least eight of the ten imputation seeds. Of these, 29 were selected by every seed, 5 by nine seeds and 4 by eight seeds; all 38 are listed with their data-field identifiers in Additional file 1: Table~S15. Further variables fell just below the cutoff: five were selected in seven seeds, one in six seeds and two in five seeds.

The seven-seed variables were tea intake, right and left leg fat percentage, glucose and urea; omega-3 fatty acids was selected in six seeds; and seating box height and average total household income before tax were selected in five. Although these variables did not enter the final feature set, their repeated selection across half or more of the seeds suggests that they may carry additional predictive information.

\subsection{Missingness of the selected predictors}
\label{subsec:missingness-results}
Missingness in the selected predictors was associated with incident dementia. Of the 38 predictors, age was complete for all participants. For each of the remaining 37, missingness was more frequent among participants who subsequently developed dementia than among those who did not (Additional file 1: Table~S18; the corresponding results for Alzheimer's disease are in Table~S19). For 33 of the 37 predictors, this association persisted after adjustment for age, sex, education, deprivation, recruitment year, assessment center and ethnicity (adjusted odds ratios, 1.07 to 2.44). Dementia incidence rose from 1.76\% among participants with no missing predictors to 3.61\% among those missing three or four (Additional file 1: Table~S17; Supplementary Note). Baseline characteristics of participants complete on all 38 predictors and those missing at least one are compared in Additional file 1: Table~S16.

\section{Discussion}
In this UK Biobank study, an explainable machine-learning model developed directly from cohort data outperformed established dementia risk scores. Starting from approximately 8{,}000 candidate features, the wrapper-based selection pipeline identified 38 predictors. ITU-DESC achieved higher discrimination than the strongest comparator, UKB-DRP (Table~\ref{tab:risk_scores_table}). Several selected predictors that are absent from established risk scores, including leukocyte telomere length and the liver enzymes aspartate aminotransferase (AST) and alanine aminotransferase (ALT), illustrate how wrapper-based feature selection can identify biologically plausible predictors that are not included in hand-crafted scores. This discrimination was accompanied by well-calibrated predictions: after outcome-specific isotonic recalibration, calibration curves closely followed the diagonal for both outcomes, with calibration slopes near one and calibration-in-the-large near zero. ITU-DESC predictions can therefore be interpreted as estimated probabilities rather than solely as relative risk rankings, although these probabilities remain specific to the outcome definitions, follow-up periods and ascertainment framework of the present study.

The primary objective of this study was to determine whether leakage-aware machine-learning feature selection across a broad UK Biobank feature space could improve discrimination and identify additional predictors. You et al.~\cite{YOU2022101665} and Anat\"urk et al.~\cite{Anaturk2023} emphasized cost, clinical usability and implementation burden; these remain important for translation into practice but were beyond the present scope.

Age was the most important predictor in ITU-DESC, consistent with the findings of You et al.~\cite{YOU2022101665}. Both of the models identified a shared core of predictors, including age, \textit{APOE} $\varepsilon$4, cognitive and task-performance measures, medication burden, mother's vital status or age at death and long-standing illness.

Our model combined maternal survival and maternal age at death into a single variable, mother's vital status or age at death, which follows the interpretation proposed by You et al.: a missing value for mother's age at death usually indicates that the participant's mother was still alive~\cite{mothers_age_at_death_ukbiobank}. Two further features identified by You et al.\ behaved differently in our pipeline. Peak expiratory flow entered our final feature set, selected in nine of ten seeds, whereas leg fat percentage fell just below the cutoff at seven seeds. Their reaction-time measure, the duration to first press of the snap-button in each round, was selected by no seed at all; the related pairs-matching timings were retained instead.

The liver enzymes AST and ALT and leukocyte telomere length, all absent from established dementia risk scores, have independent biological support. A higher AST-to-ALT ratio has been linked with increased risk of cognitive impairment~\cite{li2022increased}. Evidence for associations involving AST and ALT also extends beyond UK Biobank: a systematic review summarized findings across multiple cohorts~\cite{cog_dec_and_dem}, including studies of both dementia and broader cognitive impairment. The adjusted telomere-to-single-copy-gene ratio was also retained, consistent with the hypothesis that telomere-related aging may provide predictive information beyond chronological age~\cite{codd2022telomere}.

Several additional selected features relate to behavior and demographics, and their associations with dementia may be indirect. Sex was not retained at all: no imputation seed selected it, which suggests that other selected variables already capture the information it carries.

Time spent watching television and types of physical activity have been examined in relation to dementia risk in previous work~\cite{television_and_computer}. Li et al.\ reported that, among socially inactive older adults, moderate computer use was associated with lower dementia risk, whereas heavy computer use and television viewing above approximately 2 hours per day were associated with higher risk. Length of mobile phone use was negatively associated with incident dementia in our model, but this association may partly reflect age or cohort effects, because younger participants are likely to have used mobile phones for longer.

Evaluating every comparator under the random-imputation policy used for ITU-DESC (Section~\ref{subsec:nan-value-data-leakage}) eliminated cohort differences that would otherwise have confounded comparisons across scores. For example, \textit{APOE} $\varepsilon$4 status was missing for 88{,}961 of the 500{,}410 eligible participants. Applying complete-case analysis, as UKBDRS specifies, would therefore have restricted the \textit{APOE}-inclusive UKBDRS to a smaller, potentially selected subset, whereas the other models could be evaluated in the full sample. Under random imputation, all eight models were instead evaluated in the same 50{,}041 holdout-test participants, using identical outcome definitions and with identical outcome prevalence. Consequently, differences in discrimination are attributable to differences between the scores rather than to differences in the cohorts on which they were evaluated.

The shared missing-data policy introduces a trade-off that should be considered when interpreting the benchmark results. UKBDRS, CAIDE, DRS and ANU-ADRI were developed under different missing-data procedures, and evaluating them on randomly imputed data may therefore attenuate their performance relative to published estimates, particularly for scores containing variables with substantial missingness. Because a randomly imputed value is independent of the participant to whom it is assigned, it contributes no participant-specific predictive information. Accordingly, differences in AP or ROC AUC between ITU-DESC and these scores should not be attributed to model structure alone.

\subsection{Limitations}

Random imputation reduced reliance on deterministic missingness patterns but may also have introduced noise and weakened genuine predictive signal. Under informative missingness, no single-imputation procedure can recover the target regression function unless information about missingness is also available to the model~\cite{josse2024consistency}. Because donors were drawn from each variable's marginal distribution, the procedure also attenuated relationships between variables that a conditional donor pool could retain~\cite{chaput2015joint}. Conditioning donor selection on the remaining covariates, or guiding it with predictive models~\cite{tripet2024donor}, would preserve that structure but would reintroduce the dependence on measurement availability we set out to suppress. Whether an imputation rule can preserve that structure while withholding missingness patterns from the model remains an open question. Our inclusive design retained all eligible participants (Section~\ref{subsec:dataset}), which maximized sample size and avoided additional selection based on missingness but may have increased the variance introduced by imputation and made performance more sensitive to the chosen missing-data strategy.

Our evaluation used binary classification rather than time-to-event analysis, and participants who died without a recorded dementia diagnosis were counted as non-cases. UKBDRS was originally developed and evaluated in a competing-risk framework that accounts for follow-up time, censoring and death as a competing event, so the AP and ROC AUC reported here should not be interpreted as reproducing its published 14-year performance. At the 5- and 10-year horizons, the sensitivity analysis treating death as a competing event supported the binary results (Additional file 1: Supplementary Note, ``Competing risk of death''), but the all-incident outcome combines follow-up durations from 15.1 to 19.7 years and therefore does not represent a fixed-horizon cumulative incidence. In addition, non-case status reflects the absence of a recorded diagnosis in the current data release, so the observed incidence is a lower bound over the same follow-up and performance estimates may change modestly as additional diagnoses accrue.

Volunteer cohorts such as UK Biobank are susceptible to healthy-volunteer bias~\cite{heathybias}. Selection also affects the imaging subsample, whose participants were healthier on average than the rest of the cohort, so analyses restricted to it may underestimate exposure--outcome associations~\cite{imaging_bias}. Phenotypes from the imaging visits lay outside our baseline-only feature set, which also prevented imaging availability from acting as a proxy for participant selection or health status.

We used default hyperparameters and did not perform hyperparameter tuning, because tuning nested within BorutaSHAP feature selection and repeated across ten imputation seeds was computationally prohibitive. This may have limited model performance and should be addressed in future work.

Because ITU-DESC was developed and internally validated in UK Biobank, its generalizability to other populations remains uncertain. We also included the full eligible cohort across the available baseline age range, whereas many existing risk scores were developed for narrower age ranges or target populations, so direct comparisons should be interpreted with this difference in mind.

\subsection{Future work}

External validation in independent cohorts with different demographic, clinical and geographic characteristics is required to confirm discrimination and to assess whether calibration transports and whether the selected feature set and missing-data strategy remain effective in other populations and data-collection settings. Clinical usability and implementation burden should also be evaluated before deployment. Application of ITU-DESC to a new dataset requires the same 38 baseline predictors and the same unconditional random hot-deck imputation procedure used here, with donor values drawn from the target population or, if unavailable, from the UK Biobank development set. ITU-DESC is intended for researchers and analysts working with tabular cohort data and requires a qualified user to prepare the inputs and interpret the calibrated probabilities. It is not intended for autonomous clinical use.

Future work could evaluate more principled approaches for informative and nonmonotone missingness, including statistical frameworks designed for MNAR data with limited complete cases~\cite{cheng2022handlingnonmonotonemissingdata}. Our study did not incorporate advanced genetic biomarkers beyond \textit{APOE} $\varepsilon$4 status and polygenic risk scores. Future work could evaluate whether richer genomic information, including whole-genome sequencing (WGS), improves dementia risk prediction, although cost, scalability and interpretability remain important considerations. Repeated UK Biobank assessments could be used to model longitudinal change, especially among participants who later develop dementia, to reveal pre-diagnostic trajectories that a single baseline visit misses. Imaging-derived predictors could also be evaluated in the imaging subsample, with attention to the selection bias described above. These extensions should be weighed against their additional complexity, data requirements and reduced interpretability.

The feature-selection and missing-data pipeline is not specific to dementia. Applying it to other UK Biobank outcomes with informative missingness would test the general framework rather than the ITU-DESC score itself and would show whether consensus feature selection and random imputation perform similarly under different missingness structures. More broadly, wrapper-based selection methods such as BorutaSHAP~\cite{borutashap} or Boruta~\cite{boruta} may be useful in other UK Biobank tabular-data studies by reducing very large candidate feature spaces to more explainable and manageable predictor sets.

\section{Conclusions}

Using UK Biobank data, we developed ITU-DESC, an explainable machine-learning dementia risk score based on 38 predictors selected from a broad baseline feature space. ITU-DESC showed higher discrimination on a holdout test set than established dementia risk scores for all-cause dementia and Alzheimer's disease, while identifying predictors absent from existing scores, such as leukocyte telomere length and liver-enzyme markers. These findings suggest that missingness-aware machine-learning feature selection can improve dementia risk prediction in large-scale cohort data. External validation is required to determine whether its performance and selected predictors generalize beyond UK Biobank.

\backmatter

\section*{Declarations}

\bmhead{Ethics approval and consent to participate}
UK Biobank has Research Tissue Bank approval from the North West -- Haydock Research Ethics Committee (REC reference 21/NW/0157), and all participants provided written informed consent. This study was conducted in accordance with the UK Biobank Ethics and Governance Framework and the Declaration of Helsinki.

\bmhead{Consent for publication}
Not applicable.

\bmhead{Availability of data and materials}
{\sloppy
UK Biobank data~\cite{sudlow2015ukbiobank} are used under license and not publicly available; they are available to approved researchers through the UK Biobank Access Management System~\cite{ukbiobank_apply_for_access}. The columns of the candidate feature set and the entries excluded a priori are listed in Additional file 2. The custom analysis code used in this study is not publicly available but will be made available to editors and reviewers upon request.
\par}

\bmhead{Competing interests}
The authors declare that they have no competing interests.

\bmhead{Funding}
Abdullah Kavakl\i{} received support from the Global Researcher Access Fund to cover the UK Biobank application costs. Compute and storage costs were supported by the UK Biobank Platform Credits Programme. Faik Boray Tek's contribution to this work was supported by the Scientific and Technological Research Council of T\"urkiye (T\"UB\.ITAK) under Grant Number 122E509. The funders had no role in study design; data collection; analysis or interpretation; manuscript preparation; or the decision to submit the manuscript for publication.

\bmhead{Authors' contributions}
A.K. and F.B.T. jointly designed and planned the study. A.K. curated and prepared the data; carried out the experiments and analyses; and wrote the initial draft of the manuscript. F.B.T. supervised the project and critically revised the manuscript. Both authors read and approved the final manuscript.

\bmhead{Acknowledgements}
This research has been conducted using the UK Biobank Resource under Application Number 412140.

\section*{Additional files}

\textbf{Additional file 1} (.pdf)\quad\textbf{Supplementary Information.}
Supplementary Methods, Supplementary Note, Tables and Figures.

\noindent\textbf{Additional file 2} (.xlsx)\quad\textbf{Candidate feature set.}
Candidate columns and entries excluded a priori.

\end{document}

\title{ITU-DESC: a missingness-aware machine-learning dementia risk score in the UK Biobank cohort}
\subtitle{Supplementary Information}

\author*[1]{\fnm{Abdullah} \sur{Kavakl\i{}}}\email{kavakli23@itu.edu.tr}
\author[1]{\fnm{Faik Boray} \sur{Tek}}

\affil*[1]{\orgdiv{Department of Artificial Intelligence and Data Engineering}, \orgname{Istanbul Technical University}, \orgaddress{\city{Istanbul}, \country{Turkey}}}

\maketitle

\section{Supplementary Methods}

\subsection{Development and holdout test split}

The eligible cohort of 500{,}410 participants was divided into a 90\% development set (450{,}369 participants) and a 10\% holdout test set (50{,}041 participants), as described in the main text. This section documents the stratification scheme, the acceptance criteria and the resulting balance between the two sets. Ethnicity was self-reported (UK Biobank Field 21000) and grouped into the seven categories in Table~\ref{tab:split_balance}. It served as a split acceptance criterion and as an adjustment covariate in the missingness sensitivity analysis.

The stratification label combined sex with the six binary outcome indicators: 5-year, 10-year and all-incident outcomes for both all-cause dementia and Alzheimer's disease. For strata with fewer than 10 participants, we omitted sex and merged those participants into the corresponding sex-independent outcome pattern. We then merged outcome patterns with fewer than two participants into a residual stratum.

We prespecified three candidate random seeds and four balance criteria, then evaluated the seeds in order and retained the first that met all four. Two criteria concerned the outcomes: at least one case of each outcome in the holdout test set and a holdout share of cases for each outcome within five percentage points of the target 10\%. Because we stratified the split on outcome, both criteria only confirmed that the stratification worked. The two substantive criteria concerned covariates outside the stratification scheme: a standardized mean difference (SMD) for age below 0.10 and every assessment-center and ethnicity category within two percentage points between the development and holdout test sets. The first candidate seed met all four criteria, so we never evaluated the other two.

Table~\ref{tab:split_outcomes} reports outcome counts in the development and holdout test sets. Every outcome had cases in the holdout test set, and for every outcome the case proportion in the holdout test set differed from that in the development set by less than 0.01 percentage points, so the holdout share of cases was close to the 10\% target and both outcome criteria were met.

\begin{table}[htb]
    \centering
    \caption{Outcome counts and prevalences in the development and holdout test sets.}
    \label{tab:split_outcomes}
    \small
    \begin{tabular}{lrrrr}
    \toprule
& \multicolumn{2}{c}{Development ($n=450{,}369$)} & \multicolumn{2}{c}{Holdout test ($n=50{,}041$)} \\
    \cmidrule(lr){2-3}\cmidrule(lr){4-5}
Outcome & Cases & \% & Cases & \% \\
    \midrule
Dementia, 5-year & 681 & 0.151 & 75 & 0.150 \\
Dementia, 10-year & 3{,}802 & 0.844 & 421 & 0.841 \\
Dementia, all-incident & 10{,}242 & 2.274 & 1{,}138 & 2.274 \\
Alzheimer's disease, 5-year & 231 & 0.051 & 26 & 0.052 \\
Alzheimer's disease, 10-year & 1{,}506 & 0.334 & 167 & 0.334 \\
Alzheimer's disease, all-incident & 4{,}609 & 1.023 & 513 & 1.025 \\
    \bottomrule
    \end{tabular}
    \end{table}
    
    Table~\ref{tab:split_balance} reports the full balance summary. Sex, the only non-outcome variable in the stratification scheme, differed by 0.002 percentage points between the development and holdout test sets. Three further variables served as acceptance criteria without entering the stratification scheme, and all were similarly well balanced. Age at recruitment had an SMD of 0.0057 (mean, 56.54 versus 56.50 years), well below the 0.10 threshold. Category proportions differed by at most 0.21 percentage points across the 22 assessment centers and 0.06 across the seven ethnicity categories, both within the two-percentage-point margin. Availability of the principal predictors entered neither the stratification scheme nor the acceptance criteria, but was also closely balanced, with the largest difference 0.16 percentage points for \textit{APOE} $\varepsilon$4 genotype. This balance matters because informative missingness could inflate performance estimates in this cohort: the holdout test set closely matched the development set in the availability of measured predictors.    

\begingroup\tablebodyfont %
\setlength{\tabcolsep}{4pt}
\setlength{\LTcapwidth}{\linewidth}\setlength{\LTleft}{0pt}\setlength{\LTright}{0pt}
\begin{longtable}{@{\extracolsep{\fill}}>{\raggedright\arraybackslash}p{6.2cm}rrrrr}
\caption{Balance of the 90\% development and 10\% holdout test sets. Percentages are column percentages within each set; \textit{Diff.} is the absolute difference in percentage points. For age at recruitment, a continuous variable, the \% cells give the mean age in years and \textit{Diff.} gives the standardized mean difference (SMD). For predictor availability, $n$ and \% are the participants with a recorded value. Assessment centers are given as UK Biobank Field 54 codes. Ethnicity categories derive from UK Biobank Field 21000. The split was stratified on the six outcomes (Table~\ref{tab:split_outcomes}) and sex. Age, ethnicity and assessment center were prespecified acceptance criteria checked after the split, not stratification variables. Predictor availability was reported but neither stratified on nor subject to a threshold.}
\label{tab:split_balance}\\
\toprule
& \multicolumn{2}{c}{Development} & \multicolumn{2}{c}{Holdout test} & \\
\cmidrule(lr){2-3}\cmidrule(lr){4-5}
Variable & $n$ & \% & $n$ & \% & Diff. \\
\midrule
\endfirsthead
\multicolumn{6}{l}{\textit{Table~\ref{tab:split_balance} continued from previous page}}\\
\toprule
& \multicolumn{2}{c}{Development} & \multicolumn{2}{c}{Holdout test} & \\
\cmidrule(lr){2-3}\cmidrule(lr){4-5}
Variable & $n$ & \% & $n$ & \% & Diff. \\
\midrule
\endhead
\midrule
\multicolumn{6}{r}{\textit{Continued on next page}}\\
\endfoot
\bottomrule
\endlastfoot
\multicolumn{6}{l}{\textbf{Sex (stratified on)}}\\
\quad Female & 245{,}018 & 54.404 & 27{,}225 & 54.405 & 0.002 \\
\quad Male & 205{,}351 & 45.596 & 22{,}816 & 45.595 & 0.002 \\
\multicolumn{6}{l}{\textbf{Age at recruitment (acceptance criterion)}}\\
\quad Mean, years & 450{,}369 & 56.541 & 50{,}041 & 56.495 & 0.0057 \\
\multicolumn{6}{l}{\textbf{Ethnicity (acceptance criterion)}}\\
\quad Asian or Asian British & 8{,}856 & 1.966 & 979 & 1.956 & 0.010 \\
\quad Black or Black British & 7{,}235 & 1.606 & 775 & 1.549 & 0.058 \\
\quad Chinese & 1{,}385 & 0.308 & 177 & 0.354 & 0.046 \\
\quad Mixed & 2{,}645 & 0.587 & 289 & 0.578 & 0.010 \\
\quad Other & 4{,}083 & 0.907 & 440 & 0.879 & 0.027 \\
\quad Unknown/Not stated & 2{,}448 & 0.544 & 303 & 0.606 & 0.062 \\
\quad White & 423{,}717 & 94.082 & 47{,}078 & 94.079 & 0.003 \\
\multicolumn{6}{l}{\textbf{Assessment center (acceptance criterion)}}\\
\quad 10003 & 3{,}407 & 0.756 & 364 & 0.727 & 0.029 \\
\quad 11001 & 12{,}496 & 2.775 & 1{,}354 & 2.706 & 0.069 \\
\quad 11002 & 12{,}529 & 2.782 & 1{,}451 & 2.900 & 0.118 \\
\quad 11003 & 16{,}071 & 3.568 & 1{,}769 & 3.535 & 0.033 \\
\quad 11004 & 16{,}742 & 3.717 & 1{,}851 & 3.699 & 0.018 \\
\quad 11005 & 15{,}421 & 3.424 & 1{,}729 & 3.455 & 0.031 \\
\quad 11006 & 17{,}450 & 3.875 & 1{,}904 & 3.805 & 0.070 \\
\quad 11007 & 26{,}293 & 5.838 & 2{,}928 & 5.851 & 0.013 \\
\quad 11008 & 25{,}304 & 5.619 & 2{,}879 & 5.753 & 0.135 \\
\quad 11009 & 33{,}222 & 7.377 & 3{,}676 & 7.346 & 0.031 \\
\quad 11010 & 39{,}625 & 8.798 & 4{,}421 & 8.835 & 0.036 \\
\quad 11011 & 38{,}489 & 8.546 & 4{,}332 & 8.657 & 0.111 \\
\quad 11012 & 11{,}195 & 2.486 & 1{,}284 & 2.566 & 0.080 \\
\quad 11013 & 30{,}287 & 6.725 & 3{,}445 & 6.884 & 0.159 \\
\quad 11014 & 27{,}331 & 6.069 & 2{,}948 & 5.891 & 0.177 \\
\quad 11016 & 29{,}437 & 6.536 & 3{,}257 & 6.509 & 0.028 \\
\quad 11017 & 19{,}124 & 4.246 & 2{,}102 & 4.201 & 0.046 \\
\quad 11018 & 25{,}877 & 5.746 & 2{,}832 & 5.659 & 0.086 \\
\quad 11020 & 24{,}462 & 5.432 & 2{,}788 & 5.571 & 0.140 \\
\quad 11021 & 22{,}963 & 5.099 & 2{,}448 & 4.892 & 0.207 \\
\quad 11022 & 2{,}064 & 0.458 & 210 & 0.420 & 0.039 \\
\quad 11023 & 580 & 0.129 & 69 & 0.138 & 0.009 \\
\multicolumn{6}{l}{\textbf{Predictor availability (reported only)}}\\
\quad \textit{APOE} $\varepsilon$4 genotype & 370{,}376 & 82.238 & 41{,}073 & 82.079 & 0.160 \\
\quad Leukocyte telomere length (Field 22191) & 423{,}613 & 94.059 & 47{,}065 & 94.053 & 0.006 \\
\quad Alzheimer's disease PRS (Field 26206) & 435{,}801 & 96.765 & 48{,}476 & 96.873 & 0.107 \\
\quad Alanine aminotransferase (Field 30620) & 420{,}799 & 93.434 & 46{,}698 & 93.319 & 0.115 \\
\quad Pairs-matching completion time (Field 400) & 446{,}244 & 99.084 & 49{,}577 & 99.073 & 0.011 \\
\end{longtable}
\endgroup

Age, assessment center and ethnicity constrained the split through rejection thresholds rather than stratification. Because the first candidate seed satisfied all three criteria, these thresholds never rejected a seed, and the observed balance reflects the large cohort size rather than anything the design enforced.

\subsection{Column construction and a priori exclusions}

This section describes how we transformed the UK Biobank extract into the candidate feature matrix and which a priori exclusions we applied during preprocessing. It ends before the SHapley Additive exPlanations (SHAP)~\cite{SHAP} screening and BorutaSHAP~\cite{borutashap} stages, which belong to feature selection rather than preprocessing.

\subsubsection{From the raw extract to the feature matrix}

The extract contained 32{,}612 columns for 501{,}936 participants, whereas the final candidate feature matrix contained 7{,}977 features. The column count fell, rose above the raw-extract total, then fell again. Four construction steps account for the fall and the rise, and the a priori exclusions below for the final fall. Table~\ref{tab:column-funnel} reports the column count at each stage.

\paragraph{Baseline instance only.}
UK Biobank records repeated assessments of the same field across assessment instances (\texttt{\_i0} to \texttt{\_i4}). Instances \texttt{\_i1}--\texttt{\_i4} correspond to repeat assessment-center visits attended by only a subset of participants. We retained only the instance-0 column for each field and discarded columns from repeat visits.

\paragraph{Per-field compression and typing.}
We grouped the remaining columns by field identifier and, for fields with arrays, by array index, then collapsed each group to its first non-missing value and cast the result to the data type specified in the UK Biobank schema. We made two schema exceptions: Fields 22009 and 23376--23384 are declared numeric but contain delimited strings, so we read them as strings. This step reduced the dataset to 11{,}926 columns.

\paragraph{Early removal of dementia and diagnosis fields.}
Before the expansion we removed the dementia outcome fields and the records that define them: hospital inpatient diagnoses and their first-diagnosis dates; causes of death; and self-reported illness, medication and operation codes. That step removed 586 columns, most of them arrays of first-diagnosis dates, and left 11{,}340. 

\paragraph{One-hot expansion of multiple-choice fields.}
UK Biobank multiple-choice fields store all selected responses in a single cell, separated by vertical bars (for example, \texttt{1|3|-7}). We expanded each field into one binary indicator for every observed response code. We named each indicator by appending a double underscore and the raw response code to the parent column name; for example, response $-7$ for Field 6138 (qualifications) became \texttt{p6138\_i0\_\_-7}. For participants with a missing parent field, we set all corresponding indicators to missing rather than zero, keeping ``not asked'' distinct from ``not selected''.

The 148 multiple-choice fields became 26{,}501 indicator columns, raising the total from 11{,}340 to 37{,}693, more than in the raw extract. Much of this increase arose from a few high-cardinality hospital-record fields, particularly operative procedure codes (Fields 41272 and 41273) and hospital admission and discharge administration fields (Fields 41245--41251). We removed these parent fields, and with them every indicator they had generated, at the a priori exclusion stage that follows. Most of the 28{,}057 columns lost there were therefore indicators of those excluded parent fields, not distinct measurements. The expansion also generated indicators for the ``Do not know'' ($-1$) and ``Prefer not to answer'' ($-3$) response codes; the missing-value subsection below explains how we handled them.

\begin{table}[htbp]
  \centering
  \small
  \caption{Column count at each stage of construction and exclusion. Counts are columns, including the participant identifier.}
  \label{tab:column-funnel}
  \begin{tabular}{@{}llr@{}}
    \toprule
    Stage & Operation & Columns \\
    \midrule
    Raw extract          & ---                                                  & 32{,}612 \\
    Compression          & Baseline instance only; per-field collapse and typing & 11{,}926 \\
    Outcome removal      & Dementia- and outcome-linked fields ($-586$)          & 11{,}340 \\
    Expansion            & 148 multiple-choice fields \emph{replaced by} 26{,}501 indicators & 37{,}693 \\
    Preprocessing        & A priori exclusions ($-28{,}057$)                     &  9{,}636 \\
    Final selection      & Remaining a priori exclusions and derived features    &  7{,}978 \\
    \midrule
    \multicolumn{2}{@{}l}{Candidate features (excluding the participant identifier)} & \textbf{7{,}977} \\
    \bottomrule
  \end{tabular}
\end{table}

\subsubsection{Exclusion groups}

Features were excluded a priori, before model fitting and therefore before feature selection. Table~\ref{tab:exclusion-categories} lists the 3{,}452 excluded entries by group; each group is described below.

\begin{table}[htbp]
  \centering
  \small
  \caption{A priori feature exclusions by group. Counts are distinct excluded entries, deduplicated across pipeline stages, and are a lower bound on the number of columns removed.}
  \label{tab:exclusion-categories}
  \begin{tabular}{@{}lrl@{}}
    \toprule
    Group & Entries & Representative fields \\
    \midrule
    Outcome or temporal leakage      & 2{,}692 & 42018, 42020, 130000--132605, 41272 \\
    Diet: single-day 24-hour recall  &      8 & 20077, 26068, 105010 \\
    Study administration and timing  &    243 & 21711--21771, 20751, 32365--32368 \\
    COVID-19                         &    201 & 27981--28174, 29156--29161 \\
    Employment history               &     81 & 22601, 22602, 22603 \\
    Assay and genotyping quality control & 72 & 22000--22030, 32050--32065, 3061 \\
    Self-reported mental health      &     44 & 20544, 29000 \\
    Environmental exposure           &     43 & 24003--24024, 24500--24508, 21100--21105 \\
    Bulk-data pointers               &     37 & 21011--21018, 23145--23179 \\
    Derived accelerometry             &     16 & 40030--40045 \\
    Redundant with a retained field  &      8 & 21022, 22190, 22192, 22194 \\
    Subjective health                &      4 & 2178, 4548, 20459, 120103 \\
    Not otherwise classified         &      3 & 2345, 22501, 4144 \\
    \midrule
    Total                            & \textbf{3{,}452} & \\
    \bottomrule
  \end{tabular}
\end{table}

\paragraph{Outcome and temporal leakage.}
The largest exclusion group included the dementia and Alzheimer's disease outcome fields; the corresponding ``source of report'' fields published alongside first-occurrence and algorithmically defined outcomes; hospital inpatient diagnosis and operative-procedure records; death-registry fields recording date, age and cause of death; primary-care record-availability flags; the first-occurrence date for each recorded ICD-10 code (Fields 130000--132605); and the dates of all algorithmically defined outcomes (Fields 42000--42036). We also excluded self-reported non-cancer illness codes because they directly encode diagnoses.

\paragraph{Measurement process rather than measurement.}
We excluded variables that described the measurement process rather than participant characteristics. These included assay acquisition-route flags; sample dilution factors; genotyping batch and quality-control fields; whole-genome sequencing batch and quality metrics; and spirometry blow-acceptability codes. We excluded bulk-data fields for the same reason: they point to image or variant files rather than contain measured values. We also excluded assessment-center timestamps; fasting time before blood sampling; consent and authorization flags; and calendar artifacts such as month of birth and month of assessment.

\paragraph{Groups excluded on substantive grounds.}
We excluded COVID-19 antibody, serology and exposure-questionnaire fields because they postdate the baseline assessment and were unavailable as baseline predictors. We excluded environmental exposure fields (including air pollution; greenspace and natural-environment coverage; water hardness; and residential location history) because they derive from place of residence and may change during follow-up, making a single baseline value an unstable measure of longer-term exposure. We excluded subjective health items (overall health rating; health satisfaction; general happiness with one's health; and current health rating) because self-ratings may partly reflect undiagnosed illness.

\paragraph{Redundant with a retained field.}
Where multiple fields represented the same underlying quantity, we retained a single preferred measure. Recorded age at recruitment is available only in whole years, so we replaced it with age derived from the assessment date and the participant's month and year of birth. Among the telomere-length measures, we retained the value adjusted for technical variation and excluded the alternatives: the unadjusted measure, which retains assay-related technical variation; and the Z-standardized measure, whose standardization against the full released cohort could transmit population-level information across the development--holdout split.

\subsubsection{The candidate feature set}

After these exclusions, 7{,}977 candidate features remained. This matrix entered feature selection, and every subsequent model we developed used only features from it. Table~\ref{tab:candidate-composition} summarizes its composition.

The 7{,}977 columns came from 3{,}838 distinct UK Biobank fields. Most fields contributed a single column. A further 112 multiple-choice fields contributed 1{,}119 binary indicators, the remainder of the 26{,}501 created during preprocessing after we excluded outcome-linked parent fields. Each field appeared either as a direct column or as a set of binary indicators, never both.

We constructed four additional columns for this study instead of taking them directly from UK Biobank fields: exact age at baseline assessment, calculated from birth and assessment dates; \textit{APOE} $\varepsilon$4 allele copy number, derived from genotype; and one combined parental status-or-age variable for each parent. For the last of these, we combined parental vital status and age at death into a single variable, using a sentinel value for living parents and the age at death for deceased ones (see the ``Feature sets and related data fields'' subsection).

\begin{table}[htbp]
  \centering
  \small
  \caption{Composition of the 7{,}977-column candidate feature set that remained after the a priori exclusions and entered feature selection.}
  \label{tab:candidate-composition}
  \begin{tabular}{@{}lrr@{}}
    \toprule
    Origin & Columns & Parent fields \\
    \midrule
    UK Biobank field, one column each  & 6{,}854 & 3{,}726 \\
    Multiple-choice one-hot indicators & 1{,}119 &    112 \\
    Derived for this work              &       4 &    --- \\
    \midrule
    Total                              & \textbf{7{,}977} & \textbf{3{,}838} \\
    \bottomrule
  \end{tabular}
\end{table}

\subsubsection{Additional file 2}

Additional file 2 (\texttt{column\_status\_supplementary.xlsx}) lists the 7{,}977 candidate columns and the 3{,}452 entries excluded a priori, one per row, each labeled \texttt{retained} or \texttt{excluded}.

\subsection{Missing-value handling}

Two preprocessing steps introduced missing values before imputation. Both recoded values that UK Biobank stores as ordinary numeric entries but that do not represent substantive measurements or responses.

\subsubsection{Coded non-answers}

UK Biobank records responses such as ``Do not know'' and ``Prefer not to answer'' within each field's data coding, most commonly as $-1$ and $-3$. A gradient-boosting model treats these sentinel codes as ordinary numeric values. Unless a split isolates them explicitly, the model may group them with nearby genuine low-end values, such as ``0, No difficulty'', distorting the fitted relationship. We therefore recoded these non-response values as NaN before imputation.

We identified non-responses from the descriptive text of each UK Biobank data coding rather than from the sign of the coded value. Sign alone fails because valid responses may also use negative codes, including $-10$ for ``Less than a year'' and $-7$ for ``None of the above''; in coding 1018, all values are negative and only $-818$ denotes a non-response. This approach therefore preserved valid negative-valued responses while identifying true non-answers.

This procedure affected 1{,}336 variables and recoded 3{,}994{,}257 values as NaN. The one-hot expansion also produced 107 indicators for non-response categories. We set these indicators to NaN instead of dropping the columns, keeping non-response status out of the model. We then filled all resulting NaN values using the random-imputation procedure described in the main text.

\subsubsection{Pairs-matching completion times}

Field 400 records the time to complete each round of the pairs-matching test, where 0 means ``Test not completed'' rather than an instantaneous response~\cite{ukbiobank_field400}. Left as recorded, these zeros would signal the fastest possible completion times, reversing the meaning of a predictor that ranks among the most important features in ITU-DESC. We therefore recoded zeros as NaN in \texttt{p400\_i0\_a1} and \texttt{p400\_i0\_a2}. We then filled these values with the same random-imputation procedure used for other missing entries.

\subsection{Feature sets and related data fields}
The dietary feature ``Eats eggs, dairy, wheat, or sugar'' identifies participants who reported eating all of these food groups. It comes from Field 6144~\cite{ukbiobank_field6144}, where response option 5 reads ``I eat all of the above''. Although the parent field is titled ``Never eat eggs, dairy, wheat, sugar'', this indicator means the opposite.

The feature ``Types of physical activity in the last 4 weeks'' is derived from data coding 100316~\cite{data_coding_100316}. Response code 2 corresponds to ``Other exercises (e.g.\ swimming, cycling, keep fit, bowling)'', and this binary indicator was retained in the final feature set.

``Length of mobile phone use'' is derived from Field 1110~\cite{ukbiobank_field1110_mobile_phone_use}, which asks how many years the participant has used a mobile phone at least once per week to make or receive calls.

``Mother's vital status or age at death'' combines Field 1835, ``Mother still alive'', with Field 3526, ``Mother's age at death''. We assigned the mother's age at death to participants whose mother had died, and a sentinel value below the observed range to those whose mother was still alive. We used a sentinel rather than an imputed age because age at death is undefined for a living parent, not merely unobserved. Imputing such a value would imply a death that had not occurred. A sentinel below the observed range lets tree-based models separate living from deceased parents while preserving the ordering of observed ages at death. We built ``Father's vital status or age at death'' the same way, from Field 1797, ``Father still alive'', and Field 1807, ``Father's age at death''. This feature entered the candidate set, but only two of the ten random-imputation seeds selected it, so it stayed out of the final ITU-DESC feature set.

``Number of vehicles in household'' is derived from Field 728 and enters the model as an ordinal count.

\subsection{Software and implementation}

All analyses used Python 3.12.2 with LightGBM 4.6.0~\cite{lightgbm}, CatBoost 1.2.8~\cite{catboost}, XGBoost 2.1.4~\cite{xgb}, BorutaSHAP 1.0.17, shap 0.48.0, scikit-learn 1.5.2, pandas 2.3.2, NumPy 2.1.3, SciPy 1.15.1, statsmodels 0.14.5, lifelines 0.30.3 and Matplotlib 3.10.5.

Table~\ref{tab:model-config} reports the full training configuration of the two gradient-boosting learners: CatBoost, used for the final ITU-DESC model; and XGBoost, used for the stage-one SHAP screen and as the BorutaSHAP base model. Apart from XGBoost's native categorical support and its automatic tree-method selection, all model-forming hyperparameters used the library defaults. Everything else we set was a runtime option or the evaluation metric, and because neither learner used early stopping or a separate evaluation set, the evaluation metric affected only progress reporting.

The learning rate in Table~\ref{tab:model-config} is the value CatBoost chose for the fit on the full development set; it recalculates the rate for each cross-fitting fold. Neither model used class weighting, so the approximately 2.3\% event prevalence entered training unchanged. CatBoost's missing-value mode, \emph{minimum}, routes missing values as if they were smaller than any observed value. Random imputation removes missing values before fitting, so this mode governs only the native missing-value handling in the sensitivity analysis under ``Feature importance without random imputation'' in Supplementary Note.

\setlength{\LTcapwidth}{\linewidth}\setlength{\LTleft}{0pt}\setlength{\LTright}{0pt}
\begin{longtable}{@{\extracolsep{\fill}}p{0.14\linewidth}p{0.34\linewidth}p{0.32\linewidth}l}
\caption{Training configuration of the gradient-boosting models. \emph{Set} marks a value we passed to the constructor; \emph{supplied} marks one the library chose, and those follow from the versions listed above. Random seeds were held fixed across runs. MVS, minimal variance sampling.}
\label{tab:model-config} \\
\toprule
Library & Parameter & Value & Source \\
\midrule
\endfirsthead
\toprule
Library & Parameter & Value & Source \\
\midrule
\endhead
\midrule
\multicolumn{4}{r}{\footnotesize\textit{Continued on next page}}\\
\endfoot
\bottomrule
\endlastfoot
CatBoost & Evaluation metric & PRAUC & set \\
 & Task type & CPU & set \\
 & Threads & all available & set \\
 & Verbosity & silent, no on-disk logs & set \\
 & Random seed & fixed & set \\
 & Loss function & Logloss & supplied \\
 & Iterations & $1{,}000$ & supplied \\
 & Learning rate & $0.140$, chosen automatically & supplied \\
 & Tree depth & $6$ & supplied \\
 & L2 leaf regularization & $3$ & supplied \\
 & Grow policy & symmetric tree & supplied \\
 & Boosting type & plain & supplied \\
 & Bootstrap type & MVS, subsample $0.8$ & supplied \\
 & Feature subsampling per tree & $1$ & supplied \\
 & Split-border count & $254$ & supplied \\
 & Minimum data in leaf & $1$ & supplied \\
 & Random strength & $1$ & supplied \\
 & Leaf estimation & Newton, $10$ iterations & supplied \\
 & Score function & cosine & supplied \\
 & Missing-value mode & minimum & supplied \\
 & Class weighting & none & supplied \\
\midrule
XGBoost & Evaluation metric & aucpr & set \\
 & Tree method & \texttt{auto}, resolved to histogram & set \\
 & Categorical feature support & enabled & set \\
 & Threads & all available & set \\
 & Random seed & fixed & set \\
 & Objective & \texttt{binary:logistic} & supplied \\
 & Learning rate ($\eta$) & $0.3$ & supplied \\
 & Maximum depth & $6$ & supplied \\
 & Minimum child weight & $1$ & supplied \\
 & Row subsampling & $1$ & supplied \\
 & Column subsampling & $1$ per tree, level and node & supplied \\
 & L2 and L1 regularization & $1$ and $0$ & supplied \\
 & Minimum split loss ($\gamma$) & $0$ & supplied \\
 & Maximum delta step & $0$ & supplied \\
 & Grow policy & depthwise & supplied \\
 & Maximum bins & $256$ & supplied \\
 & Positive-class weight & $1$ & supplied \\
\end{longtable}

\section{Supplementary Note}

\subsection{Baseline characteristics and univariate associations}

Table~\ref{tab:main_baseline_characteristics} summarizes the baseline characteristics of the 38 selected predictors by incident dementia and Alzheimer's disease status. Table~\ref{tab:summary_table} extends it: for each selected predictor it adds the UK Biobank column identifier, the number of participants with a recorded value in each group and the complete response distribution of time spent driving. Table~\ref{tab:univariate_or} reports the corresponding univariate odds ratios for incident all-cause dementia and Alzheimer's disease.

\clearpage
\begin{landscape}
\begingroup
\tablebodyfont %
\setlength{\hfuzz}{2pt}
\setlength{\tabcolsep}{2pt}
\renewcommand{\arraystretch}{1.1}
\setlength{\LTcapwidth}{\linewidth}%
\newcommand{\catrow}[1]{\hangindent=1.25em\hangafter=0 #1}%
\begin{ThreePartTable}
\begin{TableNotes}[flushleft]%
\footnotesize\item[] Data are presented as median [Q1, Q3] for continuous variables and n (\%) for discrete variables. SMD = standardized mean difference. T/S ratio = telomere-to-single-copy-gene ratio, adjusted for assay-related technical parameters. The SMD (dementia) column compares participants with incident dementia with participants with no incident dementia, and the SMD (AD) column compares participants with incident Alzheimer's disease with participants with no incident dementia; both were computed on observed values. For categorical variables, parent rows report the overall effect size and indented rows show category counts and percentages. Absolute SMD values $\geq$ 0.1 were considered meaningful. Column headers give the total participants per group; the number of participants with a non-missing value varies by feature and is reported per feature in Table~\ref{tab:summary_table}; univariate odds ratios are reported in Table~\ref{tab:univariate_or}. Units are given with each variable name. UK Biobank records pairs-matching completion time and touchscreen duration in deciseconds; both are reported here in the units shown. Time spent driving is omitted here because of its many response categories and is reported in full in Table~\ref{tab:summary_table}. UK Biobank field IDs for each variable are listed in Table~\ref{tab:all_fields_presence}.
\end{TableNotes}
\begin{xltabular}{\linewidth}{X *{6}{c}}%
\caption{Baseline characteristics of selected predictors by incident dementia and Alzheimer's disease status.}\label{tab:main_baseline_characteristics}\\
\toprule
Variable & \shortstack{Overall\\(n=500{,}410)} & \shortstack{No incident\\dementia\\(n=489{,}030)} & \shortstack{Dementia\\(n=11{,}380)} & \shortstack{AD\\(n=5{,}122)} & \shortstack{SMD\\(dementia)} & \shortstack{SMD\\(AD)} \\
\midrule
\endfirsthead
\multicolumn{7}{c}{\footnotesize\textit{Table \ref{tab:main_baseline_characteristics} (continued)}}\\
\toprule
Variable & \shortstack{Overall\\(n=500{,}410)} & \shortstack{No incident\\dementia\\(n=489{,}030)} & \shortstack{Dementia\\(n=11{,}380)} & \shortstack{AD\\(n=5{,}122)} & \shortstack{SMD\\(dementia)} & \shortstack{SMD\\(AD)} \\
\midrule
\endhead
\midrule
\multicolumn{7}{r}{\footnotesize\textit{continued on next page}}\\
\endfoot
\bottomrule
\insertTableNotes
\endlastfoot
Age, y & 58.3 [50.6, 63.7] & 58.1 [50.4, 63.5] & 65.9 [62.6, 68.2] & 66.1 [63.1, 68.4] & 1.19 & 1.29 \\
Standard PRS for Alzheimer's disease (AD) & -0.09 [-0.65, 0.63] & -0.10 [-0.66, 0.61] & 0.53 [-0.25, 1.49] & 0.86 [-0.05, 1.72] & 0.56 & 0.76 \\
Long-standing illness, disability or infirmity (yes) & 159,199 (32.7\%) & 153,480 (32.2\%) & 5,719 (52.3\%) & 2,269 (46.1\%) & 0.41 & 0.29 \\
Number of treatments\slash medications taken & 2 [0, 4] & 2 [0, 4] & 4 [2, 6] & 3 [1, 6] & 0.56 & 0.47 \\
Alanine aminotransferase, U/L & 20.1 [15.4, 27.4] & 20.1 [15.4, 27.4] & 19.7 [15.4, 26.1] & 19.4 [15.2, 25.8] & -0.06 & -0.08 \\
Touchscreen duration, min & 41.8 [33.8, 52.2] & 41.6 [33.7, 52.0] & 47.7 [38.7, 60.4] & 48.5 [39.5, 61.2] & 0.41 & 0.46 \\
Aspartate aminotransferase, U/L & 24.4 [21.0, 28.8] & 24.4 [21.0, 28.8] & 25.3 [21.9, 29.7] & 25.5 [22.1, 29.9] & 0.11 & 0.11 \\
Mother still alive (yes) & 197,356 (40.2\%) & 195,950 (40.8\%) & 1,406 (12.7\%) & 549 (11.0\%) & -0.67 & -0.73 \\
Mother's age at death, y & 76 [67, 84] & 76 [67, 84] & 78 [69, 84] & 78 [69, 84] & 0.10 & 0.12 \\
\textit{APOE} \(\varepsilon\)4 &  &  &  &  & 0.58 & 0.77 \\*
\catrow{Non-carrier} & 294,093 (71.5\%) & 289,837 (72.0\%) & 4,256 (46.6\%) & 1,573 (38.1\%) &  &  \\*
\catrow{One \(\varepsilon\)4 allele} & 107,585 (26.1\%) & 103,686 (25.8\%) & 3,899 (42.7\%) & 1,984 (48.0\%) &  &  \\*
\catrow{Two \(\varepsilon\)4 alleles} & 9,771 (2.4\%) & 8,786 (2.2\%) & 985 (10.8\%) & 574 (13.9\%) &  &  \\
Basophil count, 10\textsuperscript{9} cells/L & 0.02 [0.00, 0.04] & 0.02 [0.00, 0.04] & 0.02 [0.00, 0.04] & 0.02 [0.00, 0.04] & -0.02 & -0.03 \\
Mean time to correctly identify matches, ms & 539 [481, 613] & 536 [480, 610] & 586 [520, 676] & 583 [519, 668] & 0.43 & 0.41 \\
Hand grip strength (left), kg & 28 [21, 38] & 28 [21, 38] & 26 [18, 34] & 24 [18, 34] & -0.26 & -0.30 \\
Time spent watching television (TV), h/day & 3 [2, 4] & 3 [2, 4] & 3 [2, 4] & 3 [2, 4] & 0.29 & 0.28 \\
Pairs-matching completion time, round 1, s & 8.5 [7.0, 12.6] & 8.5 [7.0, 12.5] & 11.2 [8.0, 19.0] & 11.5 [8.1, 19.2] & 0.34 & 0.36 \\
Saturated Fatty Acids to Total Fatty Acids percentage & 33.9 [32.8, 35.2] & 33.9 [32.8, 35.2] & 34.1 [32.9, 35.4] & 34.1 [32.9, 35.3] & 0.09 & 0.08 \\
Pairs-matching completion time, round 2, s & 28.0 [21.7, 36.7] & 27.9 [21.7, 36.5] & 33.6 [25.9, 44.9] & 34.1 [26.2, 45.5] & 0.41 & 0.44 \\
Other exercises (swimming, cycling, keep fit, bowling) (yes) & 235,146 (47.6\%) & 231,266 (47.9\%) & 3,880 (35.1\%) & 1,845 (36.8\%) & -0.26 & -0.23 \\
Falls in the last year &  &  &  &  & 0.25 & 0.20 \\*
\catrow{No falls} & 399,102 (80.2\%) & 391,173 (80.4\%) & 7,929 (70.4\%) & 3,656 (72.0\%) &  &  \\*
\catrow{Only one fall} & 65,699 (13.2\%) & 63,747 (13.1\%) & 1,952 (17.3\%) & 891 (17.6\%) &  &  \\*
\catrow{More than one fall} & 32,967 (6.6\%) & 31,579 (6.5\%) & 1,388 (12.3\%) & 528 (10.4\%) &  &  \\
Sleeplessness / insomnia &  &  &  &  & 0.06 & 0.01 \\*
\catrow{Never/rarely} & 120,184 (24.1\%) & 117,488 (24.1\%) & 2,696 (23.8\%) & 1,253 (24.6\%) &  &  \\*
\catrow{Sometimes} & 237,911 (47.7\%) & 232,765 (47.7\%) & 5,146 (45.5\%) & 2,414 (47.4\%) &  &  \\*
\catrow{Usually} & 140,831 (28.2\%) & 137,360 (28.2\%) & 3,471 (30.7\%) & 1,428 (28.0\%) &  &  \\
Mood swings (yes) & 221,547 (45.6\%) & 216,371 (45.5\%) & 5,176 (47.5\%) & 2,269 (46.3\%) & 0.04 & 0.01 \\
Eats eggs, dairy, wheat, sugar (yes) & 380,897 (76.8\%) & 373,697 (77.1\%) & 7,200 (64.4\%) & 3,170 (62.9\%) & -0.28 & -0.32 \\
Worry too long after embarrassment (yes) & 228,141 (47.7\%) & 223,642 (47.8\%) & 4,499 (42.3\%) & 1,958 (41.0\%) & -0.11 & -0.14 \\
IGF-1, nmol/L & 21.2 [17.5, 24.8] & 21.3 [17.5, 24.8] & 20.1 [16.2, 23.7] & 19.9 [16.2, 23.4] & -0.21 & -0.25 \\
Standard PRS for Parkinson's disease (PD) & -0.15 [-0.83, 0.54] & -0.15 [-0.83, 0.54] & -0.10 [-0.77, 0.59] & -0.11 [-0.79, 0.58] & 0.05 & 0.04 \\
Water intake, glasses/day & 2 [1, 4] & 2 [1, 4] & 2 [1, 4] & 2 [1, 4] & -0.01 & 0.03 \\
Adjusted T/S ratio & 0.82 [0.74, 0.91] & 0.82 [0.74, 0.91] & 0.79 [0.71, 0.87] & 0.79 [0.71, 0.87] & -0.27 & -0.29 \\
C-reactive protein, mg/L & 1.33 [0.66, 2.76] & 1.33 [0.66, 2.76] & 1.42 [0.69, 3.00] & 1.31 [0.65, 2.83] & 0.06 & 0.01 \\
Number of vehicles in household &  &  &  &  & 0.37 & 0.36 \\*
\catrow{None} & 43,952 (8.9\%) & 42,318 (8.7\%) & 1,634 (14.7\%) & 639 (12.7\%) &  &  \\*
\catrow{One} & 207,693 (41.9\%) & 201,893 (41.7\%) & 5,800 (52.0\%) & 2,683 (53.3\%) &  &  \\*
\catrow{Two} & 188,284 (38.0\%) & 185,116 (38.2\%) & 3,168 (28.4\%) & 1,474 (29.3\%) &  &  \\*
\catrow{Three} & 42,510 (8.6\%) & 42,082 (8.7\%) & 428 (3.8\%) & 182 (3.6\%) &  &  \\*
\catrow{Four or more} & 13,375 (2.7\%) & 13,252 (2.7\%) & 123 (1.1\%) & 52 (1.0\%) &  &  \\
Peak expiratory flow (PEF), L/min & 365 [287, 460] & 366 [288, 462] & 318 [232, 412] & 313 [231, 406] & -0.37 & -0.40 \\
Taking other prescription medications (yes) & 233,596 (47.0\%) & 225,899 (46.5\%) & 7,697 (68.4\%) & 3,316 (65.3\%) & 0.45 & 0.39 \\
Hearing difficulty\slash problems with background noise (yes) & 183,436 (37.5\%) & 178,086 (37.3\%) & 5,350 (48.5\%) & 2,322 (46.9\%) & 0.23 & 0.20 \\
Maternal history of high blood pressure (yes) & 144,074 (29.4\%) & 141,803 (29.6\%) & 2,271 (20.4\%) & 970 (19.4\%) & -0.21 & -0.24 \\
Length of mobile phone use &  &  &  &  & 0.37 & 0.39 \\*
\catrow{Never used mobile phone at least once per week} & 73,526 (14.9\%) & 70,631 (14.7\%) & 2,895 (26.4\%) & 1,311 (26.6\%) &  &  \\*
\catrow{One year or less} & 13,614 (2.8\%) & 13,108 (2.7\%) & 506 (4.6\%) & 248 (5.0\%) &  &  \\*
\catrow{Two to four years} & 86,879 (17.6\%) & 84,694 (17.6\%) & 2,185 (20.0\%) & 1,034 (21.0\%) &  &  \\*
\catrow{Five to eight years} & 151,533 (30.8\%) & 148,758 (30.9\%) & 2,775 (25.4\%) & 1,214 (24.7\%) &  &  \\*
\catrow{More than eight years} & 166,865 (33.9\%) & 164,278 (34.1\%) & 2,587 (23.6\%) & 1,115 (22.6\%) &  &  \\
Creatinine (enzymatic) in urine, \(\mu\)mol/L & 7544.0 [4377.0, 12059.0] & 7542.0 [4373.0, 12060.0] & 7662.0 [4551.0, 12027.0] & 7212.5 [4334.8, 11493.5] & 0.02 & -0.06 \\
Neutrophil percentage & 61.1 [55.6, 66.5] & 61.1 [55.6, 66.4] & 62.5 [56.9, 67.8] & 62.2 [56.6, 67.4] & 0.15 & 0.11 \\
Vitamin D, nmol/L & 46.8 [32.4, 62.4] & 46.8 [32.4, 62.4] & 46.7 [32.0, 62.5] & 47.6 [32.8, 63.8] & 0.00 & 0.04 \\
Maternal history of Alzheimer's disease\slash dementia (yes) & 38,220 (7.8\%) & 36,602 (7.6\%) & 1,618 (14.6\%) & 858 (17.1\%) & 0.22 & 0.29 \\
\end{xltabular}
\end{ThreePartTable}
\endgroup
 \end{landscape}
\clearpage
\begin{landscape}
    \addtolength{\footskip}{158pt}
\begingroup
\footnotesize
\setlength{\hfuzz}{2pt}
\setlength{\tabcolsep}{2pt}
\renewcommand{\arraystretch}{1.0}
\setlength{\LTleft}{-160pt}\setlength{\LTright}{-20pt}%
\setlength{\LTcapwidth}{\dimexpr\linewidth+180pt\relax}%
\begin{ThreePartTable}
\begin{TableNotes}[flushleft]
\footnotesize\item[] Data are presented as median [Q1, Q3] for continuous variables and n (\%) for discrete variables. For continuous variables, the number of participants with a recorded value in each group is given beneath the summary. SMD = standardized mean difference. For continuous variables, SMD was computed using means and pooled standard deviations. For binary variables, SMD was computed from group proportions. For multi-category variables, one global SMD was computed from the full category distribution. Units are given with each variable name. UK Biobank records pairs-matching completion time and touchscreen duration in deciseconds; both are reported here in the units shown. Absolute SMD values $\geq$ 0.1 were considered meaningful.
\end{TableNotes}
\begin{xltabular}{\LTcapwidth}{>{\raggedright\arraybackslash}X *{6}{c}}
\caption{Baseline characteristics of selected predictors by incident dementia and Alzheimer's disease status, including every response category and the number of participants with a recorded value.}\label{tab:summary_table}\\
\toprule
 & & & & & \multicolumn{1}{c}{\shortstack{Dementia vs\\no incident dementia}} & \multicolumn{1}{c}{\shortstack{AD vs\\no incident dementia}} \\
\cmidrule(lr){6-6}\cmidrule(lr){7-7}
Variable & Overall & \shortstack{No incident\\dementia} & Dementia & AD & SMD & SMD \\
\midrule
\endfirsthead
\multicolumn{7}{c}{\footnotesize\textit{Table \ref{tab:summary_table} (continued)}}\\
\toprule
 & & & & & \multicolumn{1}{c}{\shortstack{Dementia vs\\no incident dementia}} & \multicolumn{1}{c}{\shortstack{AD vs\\no incident dementia}} \\
\cmidrule(lr){6-6}\cmidrule(lr){7-7}
Variable & Overall & \shortstack{No incident\\dementia} & Dementia & AD & SMD & SMD \\
\midrule
\endhead
\midrule
\multicolumn{7}{r}{\footnotesize\textit{continued on next page}}\\
\endfoot
\bottomrule
\insertTableNotes
\endlastfoot
Age, y (Age) & \begin{tabular}[t]{@{}c@{}}58.3 [50.6, 63.7]\\(n=500,410)\end{tabular} & \begin{tabular}[t]{@{}c@{}}58.1 [50.4, 63.5]\\(n=489,030)\end{tabular} & \begin{tabular}[t]{@{}c@{}}65.9 [62.6, 68.2]\\(n=11,380)\end{tabular} & \begin{tabular}[t]{@{}c@{}}66.1 [63.1, 68.4]\\(n=5,122)\end{tabular} & 1.192 & 1.288 \\
Standard PRS for Alzheimer's disease (AD) (p26206) & \begin{tabular}[t]{@{}c@{}}-0.09 [-0.65, 0.63]\\(n=484,277)\end{tabular} & \begin{tabular}[t]{@{}c@{}}-0.10 [-0.66, 0.61]\\(n=473,386)\end{tabular} & \begin{tabular}[t]{@{}c@{}}0.53 [-0.25, 1.49]\\(n=10,891)\end{tabular} & \begin{tabular}[t]{@{}c@{}}0.86 [-0.05, 1.72]\\(n=4,922)\end{tabular} & 0.563 & 0.761 \\
Long-standing illness, disability or infirmity (p2188\_i0) = 0.0 & 327,933 (67.3\%) & 322,719 (67.8\%) & 5,214 (47.7\%) & 2,651 (53.9\%) & 0.415 & 0.287 \\*
Long-standing illness, disability or infirmity (p2188\_i0) = 1.0 & 159,199 (32.7\%) & 153,480 (32.2\%) & 5,719 (52.3\%) & 2,269 (46.1\%) &  &  \\
Number of treatments/medications taken (p137\_i0) & \begin{tabular}[t]{@{}c@{}}2 [0, 4]\\(n=499,572)\end{tabular} & \begin{tabular}[t]{@{}c@{}}2 [0, 4]\\(n=488,237)\end{tabular} & \begin{tabular}[t]{@{}c@{}}4 [2, 6]\\(n=11,335)\end{tabular} & \begin{tabular}[t]{@{}c@{}}3 [1, 6]\\(n=5,105)\end{tabular} & 0.561 & 0.472 \\
Alanine aminotransferase, U/L (p30620\_i0) & \begin{tabular}[t]{@{}c@{}}20.1 [15.4, 27.4]\\(n=467,497)\end{tabular} & \begin{tabular}[t]{@{}c@{}}20.1 [15.4, 27.4]\\(n=456,998)\end{tabular} & \begin{tabular}[t]{@{}c@{}}19.7 [15.4, 26.1]\\(n=10,499)\end{tabular} & \begin{tabular}[t]{@{}c@{}}19.4 [15.2, 25.8]\\(n=4,759)\end{tabular} & -0.061 & -0.081 \\
Touchscreen duration, min (p630\_i0) & \begin{tabular}[t]{@{}c@{}}41.8 [33.8, 52.2]\\(n=499,569)\end{tabular} & \begin{tabular}[t]{@{}c@{}}41.6 [33.7, 52.0]\\(n=488,221)\end{tabular} & \begin{tabular}[t]{@{}c@{}}47.7 [38.7, 60.4]\\(n=11,348)\end{tabular} & \begin{tabular}[t]{@{}c@{}}48.5 [39.5, 61.2]\\(n=5,109)\end{tabular} & 0.406 & 0.457 \\
Aspartate aminotransferase, U/L (p30650\_i0) & \begin{tabular}[t]{@{}c@{}}24.4 [21.0, 28.8]\\(n=465,900)\end{tabular} & \begin{tabular}[t]{@{}c@{}}24.4 [21.0, 28.8]\\(n=455,433)\end{tabular} & \begin{tabular}[t]{@{}c@{}}25.3 [21.9, 29.7]\\(n=10,467)\end{tabular} & \begin{tabular}[t]{@{}c@{}}25.5 [22.1, 29.9]\\(n=4,748)\end{tabular} & 0.107 & 0.113 \\
Mother still alive (p1835\_i0) = 0.0 & 293,520 (59.8\%) & 283,816 (59.2\%) & 9,704 (87.3\%) & 4,462 (89.0\%) & -0.672 & -0.726 \\*
Mother still alive (p1835\_i0) = 1.0 & 197,356 (40.2\%) & 195,950 (40.8\%) & 1,406 (12.7\%) & 549 (11.0\%) &  &  \\
Mother's age at death, y (p3526\_i0) & \begin{tabular}[t]{@{}c@{}}76 [67, 84]\\(n=291,450)\end{tabular} & \begin{tabular}[t]{@{}c@{}}76 [67, 84]\\(n=281,964)\end{tabular} & \begin{tabular}[t]{@{}c@{}}78 [69, 84]\\(n=9,486)\end{tabular} & \begin{tabular}[t]{@{}c@{}}78 [69, 84]\\(n=4,354)\end{tabular} & 0.097 & 0.119 \\
\textit{APOE} \(\varepsilon\)4 (e4\_copies) &  &  &  &  & 0.576 & 0.768 \\*
\quad Non-carrier & 294,093 (71.5\%) & 289,837 (72.0\%) & 4,256 (46.6\%) & 1,573 (38.1\%) &  &  \\*
\quad One \(\varepsilon\)4 allele & 107,585 (26.1\%) & 103,686 (25.8\%) & 3,899 (42.7\%) & 1,984 (48.0\%) &  &  \\*
\quad Two \(\varepsilon\)4 alleles & 9,771 (2.4\%) & 8,786 (2.2\%) & 985 (10.8\%) & 574 (13.9\%) &  &  \\
Basophil count, 10\textsuperscript{9} cells/L (p30160\_i0) & \begin{tabular}[t]{@{}c@{}}0.02 [0.00, 0.04]\\(n=475,347)\end{tabular} & \begin{tabular}[t]{@{}c@{}}0.02 [0.00, 0.04]\\(n=464,656)\end{tabular} & \begin{tabular}[t]{@{}c@{}}0.02 [0.00, 0.04]\\(n=10,691)\end{tabular} & \begin{tabular}[t]{@{}c@{}}0.02 [0.00, 0.04]\\(n=4,855)\end{tabular} & -0.017 & -0.030 \\
Mean time to correctly identify matches, ms (p20023\_i0) & \begin{tabular}[t]{@{}c@{}}539 [481, 613]\\(n=494,648)\end{tabular} & \begin{tabular}[t]{@{}c@{}}536 [480, 610]\\(n=483,578)\end{tabular} & \begin{tabular}[t]{@{}c@{}}586 [520, 676]\\(n=11,070)\end{tabular} & \begin{tabular}[t]{@{}c@{}}583 [519, 668]\\(n=4,997)\end{tabular} & 0.429 & 0.406 \\
Hand grip strength (left), kg (p46\_i0) & \begin{tabular}[t]{@{}c@{}}28 [21, 38]\\(n=497,050)\end{tabular} & \begin{tabular}[t]{@{}c@{}}28 [21, 38]\\(n=485,798)\end{tabular} & \begin{tabular}[t]{@{}c@{}}26 [18, 34]\\(n=11,252)\end{tabular} & \begin{tabular}[t]{@{}c@{}}24 [18, 34]\\(n=5,075)\end{tabular} & -0.257 & -0.300 \\
Time spent watching television (TV), h/day (p1070\_i0) & \begin{tabular}[t]{@{}c@{}}3 [2, 4]\\(n=495,029)\end{tabular} & \begin{tabular}[t]{@{}c@{}}3 [2, 4]\\(n=483,898)\end{tabular} & \begin{tabular}[t]{@{}c@{}}3 [2, 4]\\(n=11,131)\end{tabular} & \begin{tabular}[t]{@{}c@{}}3 [2, 4]\\(n=5,014)\end{tabular} & 0.286 & 0.278 \\
Pairs-matching completion time, round 1, s (p400\_i0\_a1) & \begin{tabular}[t]{@{}c@{}}8.5 [7.0, 12.6]\\(n=481,756)\end{tabular} & \begin{tabular}[t]{@{}c@{}}8.5 [7.0, 12.5]\\(n=471,453)\end{tabular} & \begin{tabular}[t]{@{}c@{}}11.2 [8.0, 19.0]\\(n=10,303)\end{tabular} & \begin{tabular}[t]{@{}c@{}}11.5 [8.1, 19.2]\\(n=4,647)\end{tabular} & 0.340 & 0.361 \\
Saturated Fatty Acids to Total Fatty Acids percentage (p23455\_i0) & \begin{tabular}[t]{@{}c@{}}33.9 [32.8, 35.2]\\(n=486,027)\end{tabular} & \begin{tabular}[t]{@{}c@{}}33.9 [32.8, 35.2]\\(n=475,053)\end{tabular} & \begin{tabular}[t]{@{}c@{}}34.1 [32.9, 35.4]\\(n=10,974)\end{tabular} & \begin{tabular}[t]{@{}c@{}}34.1 [32.9, 35.3]\\(n=4,948)\end{tabular} & 0.092 & 0.077 \\
Pairs-matching completion time, round 2, s (p400\_i0\_a2) & \begin{tabular}[t]{@{}c@{}}28.0 [21.7, 36.7]\\(n=480,704)\end{tabular} & \begin{tabular}[t]{@{}c@{}}27.9 [21.7, 36.5]\\(n=470,479)\end{tabular} & \begin{tabular}[t]{@{}c@{}}33.6 [25.9, 44.9]\\(n=10,225)\end{tabular} & \begin{tabular}[t]{@{}c@{}}34.1 [26.2, 45.5]\\(n=4,610)\end{tabular} & 0.406 & 0.437 \\
Other exercises (swimming, cycling, keep fit, bowling) (p6164\_i0\_\_2) = 0.0 & 259,005 (52.4\%) & 251,822 (52.1\%) & 7,183 (64.9\%) & 3,168 (63.2\%) & -0.262 & -0.225 \\*
Other exercises (swimming, cycling, keep fit, bowling) (p6164\_i0\_\_2) = 1.0 & 235,146 (47.6\%) & 231,266 (47.9\%) & 3,880 (35.1\%) & 1,845 (36.8\%) &  &  \\
Falls in the last year (p2296\_i0) &  &  &  &  & 0.249 & 0.202 \\*
\quad 1.0 & 399,102 (80.2\%) & 391,173 (80.4\%) & 7,929 (70.4\%) & 3,656 (72.0\%) &  &  \\*
\quad 2.0 & 65,699 (13.2\%) & 63,747 (13.1\%) & 1,952 (17.3\%) & 891 (17.6\%) &  &  \\*
\quad 3.0 & 32,967 (6.6\%) & 31,579 (6.5\%) & 1,388 (12.3\%) & 528 (10.4\%) &  &  \\
Sleeplessness / insomnia (p1200\_i0) &  &  &  &  & 0.057 & 0.012 \\*
\quad 1.0 & 120,184 (24.1\%) & 117,488 (24.1\%) & 2,696 (23.8\%) & 1,253 (24.6\%) &  &  \\*
\quad 2.0 & 237,911 (47.7\%) & 232,765 (47.7\%) & 5,146 (45.5\%) & 2,414 (47.4\%) &  &  \\*
\quad 3.0 & 140,831 (28.2\%) & 137,360 (28.2\%) & 3,471 (30.7\%) & 1,428 (28.0\%) &  &  \\
Mood swings (p1920\_i0) = 0.0 & 264,633 (54.4\%) & 258,901 (54.5\%) & 5,732 (52.5\%) & 2,636 (53.7\%) & 0.039 & 0.015 \\*
Mood swings (p1920\_i0) = 1.0 & 221,547 (45.6\%) & 216,371 (45.5\%) & 5,176 (47.5\%) & 2,269 (46.3\%) &  &  \\
Eats eggs, dairy, wheat, sugar (p6144\_i0\_\_5) = 0.0 & 114,880 (23.2\%) & 110,901 (22.9\%) & 3,979 (35.6\%) & 1,872 (37.1\%) & -0.282 & -0.315 \\*
Eats eggs, dairy, wheat, sugar (p6144\_i0\_\_5) = 1.0 & 380,897 (76.8\%) & 373,697 (77.1\%) & 7,200 (64.4\%) & 3,170 (62.9\%) &  &  \\
Worry too long after embarrassment (p2000\_i0) = 0.0 & 250,042 (52.3\%) & 243,900 (52.2\%) & 6,142 (57.7\%) & 2,814 (59.0\%) & -0.112 & -0.137 \\*
Worry too long after embarrassment (p2000\_i0) = 1.0 & 228,141 (47.7\%) & 223,642 (47.8\%) & 4,499 (42.3\%) & 1,958 (41.0\%) &  &  \\
IGF-1, nmol/L (p30770\_i0) & \begin{tabular}[t]{@{}c@{}}21.2 [17.5, 24.8]\\(n=465,145)\end{tabular} & \begin{tabular}[t]{@{}c@{}}21.3 [17.5, 24.8]\\(n=454,705)\end{tabular} & \begin{tabular}[t]{@{}c@{}}20.1 [16.2, 23.7]\\(n=10,440)\end{tabular} & \begin{tabular}[t]{@{}c@{}}19.9 [16.2, 23.4]\\(n=4,734)\end{tabular} & -0.210 & -0.246 \\
Time spent driving (p1090\_i0) &  &  &  &  & 0.233 & 0.203 \\*
\quad No driving & 105,589 (21.5\%) & 102,165 (21.3\%) & 3,424 (31.2\%) & 1,478 (29.8\%) &  &  \\*
\quad $<$1 hour/day & 153,079 (31.2\%) & 150,066 (31.3\%) & 3,013 (27.5\%) & 1,420 (28.6\%) &  &  \\*
\quad 1 hour/day & 139,610 (28.5\%) & 136,785 (28.5\%) & 2,825 (25.8\%) & 1,277 (25.7\%) &  &  \\*
\quad 2 hours/day & 60,698 (12.4\%) & 59,533 (12.4\%) & 1,165 (10.6\%) & 538 (10.8\%) &  &  \\*
\quad 3 hours/day & 14,637 (3.0\%) & 14,367 (3.0\%) & 270 (2.5\%) & 122 (2.5\%) &  &  \\*
\quad 4 hours/day & 6,499 (1.3\%) & 6,392 (1.3\%) & 107 (1.0\%) & 44 (0.9\%) &  &  \\*
\quad 5 hours/day & 3,137 (0.6\%) & 3,084 (0.6\%) & 53 (0.5\%) & 26 (0.5\%) &  &  \\*
\quad 6 hours/day & 2,599 (0.5\%) & 2,551 (0.5\%) & 48 (0.4\%) & 26 (0.5\%) &  &  \\*
\quad 7 hours/day & 804 (0.2\%) & 796 (0.2\%) & 8 (0.1\%) & 5 (0.1\%) &  &  \\*
\quad 8 hours/day & 1,728 (0.3\%) & 1,710 (0.4\%) & 18 (0.2\%) & 9 (0.2\%) &  &  \\*
\quad 9 hours/day & 881 (0.2\%) & 868 (0.2\%) & 13 (0.1\%) & 8 (0.2\%) &  &  \\*
\quad 10 hours/day & 793 (0.2\%) & 779 (0.2\%) & 14 (0.1\%) & 6 (0.1\%) &  &  \\*
\quad 11 hours/day & 111 (0.0\%) & 107 (0.0\%) & 4 (0.0\%) & 2 (0.0\%) &  &  \\*
\quad 12 hours/day & 249 (0.1\%) & 243 (0.1\%) & 6 (0.1\%) & 4 (0.1\%) &  &  \\*
\quad 13 hours/day & 24 (0.0\%) & 24 (0.0\%) & 0 (0.0\%) & 0 (0.0\%) &  &  \\*
\quad 14 hours/day & 34 (0.0\%) & 33 (0.0\%) & 1 (0.0\%) & 0 (0.0\%) &  &  \\*
\quad 15 hours/day & 25 (0.0\%) & 24 (0.0\%) & 1 (0.0\%) & 0 (0.0\%) &  &  \\*
\quad 16 hours/day & 6 (0.0\%) & 6 (0.0\%) & 0 (0.0\%) & 0 (0.0\%) &  &  \\*
\quad 17 hours/day & 1 (0.0\%) & 1 (0.0\%) & 0 (0.0\%) & 0 (0.0\%) &  &  \\*
\quad 18 hours/day & 4 (0.0\%) & 4 (0.0\%) & 0 (0.0\%) & 0 (0.0\%) &  &  \\*
\quad 19 hours/day & 1 (0.0\%) & 1 (0.0\%) & 0 (0.0\%) & 0 (0.0\%) &  &  \\*
\quad 20 hours/day & 15 (0.0\%) & 14 (0.0\%) & 1 (0.0\%) & 0 (0.0\%) &  &  \\*
\quad 24 hours/day & 1 (0.0\%) & 0 (0.0\%) & 1 (0.0\%) & 0 (0.0\%) &  &  \\
Standard PRS for Parkinson's disease (PD) (p26260) & \begin{tabular}[t]{@{}c@{}}-0.15 [-0.83, 0.54]\\(n=484,277)\end{tabular} & \begin{tabular}[t]{@{}c@{}}-0.15 [-0.83, 0.54]\\(n=473,386)\end{tabular} & \begin{tabular}[t]{@{}c@{}}-0.10 [-0.77, 0.59]\\(n=10,891)\end{tabular} & \begin{tabular}[t]{@{}c@{}}-0.11 [-0.79, 0.58]\\(n=4,922)\end{tabular} & 0.051 & 0.038 \\
Water intake, glasses/day (p1528\_i0) & \begin{tabular}[t]{@{}c@{}}2 [1, 4]\\(n=495,659)\end{tabular} & \begin{tabular}[t]{@{}c@{}}2 [1, 4]\\(n=484,451)\end{tabular} & \begin{tabular}[t]{@{}c@{}}2 [1, 4]\\(n=11,208)\end{tabular} & \begin{tabular}[t]{@{}c@{}}2 [1, 4]\\(n=5,044)\end{tabular} & -0.009 & 0.027 \\
Adjusted T/S ratio (p22191\_i0) & \begin{tabular}[t]{@{}c@{}}0.82 [0.74, 0.91]\\(n=470,678)\end{tabular} & \begin{tabular}[t]{@{}c@{}}0.82 [0.74, 0.91]\\(n=460,070)\end{tabular} & \begin{tabular}[t]{@{}c@{}}0.79 [0.71, 0.87]\\(n=10,608)\end{tabular} & \begin{tabular}[t]{@{}c@{}}0.79 [0.71, 0.87]\\(n=4,791)\end{tabular} & -0.270 & -0.289 \\
C-reactive protein, mg/L (p30710\_i0) & \begin{tabular}[t]{@{}c@{}}1.33 [0.66, 2.76]\\(n=466,666)\end{tabular} & \begin{tabular}[t]{@{}c@{}}1.33 [0.66, 2.76]\\(n=456,196)\end{tabular} & \begin{tabular}[t]{@{}c@{}}1.42 [0.69, 3.00]\\(n=10,470)\end{tabular} & \begin{tabular}[t]{@{}c@{}}1.31 [0.65, 2.83]\\(n=4,751)\end{tabular} & 0.059 & 0.011 \\
Number of vehicles in household (p728\_i0) &  &  &  &  & 0.371 & 0.358 \\*
\quad 1.0 & 43,952 (8.9\%) & 42,318 (8.7\%) & 1,634 (14.7\%) & 639 (12.7\%) &  &  \\*
\quad 2.0 & 207,693 (41.9\%) & 201,893 (41.7\%) & 5,800 (52.0\%) & 2,683 (53.3\%) &  &  \\*
\quad 3.0 & 188,284 (38.0\%) & 185,116 (38.2\%) & 3,168 (28.4\%) & 1,474 (29.3\%) &  &  \\*
\quad 4.0 & 42,510 (8.6\%) & 42,082 (8.7\%) & 428 (3.8\%) & 182 (3.6\%) &  &  \\*
\quad 5.0 & 13,375 (2.7\%) & 13,252 (2.7\%) & 123 (1.1\%) & 52 (1.0\%) &  &  \\
Peak expiratory flow (PEF), L/min (p3064\_i0\_a1) & \begin{tabular}[t]{@{}c@{}}365 [287, 460]\\(n=451,589)\end{tabular} & \begin{tabular}[t]{@{}c@{}}366 [288, 462]\\(n=441,842)\end{tabular} & \begin{tabular}[t]{@{}c@{}}318 [232, 412]\\(n=9,747)\end{tabular} & \begin{tabular}[t]{@{}c@{}}313 [231, 406]\\(n=4,474)\end{tabular} & -0.373 & -0.397 \\
Taking other prescription medications (p2492\_i0) = 0.0 & 263,774 (53.0\%) & 260,212 (53.5\%) & 3,562 (31.6\%) & 1,759 (34.7\%) & 0.454 & 0.387 \\*
Taking other prescription medications (p2492\_i0) = 1.0 & 233,596 (47.0\%) & 225,899 (46.5\%) & 7,697 (68.4\%) & 3,316 (65.3\%) &  &  \\
Hearing difficulty/problems with background noise (p2257\_i0) = 0.0 & 305,367 (62.5\%) & 299,698 (62.7\%) & 5,669 (51.5\%) & 2,631 (53.1\%) & 0.229 & 0.196 \\*
Hearing difficulty/problems with background noise (p2257\_i0) = 1.0 & 183,436 (37.5\%) & 178,086 (37.3\%) & 5,350 (48.5\%) & 2,322 (46.9\%) &  &  \\
Maternal history of high blood pressure (p20110\_i0\_\_8) = 0.0 & 346,801 (70.7\%) & 337,962 (70.4\%) & 8,839 (79.6\%) & 4,041 (80.6\%) & -0.212 & -0.239 \\*
Maternal history of high blood pressure (p20110\_i0\_\_8) = 1.0 & 144,074 (29.4\%) & 141,803 (29.6\%) & 2,271 (20.4\%) & 970 (19.4\%) &  &  \\
Length of mobile phone use (p1110\_i0) &  &  &  &  & 0.366 & 0.394 \\*
\quad 0.0 & 73,526 (14.9\%) & 70,631 (14.7\%) & 2,895 (26.4\%) & 1,311 (26.6\%) &  &  \\*
\quad 1.0 & 13,614 (2.8\%) & 13,108 (2.7\%) & 506 (4.6\%) & 248 (5.0\%) &  &  \\*
\quad 2.0 & 86,879 (17.6\%) & 84,694 (17.6\%) & 2,185 (20.0\%) & 1,034 (21.0\%) &  &  \\*
\quad 3.0 & 151,533 (30.8\%) & 148,758 (30.9\%) & 2,775 (25.4\%) & 1,214 (24.7\%) &  &  \\*
\quad 4.0 & 166,865 (33.9\%) & 164,278 (34.1\%) & 2,587 (23.6\%) & 1,115 (22.6\%) &  &  \\
Creatinine (enzymatic) in urine, \(\mu\)mol/L (p30510\_i0) & \begin{tabular}[t]{@{}c@{}}7544.0 [4377.0, 12059.0]\\(n=482,364)\end{tabular} & \begin{tabular}[t]{@{}c@{}}7542.0 [4373.0, 12060.0]\\(n=471,619)\end{tabular} & \begin{tabular}[t]{@{}c@{}}7662.0 [4551.0, 12027.0]\\(n=10,745)\end{tabular} & \begin{tabular}[t]{@{}c@{}}7212.5 [4334.8, 11493.5]\\(n=4,884)\end{tabular} & 0.022 & -0.055 \\
Neutrophil percentage (p30200\_i0) & \begin{tabular}[t]{@{}c@{}}61.1 [55.6, 66.5]\\(n=475,353)\end{tabular} & \begin{tabular}[t]{@{}c@{}}61.1 [55.6, 66.4]\\(n=464,662)\end{tabular} & \begin{tabular}[t]{@{}c@{}}62.5 [56.9, 67.8]\\(n=10,691)\end{tabular} & \begin{tabular}[t]{@{}c@{}}62.2 [56.6, 67.4]\\(n=4,855)\end{tabular} & 0.154 & 0.112 \\
Vitamin D, nmol/L (p30890\_i0) & \begin{tabular}[t]{@{}c@{}}46.8 [32.4, 62.4]\\(n=446,521)\end{tabular} & \begin{tabular}[t]{@{}c@{}}46.8 [32.4, 62.4]\\(n=436,559)\end{tabular} & \begin{tabular}[t]{@{}c@{}}46.7 [32.0, 62.5]\\(n=9,962)\end{tabular} & \begin{tabular}[t]{@{}c@{}}47.6 [32.8, 63.8]\\(n=4,515)\end{tabular} & -0.003 & 0.041 \\
Maternal history of Alzheimer's disease/dementia (p20110\_i0\_\_10) = 0.0 & 452,655 (92.2\%) & 443,163 (92.4\%) & 9,492 (85.4\%) & 4,153 (82.9\%) & 0.222 & 0.291 \\*
Maternal history of Alzheimer's disease/dementia (p20110\_i0\_\_10) = 1.0 & 38,220 (7.8\%) & 36,602 (7.6\%) & 1,618 (14.6\%) & 858 (17.1\%) &  &  \\
\end{xltabular}
\end{ThreePartTable}
\endgroup
 \end{landscape}
\clearpage
\begingroup
\footnotesize
\setlength{\hfuzz}{2pt}
\setlength{\tabcolsep}{2pt}
\renewcommand{\arraystretch}{1.0}
\begin{ThreePartTable}
\begin{TableNotes}[flushleft]
\footnotesize\item[] OR = odds ratio; CI = confidence interval. Odds ratios were estimated using univariate logistic regression. For continuous variables, predictors were z-standardized before modeling, so odds ratios represent the effect per 1-SD increase. For binary and ordinal-coded variables, odds ratios represent a one-unit increase in the coded variable. Odds ratios for non-ordinal multi-category variables should be interpreted cautiously. Units are given with each variable name.
\end{TableNotes}
\begin{xltabular}{\linewidth}{>{\raggedright\arraybackslash}X *{2}{c}}
\caption{Univariate associations of selected predictors with incident dementia and Alzheimer's disease.}\label{tab:univariate_or}\\
\toprule
 & \multicolumn{1}{c}{Dementia} & \multicolumn{1}{c}{AD} \\
\cmidrule(lr){2-2}\cmidrule(lr){3-3}
Variable & OR (95\% CI) & OR (95\% CI) \\
\midrule
\endfirsthead
\multicolumn{3}{c}{\footnotesize\textit{Table \ref{tab:univariate_or} (continued)}}\\
\toprule
 & \multicolumn{1}{c}{Dementia} & \multicolumn{1}{c}{AD} \\
\cmidrule(lr){2-2}\cmidrule(lr){3-3}
Variable & OR (95\% CI) & OR (95\% CI) \\
\midrule
\endhead
\midrule
\multicolumn{3}{r}{\footnotesize\textit{continued on next page}}\\
\endfoot
\bottomrule
\insertTableNotes
\endlastfoot
Age, y (Age) & 4.62 (4.47-4.77) & 5.35 (5.08-5.64) \\
Standard PRS for Alzheimer's disease (AD) (p26206) & 1.70 (1.67-1.73) & 1.98 (1.93-2.03) \\
Long-standing illness, disability or infirmity (p2188\_i0) & 2.31 (2.22-2.40) & 1.77 (1.68-1.88) \\
Number of treatments/medications taken (p137\_i0) & 1.55 (1.53-1.57) & 1.44 (1.41-1.46) \\
Alanine aminotransferase, U/L (p30620\_i0) & 0.93 (0.91-0.95) & 0.91 (0.88-0.94) \\
Touchscreen duration, min (p630\_i0) & 1.41 (1.39-1.43) & 1.41 (1.38-1.44) \\
Aspartate aminotransferase, U/L (p30650\_i0) & 1.07 (1.05-1.08) & 1.06 (1.04-1.08) \\
Mother still alive (p1835\_i0) & 0.21 (0.20-0.22) & 0.18 (0.17-0.20) \\
Mother's age at death, y (p3526\_i0) & 1.10 (1.08-1.13) & 1.13 (1.09-1.17) \\
\textit{APOE} \(\varepsilon\)4 (e4\_copies) & 2.69 (2.60-2.77) & 3.43 (3.28-3.59) \\
Basophil count, 10\textsuperscript{9} cells/L (p30160\_i0) & 0.98 (0.96-1.00) & 0.97 (0.93-1.00) \\
Mean time to correctly identify matches, ms (p20023\_i0) & 1.39 (1.37-1.41) & 1.36 (1.33-1.38) \\
Hand grip strength (left), kg (p46\_i0) & 0.77 (0.75-0.78) & 0.73 (0.71-0.76) \\
Time spent watching television (TV), h/day (p1070\_i0) & 1.46 (1.43-1.50) & 1.42 (1.37-1.47) \\
Pairs-matching completion time, round 1, s (p400\_i0\_a1) & 1.21 (1.20-1.22) & 1.19 (1.17-1.20) \\
Saturated Fatty Acids to Total Fatty Acids percentage (p23455\_i0) & 1.10 (1.08-1.12) & 1.08 (1.05-1.11) \\
Pairs-matching completion time, round 2, s (p400\_i0\_a2) & 1.32 (1.30-1.33) & 1.28 (1.26-1.30) \\
Other exercises (swimming, cycling, keep fit, bowling) (p6164\_i0\_\_2) & 0.59 (0.57-0.61) & 0.64 (0.60-0.68) \\
Falls in the last year (p2296\_i0) & 1.48 (1.44-1.52) & 1.36 (1.31-1.42) \\
Sleeplessness / insomnia (p1200\_i0) & 1.05 (1.03-1.08) & 0.99 (0.95-1.02) \\
Mood swings (p1920\_i0) & 1.08 (1.04-1.12) & 1.03 (0.97-1.09) \\
Eats eggs, dairy, wheat, sugar (p6144\_i0\_\_5) & 0.54 (0.52-0.56) & 0.51 (0.48-0.54) \\
Worry too long after embarrassment (p2000\_i0) & 0.80 (0.77-0.83) & 0.76 (0.72-0.81) \\
IGF-1, nmol/L (p30770\_i0) & 0.80 (0.78-0.82) & 0.77 (0.75-0.80) \\
Time spent driving (p1090\_i0) &  &  \\
Standard PRS for Parkinson's disease (PD) (p26260) & 1.05 (1.03-1.07) & 1.04 (1.01-1.07) \\
Water intake, glasses/day (p1528\_i0) & 0.99 (0.97-1.01) & 1.03 (1.00-1.06) \\
Adjusted T/S ratio (p22191\_i0) & 0.75 (0.73-0.76) & 0.74 (0.71-0.76) \\
C-reactive protein, mg/L (p30710\_i0) & 1.05 (1.04-1.07) & 1.01 (0.98-1.04) \\
Number of vehicles in household (p728\_i0) & 0.65 (0.63-0.66) & 0.67 (0.65-0.70) \\
Peak expiratory flow (PEF), L/min (p3064\_i0\_a1) & 0.69 (0.67-0.70) & 0.67 (0.65-0.69) \\
Taking other prescription medications (p2492\_i0) & 2.49 (2.39-2.59) & 2.14 (2.02-2.27) \\
Hearing difficulty/problems with background noise (p2257\_i0) & 1.59 (1.53-1.65) & 1.48 (1.39-1.56) \\
Maternal history of high blood pressure (p20110\_i0\_\_8) & 0.61 (0.58-0.64) & 0.57 (0.54-0.62) \\
Length of mobile phone use (p1110\_i0) & 0.78 (0.77-0.79) & 0.77 (0.76-0.78) \\
Creatinine (enzymatic) in urine, \(\mu\)mol/L (p30510\_i0) & 1.02 (1.00-1.04) & 0.94 (0.92-0.97) \\
Neutrophil percentage (p30200\_i0) & 1.17 (1.15-1.20) & 1.12 (1.09-1.15) \\
Vitamin D, nmol/L (p30890\_i0) & 1.00 (0.98-1.02) & 1.04 (1.01-1.07) \\
Maternal history of Alzheimer's disease/dementia (p20110\_i0\_\_10) & 2.06 (1.96-2.18) & 2.48 (2.30-2.67) \\
\end{xltabular}
\end{ThreePartTable}
\endgroup
 
\subsection{Uncalibrated discrimination}
\begin{figure}[htbp]
    \centering
    \includegraphics{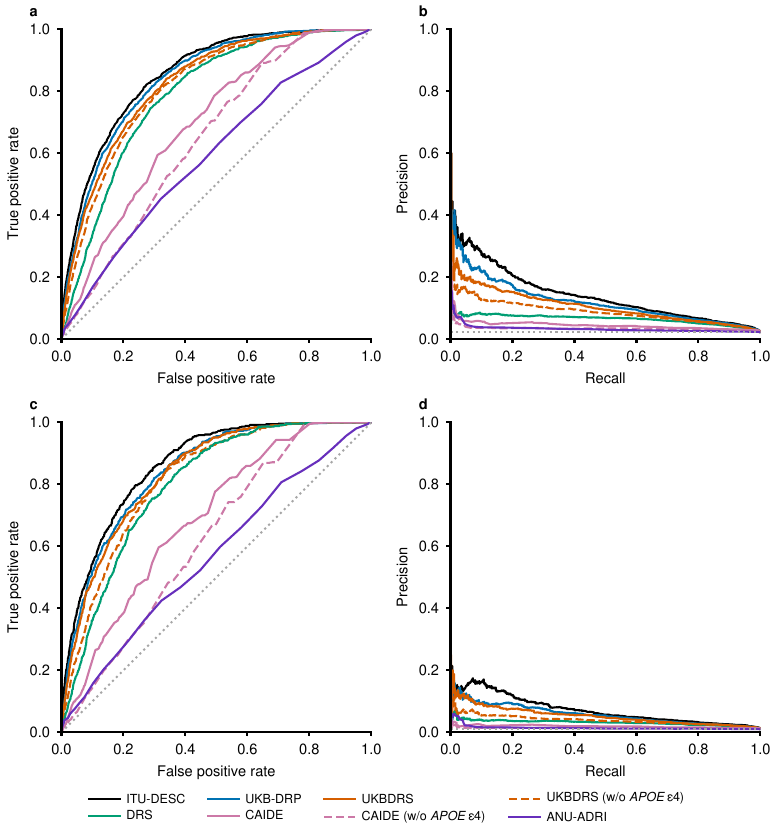}
    \caption{Uncalibrated discrimination of ITU-DESC and established dementia risk scores, computed from raw holdout-test predictions ($n = 50{,}041$) under the same random-imputation policy for every model. \textbf{a}, Receiver operating characteristic curves for all-incident all-cause dementia ($1{,}138$ incident cases). \textbf{b}, Precision--recall curves for all-incident all-cause dementia. \textbf{c}, Receiver operating characteristic curves for all-incident Alzheimer's disease ($513$ incident cases). \textbf{d}, Precision--recall curves for all-incident Alzheimer's disease. The dotted line marks chance discrimination in \textbf{a} and \textbf{c}, and the outcome prevalence in \textbf{b} and \textbf{d}. Discrimination estimates and pairwise comparisons are reported in Tables~\ref{tab:risk_scores_table_dm_uncali} and~\ref{tab:risk_scores_table_ad_uncali}.}
    \label{fig:benchmark_uncal_dm}
\end{figure}

\begin{figure}[htbp]
    \centering
    \includegraphics{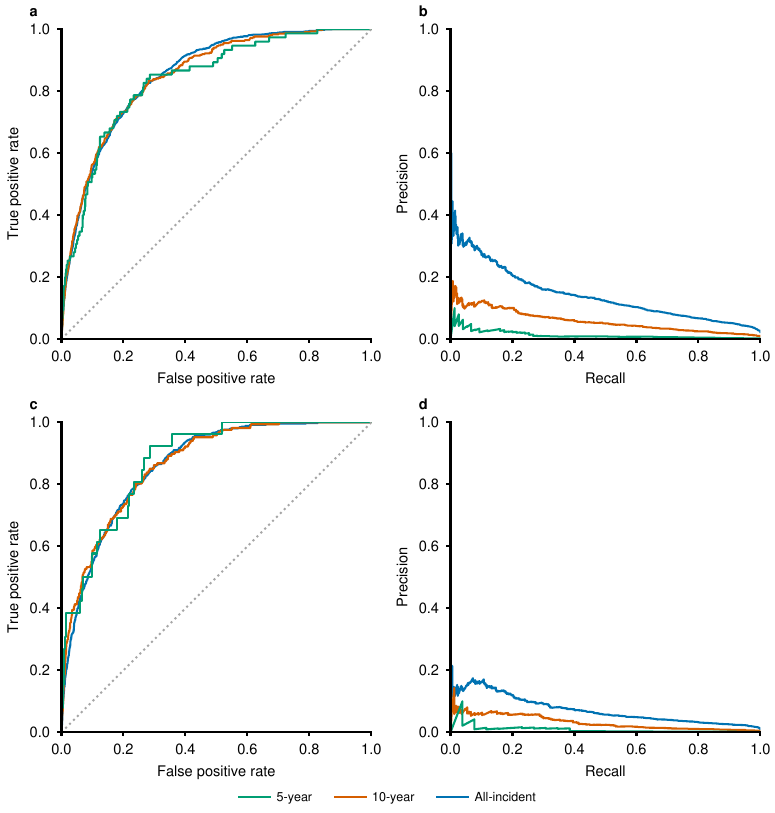}
    \caption{Uncalibrated discrimination of ITU-DESC across the 5-year, 10-year and all-incident horizons, computed from raw holdout-test predictions ($n = 50{,}041$). \textbf{a}, Receiver operating characteristic curves for all-cause dementia. \textbf{b}, Precision--recall curves for all-cause dementia. \textbf{c}, Receiver operating characteristic curves for Alzheimer's disease. \textbf{d}, Precision--recall curves for Alzheimer's disease. The same fitted model scored every participant once, so the horizons differ only in the outcome label. The dotted line marks chance discrimination in \textbf{a} and \textbf{c}. Discrimination values are reported in Table~\ref{tab:itudesc_horizons_uncali}.}
    \label{fig:temporal_uncal}
\end{figure}

We repeated both the risk-score comparison (Tables~\ref{tab:risk_scores_table_dm_uncali} and~\ref{tab:risk_scores_table_ad_uncali}; Fig.~\ref{fig:benchmark_uncal_dm}) and the analysis across time horizons (Table~\ref{tab:itudesc_horizons_uncali}; Fig.~\ref{fig:temporal_uncal}) using raw, uncalibrated holdout-test predictions rather than the isotonic-calibrated predictions reported in the main text. Model ordering held, and ITU-DESC remained the best-performing model. Uncalibrated AP was higher than the calibrated values in the main text (ITU-DESC AP $0.142$ versus $0.138$ for all-cause dementia and $0.073$ versus $0.070$ for Alzheimer's disease), consistent with isotonic ties compressing the score distribution. Table~\ref{tab:risk_scores_table_ad_cali} reports the calibrated Alzheimer's disease comparison and Table~\ref{tab:itudesc_horizons_cali} the calibrated results across horizons.

\begin{table}[htbp]
\setlength{\tabcolsep}{5pt}
\caption{Discrimination of ITU-DESC and established dementia risk scores for all-incident all-cause dementia, evaluated once on the holdout test set ($n = 50{,}041$; $1{,}138$ incident dementia cases) using raw, uncalibrated predictions. The upper panel reports average precision and the lower panel reports ROC AUC for the same models. This table repeats the main-text all-cause dementia comparison on the same models, participants and outcome definitions. Differences are ITU-DESC minus the comparator and were calculated from unrounded estimates. ROC AUC estimates, confidence intervals, differences and two-sided $P$ values were derived using DeLong's method. AP confidence intervals were obtained by paired subject-level bootstrap resampling (10{,}000 resamples; percentile method), whereas AP $P$ values were obtained using two-sided paired permutation tests (2{,}000 permutations). AP, average precision; CI, confidence interval; ROC AUC, area under the receiver operating characteristic curve.}
\label{tab:risk_scores_table_dm_uncali}
\centering
\begin{tabular}{lccc}
\toprule
\textbf{Model} & \textbf{AP} (95\% CI) & \textbf{$\Delta$AP} (95\% CI) & \textbf{$P$} \\
\midrule
ITU-DESC (our model) & 0.142 (0.127, 0.161) & --- & --- \\
UKB-DRP~\cite{YOU2022101665} & 0.122 (0.109, 0.138) & 0.021 (0.007, 0.034) & $<0.05$ \\
UKBDRS~\cite{Anaturk2023} & 0.106 (0.095, 0.120) & 0.036 (0.022, 0.051) & $<0.001$ \\
UKBDRS (w/o \textit{APOE} $\varepsilon$4) & 0.088 (0.079, 0.099) & 0.055 (0.040, 0.071) & $<0.001$ \\
DRS~\cite{DRS} & 0.065 (0.059, 0.071) & 0.078 (0.064, 0.095) & $<0.001$ \\
CAIDE~\cite{CAIDE} & 0.045 (0.041, 0.051) & 0.098 (0.083, 0.115) & $<0.001$ \\
CAIDE (w/o \textit{APOE} $\varepsilon$4) & 0.035 (0.032, 0.039) & 0.108 (0.093, 0.125) & $<0.001$ \\
ANU-ADRI~\cite{ANU_ADRI} & 0.033 (0.030, 0.038) & 0.110 (0.094, 0.128) & $<0.001$ \\
\addlinespace
\midrule
\textbf{Model} & \textbf{ROC AUC} (95\% CI) & \textbf{$\Delta$ROC AUC} (95\% CI) & \textbf{$P$} \\
\midrule
ITU-DESC (our model) & 0.854 (0.844, 0.863) & --- & --- \\
UKB-DRP~\cite{YOU2022101665} & 0.841 (0.831, 0.851) & 0.013 (0.007, 0.019) & $<0.001$ \\
UKBDRS~\cite{Anaturk2023} & 0.826 (0.815, 0.837) & 0.028 (0.021, 0.035) & $<0.001$ \\
UKBDRS (w/o \textit{APOE} $\varepsilon$4) & 0.813 (0.802, 0.824) & 0.041 (0.033, 0.049) & $<0.001$ \\
DRS~\cite{DRS} & 0.788 (0.777, 0.800) & 0.065 (0.056, 0.074) & $<0.001$ \\
CAIDE~\cite{CAIDE} & 0.697 (0.684, 0.710) & 0.157 (0.143, 0.170) & $<0.001$ \\
CAIDE (w/o \textit{APOE} $\varepsilon$4) & 0.648 (0.634, 0.661) & 0.206 (0.191, 0.221) & $<0.001$ \\
ANU-ADRI~\cite{ANU_ADRI} & 0.591 (0.575, 0.608) & 0.262 (0.244, 0.281) & $<0.001$ \\
\bottomrule
\end{tabular}
\end{table}
\begin{table}[htbp]
\setlength{\tabcolsep}{5pt}
\caption{Discrimination of ITU-DESC and established dementia risk scores for all-incident Alzheimer's disease, evaluated once on the holdout test set ($n = 50{,}041$; $513$ incident Alzheimer's disease cases) using raw, uncalibrated predictions. The upper panel reports average precision and the lower panel reports ROC AUC for the same models. This table repeats the Alzheimer's disease comparison of Table~\ref{tab:risk_scores_table_ad_cali} on the same models, participants and outcome definitions. Differences are ITU-DESC minus the comparator and were calculated from unrounded estimates. ROC AUC estimates, confidence intervals, differences and two-sided $P$ values were derived using DeLong's method. AP confidence intervals were obtained by paired subject-level bootstrap resampling (10{,}000 resamples; percentile method), whereas AP $P$ values were obtained using two-sided paired permutation tests (2{,}000 permutations). AP, average precision; CI, confidence interval; ROC AUC, area under the receiver operating characteristic curve.}
\label{tab:risk_scores_table_ad_uncali}
\centering
\begin{tabular}{lccc}
\toprule
\textbf{Model} & \textbf{AP} (95\% CI) & \textbf{$\Delta$AP} (95\% CI) & \textbf{$P$} \\
\midrule
ITU-DESC (our model) & 0.073 (0.061, 0.089) & --- & --- \\
UKB-DRP~\cite{YOU2022101665} & 0.058 (0.049, 0.070) & 0.015 (0.005, 0.027) & $<0.05$ \\
UKBDRS~\cite{Anaturk2023} & 0.054 (0.045, 0.066) & 0.019 (0.007, 0.032) & $<0.01$ \\
UKBDRS (w/o \textit{APOE} $\varepsilon$4) & 0.039 (0.034, 0.048) & 0.033 (0.022, 0.048) & $<0.001$ \\
DRS~\cite{DRS} & 0.032 (0.028, 0.037) & 0.041 (0.030, 0.056) & $<0.001$ \\
CAIDE~\cite{CAIDE} & 0.019 (0.017, 0.023) & 0.053 (0.042, 0.069) & $<0.001$ \\
CAIDE (w/o \textit{APOE} $\varepsilon$4) & 0.015 (0.013, 0.018) & 0.058 (0.047, 0.074) & $<0.001$ \\
ANU-ADRI~\cite{ANU_ADRI} & 0.014 (0.012, 0.018) & 0.058 (0.047, 0.074) & $<0.001$ \\
\addlinespace
\midrule
\textbf{Model} & \textbf{ROC AUC} (95\% CI) & \textbf{$\Delta$ROC AUC} (95\% CI) & \textbf{$P$} \\
\midrule
ITU-DESC (our model) & 0.863 (0.850, 0.876) & --- & --- \\
UKB-DRP~\cite{YOU2022101665} & 0.844 (0.830, 0.858) & 0.019 (0.010, 0.028) & $<0.001$ \\
UKBDRS~\cite{Anaturk2023} & 0.837 (0.823, 0.851) & 0.026 (0.016, 0.036) & $<0.001$ \\
UKBDRS (w/o \textit{APOE} $\varepsilon$4) & 0.817 (0.803, 0.832) & 0.046 (0.035, 0.057) & $<0.001$ \\
DRS~\cite{DRS} & 0.799 (0.784, 0.814) & 0.065 (0.052, 0.077) & $<0.001$ \\
CAIDE~\cite{CAIDE} & 0.691 (0.671, 0.710) & 0.173 (0.152, 0.193) & $<0.001$ \\
CAIDE (w/o \textit{APOE} $\varepsilon$4) & 0.625 (0.605, 0.645) & 0.238 (0.216, 0.261) & $<0.001$ \\
ANU-ADRI~\cite{ANU_ADRI} & 0.568 (0.543, 0.593) & 0.295 (0.268, 0.322) & $<0.001$ \\
\bottomrule
\end{tabular}
\end{table}
\begin{table}[htbp]
\setlength{\tabcolsep}{5pt}
\caption{Discrimination of ITU-DESC and established dementia risk scores for all-incident Alzheimer's disease, evaluated once on the holdout test set ($n = 50{,}041$; $513$ incident Alzheimer's disease cases). All models were evaluated on the same participants, under identical outcome definitions, using isotonic-calibrated predictions. The upper panel reports average precision and the lower panel reports ROC AUC for the same models. The base model was fitted once using the all-cause dementia label and recalibrated for Alzheimer's disease; every comparator was recalibrated under the same procedure. Differences are ITU-DESC minus the comparator and were calculated from unrounded estimates. ROC AUC estimates, confidence intervals, differences and two-sided $P$ values were derived using DeLong's method. AP confidence intervals were obtained by paired subject-level bootstrap resampling (10{,}000 resamples; percentile method), whereas AP $P$ values were obtained using two-sided paired permutation tests (2{,}000 permutations). AP, average precision; CI, confidence interval; ROC AUC, area under the receiver operating characteristic curve.}
\label{tab:risk_scores_table_ad_cali}
\centering
\begin{tabular}{lccc}
\toprule
\textbf{Model} & \textbf{AP} (95\% CI) & \textbf{$\Delta$AP} (95\% CI) & \textbf{$P$} \\
\midrule
ITU-DESC (our model) & 0.070 (0.059, 0.085) & --- & --- \\
UKB-DRP~\cite{YOU2022101665} & 0.056 (0.047, 0.067) & 0.014 (0.005, 0.025) & $<0.01$ \\
UKBDRS~\cite{Anaturk2023} & 0.051 (0.043, 0.062) & 0.019 (0.009, 0.030) & $<0.01$ \\
UKBDRS (w/o \textit{APOE} $\varepsilon$4) & 0.037 (0.032, 0.044) & 0.033 (0.023, 0.045) & $<0.001$ \\
DRS~\cite{DRS} & 0.030 (0.027, 0.035) & 0.040 (0.030, 0.053) & $<0.001$ \\
CAIDE~\cite{CAIDE} & 0.019 (0.017, 0.022) & 0.051 (0.041, 0.064) & $<0.001$ \\
CAIDE (w/o \textit{APOE} $\varepsilon$4) & 0.014 (0.013, 0.016) & 0.056 (0.045, 0.070) & $<0.001$ \\
ANU-ADRI~\cite{ANU_ADRI} & 0.013 (0.012, 0.016) & 0.057 (0.046, 0.071) & $<0.001$ \\
\addlinespace
\midrule
\textbf{Model} & \textbf{ROC AUC} (95\% CI) & \textbf{$\Delta$ROC AUC} (95\% CI) & \textbf{$P$} \\
\midrule
ITU-DESC (our model) & 0.863 (0.850, 0.876) & --- & --- \\
UKB-DRP~\cite{YOU2022101665} & 0.844 (0.829, 0.858) & 0.019 (0.010, 0.028) & $<0.001$ \\
UKBDRS~\cite{Anaturk2023} & 0.836 (0.822, 0.851) & 0.027 (0.016, 0.037) & $<0.001$ \\
UKBDRS (w/o \textit{APOE} $\varepsilon$4) & 0.817 (0.802, 0.831) & 0.046 (0.035, 0.058) & $<0.001$ \\
DRS~\cite{DRS} & 0.799 (0.784, 0.813) & 0.064 (0.052, 0.077) & $<0.001$ \\
CAIDE~\cite{CAIDE} & 0.701 (0.682, 0.720) & 0.162 (0.142, 0.181) & $<0.001$ \\
CAIDE (w/o \textit{APOE} $\varepsilon$4) & 0.632 (0.614, 0.650) & 0.231 (0.210, 0.252) & $<0.001$ \\
ANU-ADRI~\cite{ANU_ADRI} & 0.566 (0.542, 0.591) & 0.296 (0.270, 0.323) & $<0.001$ \\
\bottomrule
\end{tabular}
\end{table}
 
\begin{table}[htbp]
\centering
\caption{Discrimination of ITU-DESC across the 5-year, 10-year and all-incident horizons for all-cause dementia and Alzheimer's disease, evaluated once on the holdout test set ($n = 50{,}041$) using isotonic-calibrated predictions. The main-text figure plots these same point estimates. The base model, fitted once, assigned every participant the same raw score at each horizon; the horizons differ only in the outcome label and in the isotonic mapping fitted for each. AP rises with the number of cases accrued by each horizon, because the AP of a non-informative model equals the outcome prevalence. AP, average precision; ROC AUC, area under the receiver operating characteristic curve.}
\label{tab:itudesc_horizons_cali}
\begin{tabular*}{\linewidth}{@{\extracolsep{\fill}}lcccc}
\toprule
 & \multicolumn{2}{c}{All-cause dementia} & \multicolumn{2}{c}{Alzheimer's disease} \\
\cmidrule(lr){2-3} \cmidrule(lr){4-5}
Horizon & ROC AUC & AP & ROC AUC & AP \\
\midrule
5-year & 0.837 & 0.012 & 0.868 & 0.007 \\
10-year & 0.850 & 0.056 & 0.866 & 0.030 \\
All-incident & 0.853 & 0.138 & 0.863 & 0.070 \\
\bottomrule
\end{tabular*}
\end{table}
\begin{table}[htbp]
\centering
\caption{Discrimination of ITU-DESC across the 5-year, 10-year and all-incident horizons for all-cause dementia and Alzheimer's disease, evaluated once on the holdout test set ($n = 50{,}041$) using raw, uncalibrated predictions. These same point estimates are plotted in Fig.~\ref{fig:temporal_uncal}. The base model, fitted once, assigned every participant the same raw score at each horizon, so the horizons differ only in the outcome label. AP rises with the number of cases accrued by each horizon, because the AP of a non-informative model equals the outcome prevalence. AP, average precision; ROC AUC, area under the receiver operating characteristic curve.}
\label{tab:itudesc_horizons_uncali}
\begin{tabular*}{\linewidth}{@{\extracolsep{\fill}}lcccc}
\toprule
 & \multicolumn{2}{c}{All-cause dementia} & \multicolumn{2}{c}{Alzheimer's disease} \\
\cmidrule(lr){2-3} \cmidrule(lr){4-5}
Horizon & ROC AUC & AP & ROC AUC & AP \\
\midrule
5-year & 0.839 & 0.014 & 0.876 & 0.011 \\
10-year & 0.850 & 0.058 & 0.866 & 0.033 \\
All-incident & 0.854 & 0.142 & 0.863 & 0.073 \\
\bottomrule
\end{tabular*}
\end{table}

\subsection{Per-seed feature presence}
Because random imputation makes both stages of feature selection stochastic, we repeated the complete pipeline (the five-fold SHAP screen followed by BorutaSHAP) independently for each seed. Each seed therefore produced its own screened feature set and final selection. Individual lists held 40 to 59 features, and their union contained 120 distinct features (Set 1, the union); only 29 features appeared in all ten seeds (Set 10, the intersection). A further five features appeared in nine seeds and four in eight, giving the 38 features of Set 8, which we adopted as the final ITU-DESC feature set. This approximately 20-feature spread shows how much a single wrapper run depends on the imputation draw and justifies the consensus cutoff. For each candidate feature, Table~\ref{tab:all_fields_presence} reports the seeds that selected it (columns S1--S10) and the total number of selections, which defines membership in the nested consensus sets.
Feature names encode the UK Biobank field identifier with the assessment instance (\texttt{i}) and array (\texttt{a}) indices, both zero-based. Some features therefore share a field label but differ in their indices. For example, \texttt{p400\_i0\_a1} and \texttt{p400\_i0\_a2} denote Field 400, instance 0, arrays 1 and 2, which record the completion times for the two rounds of the pairs-matching test. A double-underscore suffix (\texttt{\_\_}) denotes a binary indicator for a specific response code, so features from the same field can encode different responses: \texttt{p20110\_i0\_\_8} indicates maternal high blood pressure and \texttt{p20110\_i0\_\_10} maternal Alzheimer's disease or dementia, both from Field 20110, ``Illnesses of mother''. Similarly, \texttt{p6164\_i0\_\_2} indicates response code 2 for Field 6164, ``Other exercises (eg: swimming, cycling, keep fit, bowling)''. These multiple-choice fields store selected response codes as pipe-delimited strings (for example, ``5|4|3''), and we expanded them into separate binary indicators for each observed code.

\begin{footnotesize}
\setlength{\tabcolsep}{3pt}
\renewcommand{\arraystretch}{1.1}
\setlength{\LTcapwidth}{\linewidth}\setlength{\LTleft}{0pt}\setlength{\LTright}{0pt}
\begin{longtable}{@{\extracolsep{\fill}}>{\raggedright\arraybackslash}p{2.3cm}>{\raggedright\arraybackslash}p{5.0cm}ccccccccccc}
\caption{Per-seed presence of each candidate feature across the ten random-imputation seeds. A check mark indicates the feature was selected by BorutaSHAP in that seed's run; \textit{Seeds} gives the total number of seeds (out of ten) in which the feature was selected. Features are ordered by the number of seeds in which they appear. The 38 features selected in at least eight seeds constitute Set~8, adopted as the final ITU-DESC feature set.}
\label{tab:all_fields_presence}\\
\toprule
Field & Feature & S1 & S2 & S3 & S4 & S5 & S6 & S7 & S8 & S9 & S10 & Seeds \\
\midrule
\endfirsthead
\multicolumn{13}{l}{\textit{Table~\ref{tab:all_fields_presence} continued from previous page}}\\
\toprule
Field & Feature & S1 & S2 & S3 & S4 & S5 & S6 & S7 & S8 & S9 & S10 & Seeds \\
\midrule
\endhead
\midrule
\multicolumn{13}{r}{\textit{Continued on next page}}\\
\endfoot
\bottomrule
\endlastfoot
Age & Age & \checkmark & \checkmark & \checkmark & \checkmark & \checkmark & \checkmark & \checkmark & \checkmark & \checkmark & \checkmark & 10 \\
Maternal\_\allowbreak status\_\allowbreak or\_\allowbreak age & Mother's vital status or age at death & \checkmark & \checkmark & \checkmark & \checkmark & \checkmark & \checkmark & \checkmark & \checkmark & \checkmark & \checkmark & 10 \\
e4\_\allowbreak copies & \textit{APOE} $\varepsilon$4 copies & \checkmark & \checkmark & \checkmark & \checkmark & \checkmark & \checkmark & \checkmark & \checkmark & \checkmark & \checkmark & 10 \\
p1070\_\allowbreak i0 & Time spent watching television (TV) & \checkmark & \checkmark & \checkmark & \checkmark & \checkmark & \checkmark & \checkmark & \checkmark & \checkmark & \checkmark & 10 \\
p1090\_\allowbreak i0 & Time spent driving & \checkmark & \checkmark & \checkmark & \checkmark & \checkmark & \checkmark & \checkmark & \checkmark & \checkmark & \checkmark & 10 \\
p1200\_\allowbreak i0 & Sleeplessness / insomnia & \checkmark & \checkmark & \checkmark & \checkmark & \checkmark & \checkmark & \checkmark & \checkmark & \checkmark & \checkmark & 10 \\
p137\_\allowbreak i0 & Number of treatments/medications taken & \checkmark & \checkmark & \checkmark & \checkmark & \checkmark & \checkmark & \checkmark & \checkmark & \checkmark & \checkmark & 10 \\
p1528\_\allowbreak i0 & Water intake & \checkmark & \checkmark & \checkmark & \checkmark & \checkmark & \checkmark & \checkmark & \checkmark & \checkmark & \checkmark & 10 \\
p1920\_\allowbreak i0 & Mood swings & \checkmark & \checkmark & \checkmark & \checkmark & \checkmark & \checkmark & \checkmark & \checkmark & \checkmark & \checkmark & 10 \\
p2000\_\allowbreak i0 & Worry too long after embarrassment & \checkmark & \checkmark & \checkmark & \checkmark & \checkmark & \checkmark & \checkmark & \checkmark & \checkmark & \checkmark & 10 \\
p20023\_\allowbreak i0 & Mean time to correctly identify matches & \checkmark & \checkmark & \checkmark & \checkmark & \checkmark & \checkmark & \checkmark & \checkmark & \checkmark & \checkmark & 10 \\
p2188\_\allowbreak i0 & Long-standing illness, disability or infirmity & \checkmark & \checkmark & \checkmark & \checkmark & \checkmark & \checkmark & \checkmark & \checkmark & \checkmark & \checkmark & 10 \\
p22191\_\allowbreak i0 & Adjusted T/S ratio & \checkmark & \checkmark & \checkmark & \checkmark & \checkmark & \checkmark & \checkmark & \checkmark & \checkmark & \checkmark & 10 \\
p2296\_\allowbreak i0 & Falls in the last year & \checkmark & \checkmark & \checkmark & \checkmark & \checkmark & \checkmark & \checkmark & \checkmark & \checkmark & \checkmark & 10 \\
p23455\_\allowbreak i0 & Saturated Fatty Acids to Total Fatty Acids percentage & \checkmark & \checkmark & \checkmark & \checkmark & \checkmark & \checkmark & \checkmark & \checkmark & \checkmark & \checkmark & 10 \\
p26206 & Standard PRS for Alzheimer's disease (AD) & \checkmark & \checkmark & \checkmark & \checkmark & \checkmark & \checkmark & \checkmark & \checkmark & \checkmark & \checkmark & 10 \\
p26260 & Standard PRS for Parkinson's disease (PD) & \checkmark & \checkmark & \checkmark & \checkmark & \checkmark & \checkmark & \checkmark & \checkmark & \checkmark & \checkmark & 10 \\
p30160\_\allowbreak i0 & Basophil count & \checkmark & \checkmark & \checkmark & \checkmark & \checkmark & \checkmark & \checkmark & \checkmark & \checkmark & \checkmark & 10 \\
p30620\_\allowbreak i0 & Alanine aminotransferase & \checkmark & \checkmark & \checkmark & \checkmark & \checkmark & \checkmark & \checkmark & \checkmark & \checkmark & \checkmark & 10 \\
p30650\_\allowbreak i0 & Aspartate aminotransferase & \checkmark & \checkmark & \checkmark & \checkmark & \checkmark & \checkmark & \checkmark & \checkmark & \checkmark & \checkmark & 10 \\
p30710\_\allowbreak i0 & C-reactive protein & \checkmark & \checkmark & \checkmark & \checkmark & \checkmark & \checkmark & \checkmark & \checkmark & \checkmark & \checkmark & 10 \\
p30770\_\allowbreak i0 & IGF-1 & \checkmark & \checkmark & \checkmark & \checkmark & \checkmark & \checkmark & \checkmark & \checkmark & \checkmark & \checkmark & 10 \\
p400\_\allowbreak i0\_\allowbreak a1 & Pairs-matching completion time, round 1 & \checkmark & \checkmark & \checkmark & \checkmark & \checkmark & \checkmark & \checkmark & \checkmark & \checkmark & \checkmark & 10 \\
p400\_\allowbreak i0\_\allowbreak a2 & Pairs-matching completion time, round 2 & \checkmark & \checkmark & \checkmark & \checkmark & \checkmark & \checkmark & \checkmark & \checkmark & \checkmark & \checkmark & 10 \\
p46\_\allowbreak i0 & Hand grip strength (left) & \checkmark & \checkmark & \checkmark & \checkmark & \checkmark & \checkmark & \checkmark & \checkmark & \checkmark & \checkmark & 10 \\
p6144\_\allowbreak i0\_\allowbreak \_\allowbreak 5 & Eats eggs, dairy, wheat, sugar & \checkmark & \checkmark & \checkmark & \checkmark & \checkmark & \checkmark & \checkmark & \checkmark & \checkmark & \checkmark & 10 \\
p6164\_\allowbreak i0\_\allowbreak \_\allowbreak 2 & Other exercises (swimming, cycling, keep fit, bowling) & \checkmark & \checkmark & \checkmark & \checkmark & \checkmark & \checkmark & \checkmark & \checkmark & \checkmark & \checkmark & 10 \\
p630\_\allowbreak i0 & Touchscreen duration & \checkmark & \checkmark & \checkmark & \checkmark & \checkmark & \checkmark & \checkmark & \checkmark & \checkmark & \checkmark & 10 \\
p728\_\allowbreak i0 & Number of vehicles in household & \checkmark & \checkmark & \checkmark & \checkmark & \checkmark & \checkmark & \checkmark & \checkmark & \checkmark & \checkmark & 10 \\
p1110\_\allowbreak i0 & Length of mobile phone use & \checkmark & \checkmark & \checkmark & \checkmark & \checkmark & \checkmark &  & \checkmark & \checkmark & \checkmark & 9 \\
p20110\_\allowbreak i0\_\allowbreak \_\allowbreak 8 & Maternal history of high blood pressure & \checkmark & \checkmark & \checkmark & \checkmark & \checkmark & \checkmark & \checkmark & \checkmark & \checkmark &  & 9 \\
p2257\_\allowbreak i0 & Hearing difficulty/problems with background noise &  & \checkmark & \checkmark & \checkmark & \checkmark & \checkmark & \checkmark & \checkmark & \checkmark & \checkmark & 9 \\
p2492\_\allowbreak i0 & Taking other prescription medications & \checkmark & \checkmark & \checkmark &  & \checkmark & \checkmark & \checkmark & \checkmark & \checkmark & \checkmark & 9 \\
p3064\_\allowbreak i0\_\allowbreak a1 & Peak expiratory flow (PEF) & \checkmark & \checkmark & \checkmark & \checkmark & \checkmark &  & \checkmark & \checkmark & \checkmark & \checkmark & 9 \\
p20110\_\allowbreak i0\_\allowbreak \_\allowbreak 10 & Maternal history of Alzheimer's disease/dementia & \checkmark & \checkmark & \checkmark & \checkmark & \checkmark &  & \checkmark &  & \checkmark & \checkmark & 8 \\
p30200\_\allowbreak i0 & Neutrophil percentage & \checkmark & \checkmark & \checkmark & \checkmark & \checkmark &  &  & \checkmark & \checkmark & \checkmark & 8 \\
p30510\_\allowbreak i0 & Creatinine (enzymatic) in urine & \checkmark &  & \checkmark & \checkmark & \checkmark & \checkmark & \checkmark & \checkmark &  & \checkmark & 8 \\
p30890\_\allowbreak i0 & Vitamin D & \checkmark & \checkmark & \checkmark & \checkmark & \checkmark & \checkmark & \checkmark &  &  & \checkmark & 8 \\
p1488\_\allowbreak i0 & Tea intake &  & \checkmark & \checkmark & \checkmark & \checkmark &  & \checkmark &  & \checkmark & \checkmark & 7 \\
p23111\_\allowbreak i0 & Leg fat percentage (right) & \checkmark &  &  & \checkmark & \checkmark & \checkmark & \checkmark & \checkmark &  & \checkmark & 7 \\
p23115\_\allowbreak i0 & Leg fat percentage (left) &  & \checkmark & \checkmark &  & \checkmark & \checkmark &  & \checkmark & \checkmark & \checkmark & 7 \\
p23470\_\allowbreak i0 & Glucose & \checkmark & \checkmark &  & \checkmark & \checkmark &  & \checkmark &  & \checkmark & \checkmark & 7 \\
p30670\_\allowbreak i0 & Urea & \checkmark &  &  & \checkmark & \checkmark &  & \checkmark & \checkmark & \checkmark & \checkmark & 7 \\
p23444\_\allowbreak i0 & Omega-3 Fatty Acids & \checkmark & \checkmark &  &  & \checkmark & \checkmark & \checkmark & \checkmark &  &  & 6 \\
p3077\_\allowbreak i0 & Seating box height & \checkmark & \checkmark & \checkmark &  & \checkmark &  & \checkmark &  &  &  & 5 \\
p738\_\allowbreak i0 & Average total household income before tax & \checkmark &  &  & \checkmark & \checkmark &  & \checkmark &  &  & \checkmark & 5 \\
p22009\_\allowbreak a2 & Genetic principal components &  &  & \checkmark & \checkmark & \checkmark &  &  &  & \checkmark &  & 4 \\
p23107\_\allowbreak i0 & Impedance of leg (right) & \checkmark &  & \checkmark &  & \checkmark &  &  &  & \checkmark &  & 4 \\
p23112\_\allowbreak i0 & Leg fat mass (right) & \checkmark &  & \checkmark & \checkmark & \checkmark &  &  &  &  &  & 4 \\
p1060\_\allowbreak i0 & Time spent outdoors in winter &  &  &  & \checkmark &  &  &  & \checkmark &  & \checkmark & 3 \\
p22009\_\allowbreak a11 & Genetic principal components &  & \checkmark & \checkmark & \checkmark &  &  &  &  &  &  & 3 \\
p28627 & Currently suffering from hearing loss &  &  &  & \checkmark & \checkmark &  &  & \checkmark &  &  & 3 \\
p28739 & Difficulty with standing for long periods & \checkmark &  & \checkmark &  &  &  &  &  &  & \checkmark & 3 \\
p3064\_\allowbreak i0\_\allowbreak a2 & Peak expiratory flow (PEF) &  &  & \checkmark &  & \checkmark &  & \checkmark &  &  &  & 3 \\
p47\_\allowbreak i0 & Hand grip strength (right) &  &  &  &  & \checkmark &  & \checkmark &  & \checkmark &  & 3 \\
p6164\_\allowbreak i0\_\allowbreak \_\allowbreak 5 & Types of physical activity: prefer not to answer &  &  & \checkmark & \checkmark &  &  &  &  &  & \checkmark & 3 \\
Paternal\_\allowbreak status\_\allowbreak or\_\allowbreak age & Father's vital status or age at death &  &  &  &  &  &  &  &  & \checkmark & \checkmark & 2 \\
p20162\_\allowbreak i0 & Pack years adult smoking as proportion of life span exposed to smoking &  &  &  & \checkmark &  &  &  &  &  & \checkmark & 2 \\
p2090\_\allowbreak i0 & Seen doctor (GP) for nerves, anxiety, tension or depression & \checkmark &  &  &  &  &  &  &  & \checkmark &  & 2 \\
p22009\_\allowbreak a1 & Genetic principal components &  & \checkmark &  &  &  &  & \checkmark &  &  &  & 2 \\
p23116\_\allowbreak i0 & Leg fat mass (left) &  & \checkmark &  &  &  &  &  &  & \checkmark &  & 2 \\
p26201\_\allowbreak a0 & PRS genetic principal components &  & \checkmark &  &  &  &  & \checkmark &  &  &  & 2 \\
p26227 & Standard PRS for coronary artery disease (CAD) &  &  & \checkmark &  &  &  &  &  &  & \checkmark & 2 \\
p30180\_\allowbreak i0 & Lymphocyte percentage &  &  &  &  &  & \checkmark & \checkmark &  &  &  & 2 \\
p30444 & Time usually got up in the morning during the past month &  &  &  & \checkmark &  &  & \checkmark &  &  &  & 2 \\
p30520\_\allowbreak i0 & Potassium in urine &  & \checkmark &  &  &  &  &  &  & \checkmark &  & 2 \\
p1508\_\allowbreak i0 & Coffee type &  &  &  &  & \checkmark &  &  &  &  &  & 1 \\
p20133\_\allowbreak i0\_\allowbreak a2 & Time to complete round &  &  &  &  &  &  &  &  &  & \checkmark & 1 \\
p20149\_\allowbreak i0\_\allowbreak a24 & Interval between previous point and current one in numeric path (trail \#1) & \checkmark &  &  &  &  &  &  &  &  &  & 1 \\
p20195\_\allowbreak i0 & Number of symbol digit matches attempted & \checkmark &  &  &  &  &  &  &  &  &  & 1 \\
p20230\_\allowbreak i0\_\allowbreak a10 & Duration to entering symbol choice &  &  &  &  &  &  &  &  &  & \checkmark & 1 \\
p20230\_\allowbreak i0\_\allowbreak a31 & Duration to entering symbol choice & \checkmark &  &  &  &  &  &  &  &  &  & 1 \\
p20676 & Liking for herring &  & \checkmark &  &  &  &  &  &  &  &  & 1 \\
p20717 & Liking for salty pretzels &  &  &  &  &  &  &  &  &  & \checkmark & 1 \\
p2080\_\allowbreak i0 & Frequency of tiredness / lethargy in last 2 weeks &  &  & \checkmark &  &  &  &  &  &  &  & 1 \\
p22009\_\allowbreak a21 & Genetic principal components &  &  &  &  &  &  &  &  &  & \checkmark & 1 \\
p23014\_\allowbreak i0 & LANA antigen for Kaposi's Sarcoma-Associated Herpesvirus &  &  &  &  &  &  &  & \checkmark &  &  & 1 \\
p23100\_\allowbreak i0 & Whole body fat mass &  & \checkmark &  &  &  &  &  &  &  &  & 1 \\
p23451\_\allowbreak i0 & Omega-3 Fatty Acids to Total Fatty Acids percentage &  &  &  &  &  &  &  &  & \checkmark &  & 1 \\
p23459\_\allowbreak i0 & Omega-6 Fatty Acids to Omega-3 Fatty Acids ratio &  &  & \checkmark &  &  &  &  &  &  &  & 1 \\
p23472\_\allowbreak i0 & Pyruvate &  &  &  &  &  &  &  &  &  & \checkmark & 1 \\
p23540\_\allowbreak i0 & Cholesterol in Medium LDL &  &  & \checkmark &  &  &  &  &  &  &  & 1 \\
p2405\_\allowbreak i0 & Number of children fathered &  &  &  &  &  &  &  &  &  & \checkmark & 1 \\
p26201\_\allowbreak a1 & PRS genetic principal components &  &  &  &  &  &  &  &  &  & \checkmark & 1 \\
p26201\_\allowbreak a2 & PRS genetic principal components & \checkmark &  &  &  &  &  &  &  &  &  & 1 \\
p29093 & Frequency of consuming six or more units of alcohol &  & \checkmark &  &  &  &  &  &  &  &  & 1 \\
p30140\_\allowbreak i0 & Neutrophil count &  &  &  &  & \checkmark &  &  &  &  &  & 1 \\
p30290\_\allowbreak i0 & High light scatter reticulocyte percentage &  &  & \checkmark &  &  &  &  &  &  &  & 1 \\
p30460 & Trouble sleeping in past month as disturbed by bed partner &  & \checkmark &  &  &  &  &  &  &  &  & 1 \\
p3062\_\allowbreak i0\_\allowbreak a2 & Forced vital capacity (FVC) &  &  &  &  &  &  &  &  &  & \checkmark & 1 \\
p30880\_\allowbreak i0 & Urate &  & \checkmark &  &  &  &  &  &  &  &  & 1 \\
p31147\_\allowbreak i0 & Adjusted optic disk major axis length (right) &  &  & \checkmark &  &  &  &  &  &  &  & 1 \\
p31152\_\allowbreak i0 & Optic disk orientation (left) &  &  &  &  &  & \checkmark &  &  &  &  & 1 \\
p32108 & Problems with slow thinking speed in past 7 days &  &  &  &  &  &  & \checkmark &  &  &  & 1 \\
p32109 & Trouble with figuring things out or solving problems in past 7 days &  &  &  &  &  &  & \checkmark &  &  &  & 1 \\
p32110 & Number of brothers and sisters &  &  &  & \checkmark &  &  &  &  &  &  & 1 \\
p32309 & Experiences difficulty in imagining what it would be like to be someone else &  &  &  &  & \checkmark &  &  &  &  &  & 1 \\
p399\_\allowbreak i0\_\allowbreak a2 & Number of incorrect matches in round &  &  &  &  &  &  & \checkmark &  &  &  & 1 \\
p40008\_\allowbreak i0 & Age at cancer diagnosis &  &  & \checkmark &  &  &  &  &  &  &  & 1 \\
p4196\_\allowbreak i0 & Pulse wave peak to peak time &  &  &  & \checkmark &  &  &  &  &  &  & 1 \\
p4258\_\allowbreak i0\_\allowbreak a7 & Number entered by participant &  &  &  & \checkmark &  &  &  &  &  &  & 1 \\
p4290\_\allowbreak i0 & Duration screen displayed &  &  &  &  &  &  &  &  &  & \checkmark & 1 \\
p48\_\allowbreak i0 & Waist circumference &  & \checkmark &  &  &  &  &  &  &  &  & 1 \\
p4947\_\allowbreak i0 & FI2 : duration viewed & \checkmark &  &  &  &  &  &  &  &  &  & 1 \\
p4980\_\allowbreak i0 & FI5 : duration viewed &  &  &  &  &  &  & \checkmark &  &  &  & 1 \\
p5080\_\allowbreak i0\_\allowbreak a10 & Displayed letters in round (right) &  &  &  &  & \checkmark &  &  &  &  &  & 1 \\
p5111\_\allowbreak i0\_\allowbreak a5 & 3mm asymmetry angle (left) &  & \checkmark &  &  &  &  &  &  &  &  & 1 \\
p5113\_\allowbreak i0\_\allowbreak a2 & 6mm cylindrical power angle (left) &  &  & \checkmark &  &  &  &  &  &  &  & 1 \\
p5115\_\allowbreak i0\_\allowbreak a2 & 3mm cylindrical power angle (right) &  &  &  &  &  &  &  &  & \checkmark &  & 1 \\
p5118\_\allowbreak i0\_\allowbreak a0 & 6mm cylindrical power (left) &  &  &  &  &  &  & \checkmark &  &  &  & 1 \\
p5983\_\allowbreak i0\_\allowbreak a98 & ECG, heart rate &  &  &  &  &  &  &  &  & \checkmark &  & 1 \\
p5985\_\allowbreak i0\_\allowbreak a49 & Bicycle speed &  &  &  &  &  &  &  &  & \checkmark &  & 1 \\
p6164\_\allowbreak i0\_\allowbreak \_\allowbreak 4 & Types of physical activity: none of the above &  &  &  &  &  &  &  &  & \checkmark &  & 1 \\
p90013 & Standard deviation of acceleration &  & \checkmark &  &  &  &  &  &  &  &  & 1 \\
p90038 & Average acceleration 11:00 - 11:59 &  &  &  &  &  &  &  & \checkmark &  &  & 1 \\
p90089 & No-wear time bias adjusted acceleration median &  &  &  &  &  &  & \checkmark &  &  &  & 1 \\
p90099 & Fraction acceleration <= 8 milli-gravities & \checkmark &  &  &  &  &  &  &  &  &  & 1 \\
p90109 & Fraction acceleration <= 18 milli-gravities &  &  &  &  &  &  &  &  &  & \checkmark & 1 \\
p90129 & Fraction acceleration <= 150 milli-gravities &  &  &  &  &  &  & \checkmark &  &  &  & 1 \\
p90186 & Maximum readings exceeding +/-8 gravities after calibration in a 5 second epoch &  & \checkmark &  &  &  &  &  &  &  &  & 1 \\
\end{longtable}
\end{footnotesize}

\subsection{Missingness as a predictive signal}

This section quantifies how much information measurement availability carries about subsequent dementia, using the full study cohort, outcome definitions and predictor set of the main analysis.

\subsubsection{Methods}

We classified a predictor as missing if no value remained in the analysis frame after the recoding under ``Missing-value handling'' in Supplementary Methods, so random imputation filled the entry. Of the 38 ITU-DESC predictors, age was complete for all participants. Each of the remaining 37 therefore defined a binary missingness indicator. Two pairs of predictors had identical missingness patterns: the polygenic risk scores and the maternal-history indicators.

In the full study cohort (500{,}410 participants), we estimated the association between each predictor's missingness and incident all-cause dementia or Alzheimer's disease using odds ratios from unadjusted and adjusted logistic regression models. Adjusted models included age as linear and quadratic terms, sex, education, Townsend deprivation index, recruitment year, assessment center and ethnicity. We modeled education and deprivation as categorical variables with an explicit missing category. We also report the standardized difference in missingness prevalence between cases and non-cases, together with the corresponding chi-square test. Benjamini--Hochberg \(q\)-values controlled the false discovery rate across the 37 predictors. We examined outcome incidence by the number of missing predictors per participant.

\subsubsection{Results}

For every predictor and both outcomes, missingness was more frequent among participants who subsequently developed the outcome than among non-cases (Tables~\ref{tab:missingness-association-dm} and~\ref{tab:missingness-association-ad}). For all-cause dementia, the standardized difference in missingness ranged from 0.02 to 0.25, with eight predictors at or above 0.10. For Alzheimer's disease, the corresponding range was 0.01 to 0.24, with five predictors at or above 0.10. The largest differences involved the pairs-matching test, peak expiratory flow and questionnaire variables such as length of mobile phone use. No blood assay reached 0.10 for either outcome, and the smallest differences involved touchscreen duration, the maternal-history indicators and several blood assays. After full adjustment, missingness remained associated with all-cause dementia at \(q<0.05\) for 33 of 37 predictors, with adjusted odds ratios ranging from 1.07 to 2.44; 16 of 37 predictors remained associated with Alzheimer's disease. All 37 unadjusted chi-square tests for dementia had \(P<0.05\). Given the large sample size, the standardized differences are more informative than the \(P\)-values: missingness was consistently associated with higher subsequent dementia risk, but the differences were modest for most predictors.

Dementia incidence rose from 1.76\% among participants with no missing predictors to 3.61\% among those missing three or four, then fell to 3.01\% among those missing five or more, mostly participants without blood-assay measurements (Table~\ref{tab:missingness-count}). Each additional missing predictor was associated with an adjusted odds ratio of 1.057 for incident dementia (95\% confidence interval [CI], 1.050 to 1.065). Participants with complete data for all 38 predictors made up 49.4\% of the cohort and were younger, less deprived and more highly educated than participants with at least one missing predictor. They also had lower incidence of both outcomes: 1.76\% versus 2.77\% for all-cause dementia (Table~\ref{tab:missingness-completeness}).

\subsubsection{Interpretation}

Within UK Biobank, missingness remained informative about subsequent outcomes after adjusting for measured demographic and socioeconomic characteristics. These analyses cannot distinguish missingness that is conditionally random, given unmeasured covariates, from missingness that is not at random, nor can they establish whether the observed associations would transport to other settings. The strongest signals arose from non-completion of cognitive testing and questionnaire non-response, which may reflect participant health status. Assay availability, more closely tied to assessment procedures, showed weaker associations. Because we cannot reliably separate these sources in advance, we withheld both from the model. The binary outcome definitions also classify participants who died before a dementia diagnosis as non-cases; the ``Competing risk of death'' subsection below quantifies this competing event and shows that it leaves score ranking and calibration by risk decile unchanged.

\begin{table}[htbp]
\centering
\footnotesize
\setlength{\tabcolsep}{4pt}
\caption{Participants with a recorded value for all 38 ITU-DESC predictors compared with participants missing at least one, the study cohort (500{,}410 participants). Means are shown with standard deviations and categories as column percentages. \textit{Difference} is the absolute difference in the units shown (percentage points for categories). \textit{SMD} is the standardized mean difference, complete minus incomplete, so a negative value means the complete group has the lower value; for categories it is the standardized difference of two proportions. \(P\)-values are from Welch's \(t\)-test for means and Pearson's chi-square test for categories.}
\label{tab:missingness-completeness}
\begin{tabular}{>{\raggedright\arraybackslash}p{4.4cm}llrrl}
\toprule
Characteristic & Complete & Incomplete & Difference & SMD & \(P\) \\
& (\(n=247{,}200\)) & (\(n=253{,}210\)) & & & \\
\midrule
Age at recruitment, years & 56.50 (8.10) & 57.64 (8.04) & 1.14 & $-$0.142 & $<$0.001 \\
Female, \% & 54.19 & 54.61 & 0.42 & $-$0.008 & 0.003 \\
Townsend deprivation index & $-$1.56 (2.94) & $-$1.04 (3.22) & 0.52 & $-$0.168 & $<$0.001 \\
Education, years & 14.72 (4.29) & 14.03 (4.28) & 0.70 & 0.163 & $<$0.001 \\
\multicolumn{6}{l}{Education band, \%}\\
\quad Fewer than 7 years & 0.08 & 0.21 & 0.13 & $-$0.034 & $<$0.001 \\
\quad 7 to 9 years & 0.90 & 2.07 & 1.17 & $-$0.097 & $<$0.001 \\
\quad 10 years or more & 98.95 & 96.32 & 2.63 & 0.174 & $<$0.001 \\
\quad Not recorded & 0.06 & 1.40 & 1.33 & $-$0.157 & $<$0.001 \\
\multicolumn{6}{l}{Ethnicity, \%}\\
\quad White & 95.47 & 92.73 & 2.74 & 0.116 & $<$0.001 \\
\quad Not recorded & 0.28 & 0.81 & 0.53 & $-$0.071 & $<$0.001 \\
\multicolumn{6}{l}{Incident all-cause dementia, \%}\\
\quad All-incident & 1.76 & 2.77 & 1.01 & $-$0.068 & $<$0.001 \\
\quad 10-year & 0.62 & 1.07 & 0.45 & $-$0.049 & $<$0.001 \\
\quad 5-year & 0.09 & 0.21 & 0.12 & $-$0.030 & $<$0.001 \\
\multicolumn{6}{l}{Incident Alzheimer's disease, \%}\\
\quad All-incident & 0.82 & 1.23 & 0.41 & $-$0.041 & $<$0.001 \\
\quad 10-year & 0.25 & 0.42 & 0.17 & $-$0.029 & $<$0.001 \\
\quad 5-year & 0.03 & 0.07 & 0.04 & $-$0.019 & $<$0.001 \\
\bottomrule
\end{tabular}
\end{table}

\begin{table}[htbp]
\centering
\footnotesize
\setlength{\tabcolsep}{3pt}
\caption{Incidence by number of missing ITU-DESC predictors in the study cohort (500{,}410 participants; all-incident horizon). Incidence intervals are Wilson 95\% confidence intervals. Odds ratios are relative to participants with no missing predictor, from a logistic model adjusted for age (linear and quadratic), sex, education, Townsend deprivation, recruitment year, assessment center and ethnicity. The per-predictor row enters the number of missing predictors as a linear term in the same logistic model.}
\label{tab:missingness-count}
\begin{tabular}{>{\raggedright\arraybackslash}p{4.85cm}rrlll}
\toprule
& & & Incidence, \% & Adjusted OR & \\
Missing predictors & \(n\) & Cases & (95\% CI) & (95\% CI) & \(P\) \\
\midrule
\multicolumn{6}{l}{\textbf{All-cause dementia}}\\
\quad 0 & 247{,}200 & 4{,}361 & 1.76 (1.71, 1.82) & 1 (reference) & -- \\
\quad 1 & 138{,}901 & 3{,}300 & 2.38 (2.30, 2.46) & 1.18 (1.12, 1.23) & $<$0.001 \\
\quad 2 & 45{,}172 & 1{,}465 & 3.24 (3.08, 3.41) & 1.47 (1.38, 1.56) & $<$0.001 \\
\quad 3--4 & 29{,}081 & 1{,}049 & 3.61 (3.40, 3.83) & 1.61 (1.50, 1.73) & $<$0.001 \\
\quad $\geq$5 & 40{,}056 & 1{,}205 & 3.01 (2.85, 3.18) & 1.50 (1.40, 1.61) & $<$0.001 \\
\quad Per additional missing predictor, OR & & & & 1.057 (1.050, 1.065) & $<$0.001 \\
\multicolumn{6}{l}{\textbf{Alzheimer's disease}}\\
\quad 0 & 247{,}200 & 2{,}015 & 0.82 (0.78, 0.85) & 1 (reference) & -- \\
\quad 1 & 138{,}901 & 1{,}484 & 1.07 (1.02, 1.12) & 1.13 (1.06, 1.21) & $<$0.001 \\
\quad 2 & 45{,}172 & 670 & 1.48 (1.38, 1.60) & 1.41 (1.29, 1.54) & $<$0.001 \\
\quad 3--4 & 29{,}081 & 463 & 1.59 (1.45, 1.74) & 1.51 (1.36, 1.67) & $<$0.001 \\
\quad $\geq$5 & 40{,}056 & 490 & 1.22 (1.12, 1.34) & 1.30 (1.17, 1.45) & $<$0.001 \\
\quad Per additional missing predictor, OR & & & & 1.042 (1.030, 1.053) & $<$0.001 \\
\bottomrule
\end{tabular}
\end{table}

\begin{footnotesize}
\setlength{\tabcolsep}{3pt}
\setlength{\LTcapwidth}{\linewidth}\setlength{\LTleft}{\fill}\setlength{\LTright}{\fill}
\begin{longtable}{>{\raggedright\arraybackslash}p{4.3cm}rrrrrlr}
\caption{Missingness of each ITU-DESC predictor and incident all-cause dementia in the study cohort (500{,}410 participants; all-incident horizon). \textit{Missing, \%} is the share of participants with no recorded value after the recoding described under ``Missing-value handling'' in Supplementary Methods, overall and separately among participants who did and did not develop all-cause dementia. \textit{SMD} is the standardized difference in the missing rate between cases and non-cases; 8 of 37 predictors reach 0.10 and 37 of 37 unadjusted chi-square tests have \(P<0.05\). \textit{Crude OR} is the unadjusted odds ratio of all-cause dementia for missing versus recorded. \textit{Adjusted OR} is from a logistic model adjusted for age (linear and quadratic), sex, education, Townsend deprivation, recruitment year, assessment center and ethnicity; 33 of 37 fitted predictors keep a Benjamini--Hochberg \(q<0.05\). Age was recorded for every participant. Rows are ordered by SMD.}
\label{tab:missingness-association-dm}\\
\toprule
& \multicolumn{3}{c}{Missing, \%} & & Crude & Adjusted OR & BH \\
\cmidrule(lr){2-4}
Predictor & Overall & Cases & Non-cases & SMD & OR & (95\% CI) & \(q\) \\
\midrule
\endfirsthead
\multicolumn{8}{l}{\textit{Table~\ref{tab:missingness-association-dm} continued from previous page}}\\
\toprule
& \multicolumn{3}{c}{Missing, \%} & & Crude & Adjusted OR & BH \\
\cmidrule(lr){2-4}
Predictor & Overall & Cases & Non-cases & SMD & OR & (95\% CI) & \(q\) \\
\midrule
\endhead
\midrule
\multicolumn{8}{r}{\textit{Continued on next page}}\\
\endfoot
\bottomrule
\endlastfoot
Pairs-matching completion time, round 2 & 3.94 & 10.15 & 3.79 & 0.252 & 2.86 & 2.12 (1.97, 2.28) & $<$0.001 \\
Pairs-matching completion time, round 1 & 3.73 & 9.46 & 3.59 & 0.239 & 2.80 & 2.06 (1.91, 2.22) & $<$0.001 \\
Peak expiratory flow (PEF) & 9.76 & 14.35 & 9.65 & 0.145 & 1.57 & 1.18 (1.12, 1.25) & $<$0.001 \\
Length of mobile phone use & 1.60 & 3.80 & 1.55 & 0.140 & 2.51 & 1.82 (1.64, 2.02) & $<$0.001 \\
Mean time to correctly identify matches & 1.15 & 2.72 & 1.11 & 0.117 & 2.48 & 1.84 (1.62, 2.09) & $<$0.001 \\
Mother's vital status or age at death & 3.02 & 5.30 & 2.96 & 0.117 & 1.83 & 1.47 (1.34, 1.60) & $<$0.001 \\
Other exercises (swimming, cycling, keep fit, bowling) & 1.25 & 2.79 & 1.22 & 0.112 & 2.33 & 2.24 (1.86, 2.70) & $<$0.001 \\
Time spent driving & 1.98 & 3.59 & 1.94 & 0.101 & 1.88 & 1.42 (1.27, 1.58) & $<$0.001 \\
Creatinine (enzymatic) in urine & 3.61 & 5.58 & 3.56 & 0.097 & 1.60 & 1.45 (1.32, 1.58) & $<$0.001 \\
Worry too long after embarrassment & 4.44 & 6.49 & 4.39 & 0.093 & 1.51 & 1.30 (1.20, 1.41) & $<$0.001 \\
Number of vehicles in household & 0.92 & 1.99 & 0.89 & 0.092 & 2.26 & 1.53 (1.32, 1.78) & $<$0.001 \\
Time spent watching television (TV) & 1.08 & 2.19 & 1.05 & 0.090 & 2.11 & 1.71 (1.49, 1.97) & $<$0.001 \\
Eats eggs, dairy, wheat, sugar & 0.93 & 1.77 & 0.91 & 0.075 & 1.97 & 1.80 (1.08, 2.99) & 0.027 \\
Long-standing illness, disability or infirmity & 2.65 & 3.93 & 2.62 & 0.073 & 1.52 & 1.37 (1.24, 1.52) & $<$0.001 \\
Mood swings & 2.84 & 4.15 & 2.81 & 0.073 & 1.49 & 1.24 (1.13, 1.37) & $<$0.001 \\
Standard PRS for Parkinson's disease (PD) & 3.22 & 4.30 & 3.20 & 0.058 & 1.36 & 1.25 (1.12, 1.39) & $<$0.001 \\
Standard PRS for Alzheimer's disease (AD) & 3.22 & 4.30 & 3.20 & 0.058 & 1.36 & 1.25 (1.12, 1.39) & $<$0.001 \\
Vitamin D & 10.77 & 12.46 & 10.73 & 0.054 & 1.18 & 1.14 (1.07, 1.21) & $<$0.001 \\
Hearing difficulty/problems with background noise & 2.32 & 3.17 & 2.30 & 0.053 & 1.39 & 1.41 (1.26, 1.58) & $<$0.001 \\
Falls in the last year & 0.53 & 0.98 & 0.52 & 0.053 & 1.89 & 2.44 (1.94, 3.09) & $<$0.001 \\
Water intake & 0.95 & 1.51 & 0.94 & 0.052 & 1.62 & 1.33 (1.12, 1.57) & 0.001 \\
Taking other prescription medications & 0.61 & 1.06 & 0.60 & 0.051 & 1.79 & 1.79 (1.45, 2.22) & $<$0.001 \\
\textit{APOE} $\varepsilon$4 copies & 17.78 & 19.68 & 17.73 & 0.050 & 1.14 & 1.07 (1.02, 1.13) & 0.005 \\
Hand grip strength (left) & 0.67 & 1.12 & 0.66 & 0.049 & 1.71 & 1.53 (1.27, 1.85) & $<$0.001 \\
C-reactive protein & 6.74 & 8.00 & 6.71 & 0.049 & 1.21 & 1.14 (1.06, 1.24) & $<$0.001 \\
Neutrophil percentage & 5.01 & 6.05 & 4.98 & 0.047 & 1.23 & 1.09 (1.00, 1.20) & 0.059 \\
Basophil count & 5.01 & 6.05 & 4.98 & 0.047 & 1.23 & 1.09 (1.00, 1.20) & 0.059 \\
IGF-1 & 7.05 & 8.26 & 7.02 & 0.047 & 1.19 & 1.13 (1.04, 1.22) & 0.003 \\
Alanine aminotransferase & 6.58 & 7.74 & 6.55 & 0.046 & 1.20 & 1.13 (1.05, 1.23) & 0.003 \\
Sleeplessness/insomnia & 0.30 & 0.59 & 0.29 & 0.045 & 2.04 & 2.24 (1.62, 3.11) & $<$0.001 \\
Number of treatments/medications taken & 0.17 & 0.40 & 0.16 & 0.044 & 2.44 & 2.20 (1.57, 3.09) & $<$0.001 \\
Aspartate aminotransferase & 6.90 & 8.02 & 6.87 & 0.044 & 1.18 & 1.12 (1.04, 1.21) & 0.004 \\
Saturated Fatty Acids to Total Fatty Acids percentage & 2.87 & 3.57 & 2.86 & 0.040 & 1.26 & 1.26 (1.13, 1.39) & $<$0.001 \\
Adjusted T/S ratio & 5.94 & 6.78 & 5.92 & 0.035 & 1.16 & 1.07 (0.98, 1.16) & 0.127 \\
Maternal history of high blood pressure & 1.91 & 2.37 & 1.89 & 0.033 & 1.26 & 1.22 (1.07, 1.39) & 0.003 \\
Maternal history of Alzheimer's disease/dementia & 1.91 & 2.37 & 1.89 & 0.033 & 1.26 & 1.22 (1.07, 1.39) & 0.003 \\
Touchscreen duration & 0.17 & 0.28 & 0.17 & 0.025 & 1.70 & 1.59 (0.95, 2.67) & 0.079 \\
\end{longtable}
\end{footnotesize}

\begin{footnotesize}
\setlength{\tabcolsep}{3pt}
\setlength{\LTcapwidth}{\linewidth}\setlength{\LTleft}{\fill}\setlength{\LTright}{\fill}
\begin{longtable}{>{\raggedright\arraybackslash}p{4.3cm}rrrrrlr}
\caption{Missingness of each ITU-DESC predictor and incident Alzheimer's disease in the study cohort (500{,}410 participants; all-incident horizon). \textit{Missing, \%} is the share of participants with no recorded value after the recoding described under ``Missing-value handling'' in Supplementary Methods, overall and separately among participants who did and did not develop Alzheimer's disease. \textit{SMD} is the standardized difference in the missing rate between cases and non-cases; 5 of 37 predictors reach 0.10 and 27 of 37 unadjusted chi-square tests have \(P<0.05\). \textit{Crude OR} is the unadjusted odds ratio of Alzheimer's disease for missing versus recorded. \textit{Adjusted OR} is from a logistic model adjusted for age (linear and quadratic), sex, education, Townsend deprivation, recruitment year, assessment center and ethnicity; 16 of 37 fitted predictors keep a Benjamini--Hochberg \(q<0.05\). Age was recorded for every participant. Rows are ordered by SMD.}
\label{tab:missingness-association-ad}\\
\toprule
& \multicolumn{3}{c}{Missing, \%} & & Crude & Adjusted OR & BH \\
\cmidrule(lr){2-4}
Predictor & Overall & Cases & Non-cases & SMD & OR & (95\% CI) & \(q\) \\
\midrule
\endfirsthead
\multicolumn{8}{l}{\textit{Table~\ref{tab:missingness-association-ad} continued from previous page}}\\
\toprule
& \multicolumn{3}{c}{Missing, \%} & & Crude & Adjusted OR & BH \\
\cmidrule(lr){2-4}
Predictor & Overall & Cases & Non-cases & SMD & OR & (95\% CI) & \(q\) \\
\midrule
\endhead
\midrule
\multicolumn{8}{r}{\textit{Continued on next page}}\\
\endfoot
\bottomrule
\endlastfoot
Pairs-matching completion time, round 2 & 3.94 & 10.00 & 3.88 & 0.243 & 2.75 & 2.08 (1.87, 2.31) & $<$0.001 \\
Pairs-matching completion time, round 1 & 3.73 & 9.27 & 3.67 & 0.229 & 2.68 & 2.01 (1.80, 2.24) & $<$0.001 \\
Length of mobile phone use & 1.60 & 3.90 & 1.57 & 0.143 & 2.54 & 1.82 (1.56, 2.11) & $<$0.001 \\
Mother's vital status or age at death & 3.02 & 5.29 & 2.99 & 0.116 & 1.81 & 1.51 (1.33, 1.71) & $<$0.001 \\
Worry too long after embarrassment & 4.44 & 6.83 & 4.42 & 0.105 & 1.59 & 1.35 (1.21, 1.52) & $<$0.001 \\
Mean time to correctly identify matches & 1.15 & 2.44 & 1.14 & 0.098 & 2.17 & 1.63 (1.34, 1.97) & $<$0.001 \\
Peak expiratory flow (PEF) & 9.76 & 12.65 & 9.73 & 0.093 & 1.34 & 1.01 (0.92, 1.10) & 0.873 \\
Time spent watching television (TV) & 1.08 & 2.11 & 1.06 & 0.084 & 2.00 & 1.61 (1.31, 1.97) & $<$0.001 \\
Number of vehicles in household & 0.92 & 1.80 & 0.91 & 0.077 & 1.99 & 1.39 (1.11, 1.74) & 0.011 \\
Mood swings & 2.84 & 4.24 & 2.83 & 0.076 & 1.52 & 1.28 (1.11, 1.47) & 0.002 \\
Long-standing illness, disability or infirmity & 2.65 & 3.94 & 2.64 & 0.073 & 1.51 & 1.37 (1.18, 1.58) & $<$0.001 \\
Time spent driving & 1.98 & 3.07 & 1.96 & 0.070 & 1.58 & 1.18 (1.00, 1.39) & 0.106 \\
Other exercises (swimming, cycling, keep fit, bowling) & 1.25 & 2.13 & 1.24 & 0.069 & 1.73 & 1.25 (0.89, 1.73) & 0.257 \\
Hearing difficulty/problems with background noise & 2.32 & 3.30 & 2.31 & 0.060 & 1.44 & 1.49 (1.27, 1.75) & $<$0.001 \\
Eats eggs, dairy, wheat, sugar & 0.93 & 1.56 & 0.92 & 0.058 & 1.71 & 1.67 (0.75, 3.71) & 0.259 \\
Creatinine (enzymatic) in urine & 3.61 & 4.65 & 3.60 & 0.053 & 1.31 & 1.15 (1.00, 1.33) & 0.097 \\
Water intake & 0.95 & 1.52 & 0.94 & 0.052 & 1.62 & 1.37 (1.07, 1.74) & 0.028 \\
Falls in the last year & 0.53 & 0.92 & 0.52 & 0.047 & 1.76 & 2.47 (1.75, 3.49) & $<$0.001 \\
\textit{APOE} $\varepsilon$4 copies & 17.78 & 19.35 & 17.76 & 0.041 & 1.11 & 1.07 (0.99, 1.15) & 0.158 \\
Standard PRS for Parkinson's disease (PD) & 3.22 & 3.90 & 3.22 & 0.037 & 1.22 & 1.12 (0.95, 1.33) & 0.257 \\
Standard PRS for Alzheimer's disease (AD) & 3.22 & 3.90 & 3.22 & 0.037 & 1.22 & 1.12 (0.95, 1.33) & 0.257 \\
Sleeplessness/insomnia & 0.30 & 0.53 & 0.29 & 0.036 & 1.80 & 2.19 (1.33, 3.60) & 0.006 \\
Taking other prescription medications & 0.61 & 0.92 & 0.60 & 0.036 & 1.52 & 1.62 (1.17, 2.25) & 0.011 \\
Vitamin D & 10.77 & 11.85 & 10.76 & 0.035 & 1.12 & 1.05 (0.96, 1.15) & 0.345 \\
Number of treatments/medications taken & 0.17 & 0.33 & 0.17 & 0.033 & 2.01 & 1.91 (1.13, 3.22) & 0.035 \\
Saturated Fatty Acids to Total Fatty Acids percentage & 2.87 & 3.40 & 2.87 & 0.030 & 1.19 & 1.18 (1.01, 1.38) & 0.071 \\
Hand grip strength (left) & 0.67 & 0.92 & 0.67 & 0.028 & 1.38 & 1.23 (0.91, 1.65) & 0.257 \\
Adjusted T/S ratio & 5.94 & 6.46 & 5.94 & 0.022 & 1.09 & 1.03 (0.90, 1.16) & 0.752 \\
IGF-1 & 7.05 & 7.58 & 7.04 & 0.021 & 1.08 & 1.02 (0.91, 1.15) & 0.752 \\
Alanine aminotransferase & 6.58 & 7.09 & 6.57 & 0.020 & 1.08 & 1.03 (0.91, 1.16) & 0.752 \\
C-reactive protein & 6.74 & 7.24 & 6.74 & 0.020 & 1.08 & 1.02 (0.91, 1.15) & 0.752 \\
Touchscreen duration & 0.17 & 0.25 & 0.17 & 0.019 & 1.52 & 1.68 (0.75, 3.76) & 0.259 \\
Maternal history of high blood pressure & 1.91 & 2.17 & 1.90 & 0.019 & 1.14 & 1.14 (0.94, 1.39) & 0.257 \\
Maternal history of Alzheimer's disease/dementia & 1.91 & 2.17 & 1.90 & 0.019 & 1.14 & 1.14 (0.94, 1.39) & 0.257 \\
Aspartate aminotransferase & 6.90 & 7.30 & 6.89 & 0.016 & 1.06 & 1.01 (0.90, 1.14) & 0.873 \\
Basophil count & 5.01 & 5.21 & 5.01 & 0.009 & 1.04 & 0.90 (0.78, 1.04) & 0.257 \\
Neutrophil percentage & 5.01 & 5.21 & 5.01 & 0.009 & 1.04 & 0.90 (0.78, 1.04) & 0.257 \\
\end{longtable}
\end{footnotesize}
 
\subsection{Competing risk of death}

The main analyses treat dementia as a binary outcome and count participants who died without a recorded dementia diagnosis as non-cases. This section quantifies death as a competing event and re-evaluates all scores on the holdout test set using two time-to-event frameworks. No prediction model was retrained. All analyses used the fixed holdout-test predictions from the main analysis and retained its definition of all-cause dementia; rank-based measures used the uncalibrated predictions. The hazard models used each participant's fixed score as the sole covariate to estimate its association with time to dementia or death.

\subsubsection{Methods}

Follow-up extended from the baseline visit to the earliest of a recorded dementia diagnosis, death or the administrative censoring date of 20 November 2025. Because the latest baseline visit occurred on 1 October 2010 and we excluded participants lost to follow-up during cohort construction, all participants had at least 15.1 years of potential follow-up (maximum, 19.7 years), and none was censored before 15 years. At the 5-, 10- and 15-year horizons, each participant therefore fell into exactly one of three states: dementia recorded by the horizon; death without dementia by the horizon; or alive and dementia-free at the horizon (Table~\ref{tab:competing-risk-status}).

Three consequences follow. First, the binary horizon labels used in the main analysis coincide with the observed cumulative-incidence status, and the Aalen--Johansen cumulative incidence of dementia~\cite{aalen1978} at each horizon equals the observed proportion. Second, both time-dependent AUC definitions described by Blanche et al.~\cite{blanche2013} can be estimated without inverse-probability weighting. The cumulative-incidence AUC treats as controls all participants without dementia by the horizon, including those who died, and therefore equals the corresponding binary AUC; the cause-specific AUC uses as controls only participants who were alive and dementia-free at the horizon. Third, in the absence of censoring before the horizon, the Fine--Gray subdistribution hazard model~\cite{finegray1999} reduces to a Cox model with a modified risk set in which participants who die without dementia remain at risk until the horizon~\cite{geskus2011}. We therefore fitted it alongside a cause-specific Cox model, which censored deaths at the time of death. Both models used the linear predictor of the fixed score as their only covariate.

For each score and horizon, we report both time-dependent AUCs with DeLong CIs; Harrell's concordance index, with deaths censored at the time of death; and the concordance index of Wolbers et al.~\cite{wolbers2014}, in which participants who died remain comparable as non-events until the horizon. We obtained CIs for both concordance indices by paired participant-level bootstrap resampling (2{,}000 resamples). We also report the AUC of each score for death by the horizon among participants without dementia, quantifying how well each score discriminates the competing event. For ITU-DESC, we report cause-specific and Fine--Gray hazard ratios per standard-deviation increase in the linear predictor. Across deciles of the score, we report the Aalen--Johansen cumulative incidence of dementia and death, together with the Kaplan--Meier estimate that treats death as censoring. The two concordance indices target different quantities: with deaths censored, \(C\) measures concordance among pairs still alive and dementia-free at the time of the earlier dementia event and therefore depends on the mortality distribution; with deaths retained as non-events, it targets discrimination of the observable cumulative incidence. We did not estimate an AUC that treats death as non-informative censoring because this quantity corresponds to a hypothetical setting in which death is eliminated and requires conditional independence of death and dementia given the score, an assumption that cannot be verified from these data.
 
\subsubsection{Results}

In the holdout test set, 4{,}880 participants died without a recorded dementia diagnosis during full follow-up, compared with 1{,}138 dementia cases: 4.29 deaths per case. At 5, 10 and 15 years, the corresponding ratios were 11.23, 5.41 and 4.09 (Table~\ref{tab:competing-risk-status}). The eight scores kept the same rank ordering across all frameworks at each horizon, with ITU-DESC first throughout (Table~\ref{tab:competing-risk-discrimination}). At 10 years, ITU-DESC had a cumulative-incidence AUC of 0.850 (95\% CI, 0.833 to 0.867), a cause-specific AUC of 0.857 (0.841 to 0.873), a concordance index of 0.853 with deaths censored and a concordance index of 0.849 (0.831 to 0.865) with deaths retained as non-events. ITU-DESC exceeded every comparator; at 10 and 15 years, all CIs for these differences excluded zero, whereas at 5 years, with only 75 dementia cases, the intervals for the difference from UKB-DRP included zero.

The scores also discriminated death without dementia. For ITU-DESC, the AUC for this competing event was 0.698, 0.714 and 0.732 at 5, 10 and 15 years, respectively, and the cause-specific hazard ratio for death at 10 years was 2.05 per standard-deviation increase in the linear predictor (95\% CI, 1.96 to 2.13). Consistent with this association, the 10-year Fine--Gray hazard ratio for dementia was lower than the corresponding cause-specific hazard ratio: 4.16 (3.75 to 4.62) versus 4.25 (3.83 to 4.71) per standard-deviation increase.

\begin{figure}[htbp]
    \centering
    \includegraphics{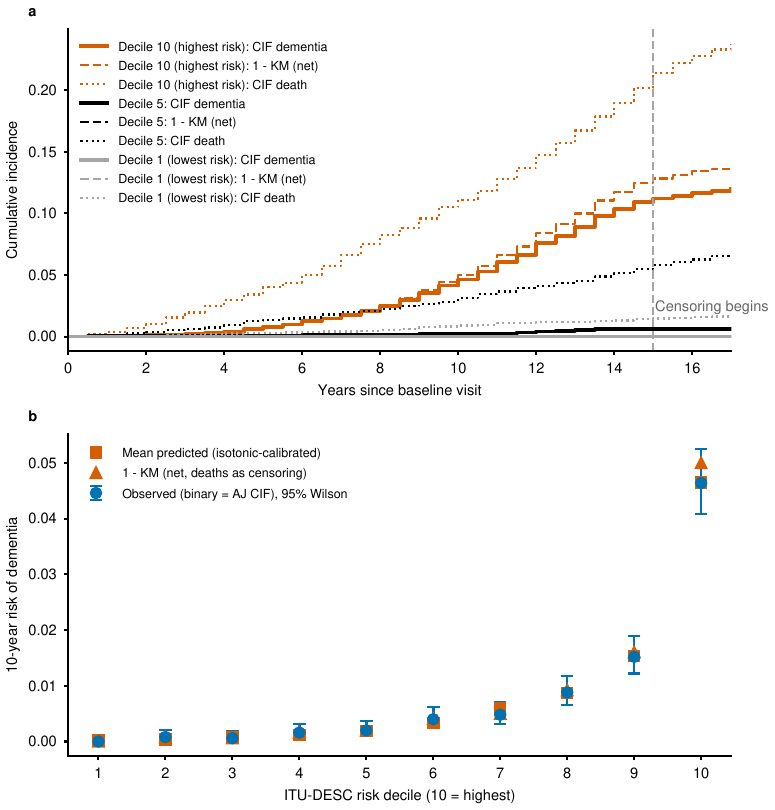}
    \caption{Competing risk of death across deciles of the ITU-DESC score in the holdout test set (50{,}041 participants). \textbf{a}, Aalen--Johansen cumulative incidence of dementia and death without dementia, together with the Kaplan--Meier estimate treating death as censoring, for the highest-, fifth- and lowest-risk deciles. The vertical line marks 15 years, after which administrative censoring begins. \textbf{b}, Ten-year risk by decile, showing the mean isotonic-calibrated prediction, the observed proportion with its Wilson 95\% confidence interval, and the Kaplan--Meier estimate treating death as censoring. The observed proportion is identical to the Aalen--Johansen cumulative incidence at this horizon.}
    \label{fig:competing-risk}
\end{figure}

The observed 10-year dementia proportion equaled the Aalen--Johansen cumulative incidence in every decile of the ITU-DESC score, and the mean calibrated prediction tracked the observed risk. In the highest-risk decile, the mean predicted risk was 4.65\%, compared with an observed risk of 4.64\% (95\% CI, 4.09 to 5.25; Table~\ref{tab:competing-risk-deciles} and Fig.~\ref{fig:competing-risk}). The corresponding Kaplan--Meier estimate, which treats death as censoring, was 5.00\%, higher than the observed proportion because it estimates risk under a hypothetical setting in which death is removed as a competing event. Deaths also clustered in the higher-risk deciles: among binary non-cases, 11.65\% of participants in the highest-risk decile had died by 10 years and 24.13\% by 15 years, compared with 0.90\% and 1.48\%, respectively, in the lowest-risk decile.

\subsubsection{Interpretation}

At fixed horizons, the binary evaluation used in the main analysis is equivalent to a cumulative-incidence evaluation because no participant was censored before 15 years. The competing event instead changes the interpretation of alternative estimands: treating death as censoring yields higher risk estimates, and the cause-specific hazard ratio exceeds the subdistribution hazard ratio because the score also predicts death. Neither difference changes score ranking or the agreement between predicted and observed risk. The remaining limitation of the binary framework is therefore the all-incident outcome, which combines follow-up durations of 15.1 to 19.7 years and is not a fixed-horizon cumulative-incidence measure, unlike the 5- and 10-year outcomes.

\begin{table}[htbp]
\centering
\footnotesize
\setlength{\tabcolsep}{4pt}
\caption{Follow-up status by horizon in the study cohort (500{,}410 participants) and the holdout test set (50{,}041 participants). At each fixed horizon every participant is in exactly one of three states: dementia recorded by the horizon, died without a dementia record by the horizon, or alive without a dementia record at the horizon; no participant was administratively censored before 15 years, and participants lost to follow-up were excluded at cohort construction. \textit{All-incident} is the manuscript's binary label over each participant's full follow-up (15.1 to 19.7 years), with the event-free column counting participants censored at the administrative date. \textit{Deaths per case} is the ratio of dementia-free deaths to dementia cases.}
\label{tab:competing-risk-status}
\begin{tabular*}{\linewidth}{@{\extracolsep{\fill}}lrrrrr}
\toprule
& & Dementia, & Died without & Alive without & Deaths \\
Horizon & \(n\) & \(n\) (\%) & dementia, \(n\) (\%) & dementia, \(n\) (\%) & per case \\
\midrule
\multicolumn{6}{l}{\textbf{Study cohort}}\\
\quad 5 years & 500{,}410 & 756 (0.15) & 8{,}565 (1.71) & 491{,}089 (98.14) & 11.33 \\
\quad 10 years & 500{,}410 & 4{,}223 (0.84) & 23{,}107 (4.62) & 473{,}080 (94.54) & 5.47 \\
\quad 15 years & 500{,}410 & 10{,}792 (2.16) & 44{,}966 (8.99) & 444{,}652 (88.86) & 4.17 \\
\quad All-incident & 500{,}410 & 11{,}380 (2.27) & 49{,}366 (9.87) & 439{,}664 (87.86) & 4.34 \\
\multicolumn{6}{l}{\textbf{Holdout test set}}\\
\quad 5 years & 50{,}041 & 75 (0.15) & 842 (1.68) & 49{,}124 (98.17) & 11.23 \\
\quad 10 years & 50{,}041 & 421 (0.84) & 2{,}279 (4.55) & 47{,}341 (94.60) & 5.41 \\
\quad 15 years & 50{,}041 & 1{,}086 (2.17) & 4{,}438 (8.87) & 44{,}517 (88.96) & 4.09 \\
\quad All-incident & 50{,}041 & 1{,}138 (2.27) & 4{,}880 (9.75) & 44{,}023 (87.97) & 4.29 \\
\bottomrule
\end{tabular*}
\end{table}

\begin{table}[htbp]
\centering
\footnotesize
\setlength{\tabcolsep}{4pt}
\caption{Ten-year risk of all-cause dementia by decile of the ITU-DESC score in the holdout test set (50{,}041 participants; decile 10 is the highest risk). \textit{Predicted} is the mean isotonic-calibrated 10-year risk. \textit{Observed} is the share of the decile with dementia recorded within 10 years, which equals the Aalen--Johansen cumulative incidence with death as a competing event because no participant was censored before 15 years; the interval is a Wilson 95\% interval. \textit{\(1-\mathrm{KM}\)} treats death as censoring and estimates the risk that would apply if no one died. \textit{Death} is the Aalen--Johansen cumulative incidence of death without dementia. The last column is the share of the decile's binary non-cases who had died by 10 years rather than remaining alive without dementia.}
\label{tab:competing-risk-deciles}
\begin{tabular}{lrrlrrr}
\toprule
Decile & \(n\) & Predicted, \% & Observed, \% (95\% CI) & \(1-\mathrm{KM}\), \% & Death, \% & Non-cases who died, \% \\
\midrule
1 & 5{,}005 & 0.02 & 0.00 (0.00, 0.08) & 0.00 & 0.90 & 0.90 \\
2 & 5{,}004 & 0.04 & 0.08 (0.03, 0.21) & 0.08 & 1.34 & 1.34 \\
3 & 5{,}004 & 0.07 & 0.06 (0.02, 0.18) & 0.06 & 1.42 & 1.42 \\
4 & 5{,}004 & 0.12 & 0.16 (0.08, 0.32) & 0.16 & 2.56 & 2.56 \\
5 & 5{,}004 & 0.19 & 0.20 (0.11, 0.37) & 0.20 & 3.12 & 3.12 \\
6 & 5{,}004 & 0.33 & 0.40 (0.26, 0.62) & 0.41 & 3.88 & 3.89 \\
7 & 5{,}004 & 0.60 & 0.48 (0.32, 0.71) & 0.50 & 5.40 & 5.42 \\
8 & 5{,}004 & 0.87 & 0.88 (0.66, 1.18) & 0.92 & 7.21 & 7.28 \\
9 & 5{,}004 & 1.53 & 1.52 (1.22, 1.90) & 1.60 & 8.61 & 8.75 \\
10 & 5{,}004 & 4.65 & 4.64 (4.09, 5.25) & 5.00 & 11.11 & 11.65 \\
\bottomrule
\end{tabular}
\end{table}
 
\clearpage
\begin{landscape}
\begin{footnotesize}
\setlength{\tabcolsep}{3pt}
\setlength{\LTcapwidth}{\linewidth}\setlength{\LTleft}{0pt}\setlength{\LTright}{0pt}
\begin{longtable}{@{\extracolsep{\fill}}>{\raggedright\arraybackslash}p{3.3cm}>{\raggedleft\arraybackslash}p{2.5cm}>{\raggedleft\arraybackslash}p{2.5cm}>{\raggedleft\arraybackslash}p{2.5cm}>{\raggedleft\arraybackslash}p{2.5cm}>{\raggedleft\arraybackslash}p{2.5cm}>{\raggedleft\arraybackslash}p{1.3cm}}
\caption{Discrimination of each score on the holdout test set (50{,}041 participants) when death without dementia is treated as a competing event, at the 5-, 10- and 15-year horizons. Predictions are the uncalibrated holdout-test predictions of the main analysis; no prediction model was retrained. \textit{AUC, cumulative incidence} takes as controls every participant without dementia by the horizon, including those who died, and equals the binary AUC of the horizon label because no participant was censored before 15 years. \textit{AUC, cause-specific} takes as controls only participants alive and dementia-free at the horizon. \textit{\(C\), deaths censored} is Harrell's concordance with deaths censored at the date of death; \textit{\(C\), deaths as non-events} is the concordance of Wolbers et al., in which participants who died remain comparable non-events until the horizon. \(\Delta C\) is ITU-DESC minus the comparator, with a paired participant-level bootstrap interval (2{,}000 resamples). \textit{AUC for death} is the AUC of each score for death by the horizon among participants without dementia by the horizon, and measures how strongly the score predicts the competing event; its DeLong interval half-width is at most \(\pm 0.021\) and is given in the source data. AUC intervals are DeLong intervals.}
\label{tab:competing-risk-discrimination}\\
\toprule
Model & AUC, cumulative incidence (95\% CI) & AUC, cause-specific (95\% CI) & \(C\), deaths censored (95\% CI) & \(C\), deaths as non-events (95\% CI) & \(\Delta C\), deaths as non-events (95\% CI) & AUC for death \\
\midrule
\endfirsthead
\multicolumn{7}{l}{\textit{Table~\ref{tab:competing-risk-discrimination} continued from previous page}}\\
\toprule
Model & AUC, cumulative incidence (95\% CI) & AUC, cause-specific (95\% CI) & \(C\), deaths censored (95\% CI) & \(C\), deaths as non-events (95\% CI) & \(\Delta C\), deaths as non-events (95\% CI) & AUC for death \\
\midrule
\endhead
\midrule
\multicolumn{7}{r}{\textit{Continued on next page}}\\
\endfoot
\bottomrule
\endlastfoot
\multicolumn{7}{l}{\textbf{5 years}: 75 dementia cases, 842 deaths without dementia}\\
ITU-DESC & 0.839 (0.796, 0.882) & 0.842 (0.798, 0.884) & 0.840 (0.793, 0.882) & 0.839 (0.792, 0.881) & -- & 0.698 \\
UKB-DRP & 0.823 (0.778, 0.868) & 0.826 (0.781, 0.870) & 0.824 (0.774, 0.867) & 0.823 (0.773, 0.866) & 0.016 ($-$0.005, 0.037) & 0.700 \\
UKBDRS & 0.800 (0.752, 0.848) & 0.802 (0.755, 0.850) & 0.801 (0.750, 0.848) & 0.800 (0.749, 0.846) & 0.039 (0.012, 0.065) & 0.673 \\
UKBDRS (w/o \textit{APOE} $\varepsilon$4) & 0.800 (0.753, 0.848) & 0.803 (0.755, 0.851) & 0.801 (0.751, 0.848) & 0.800 (0.750, 0.847) & 0.039 (0.010, 0.066) & 0.683 \\
DRS & 0.762 (0.711, 0.812) & 0.764 (0.714, 0.814) & 0.762 (0.709, 0.812) & 0.761 (0.709, 0.811) & 0.078 (0.050, 0.106) & 0.661 \\
CAIDE & 0.657 (0.604, 0.710) & 0.659 (0.606, 0.712) & 0.658 (0.604, 0.710) & 0.657 (0.604, 0.709) & 0.182 (0.122, 0.240) & 0.619 \\
CAIDE (w/o \textit{APOE} $\varepsilon$4) & 0.626 (0.576, 0.676) & 0.628 (0.579, 0.678) & 0.627 (0.576, 0.675) & 0.626 (0.575, 0.674) & 0.213 (0.151, 0.274) & 0.621 \\
ANU-ADRI & 0.633 (0.572, 0.694) & 0.635 (0.574, 0.696) & 0.634 (0.572, 0.692) & 0.633 (0.571, 0.691) & 0.206 (0.129, 0.283) & 0.594 \\
\multicolumn{7}{l}{\textbf{10 years}: 421 dementia cases, 2{,}279 deaths without dementia}\\
ITU-DESC & 0.850 (0.833, 0.867) & 0.857 (0.841, 0.873) & 0.853 (0.836, 0.869) & 0.849 (0.831, 0.865) & -- & 0.714 \\
UKB-DRP & 0.836 (0.819, 0.854) & 0.844 (0.827, 0.861) & 0.840 (0.823, 0.856) & 0.835 (0.818, 0.852) & 0.014 (0.004, 0.024) & 0.717 \\
UKBDRS & 0.813 (0.794, 0.832) & 0.820 (0.802, 0.839) & 0.816 (0.798, 0.834) & 0.812 (0.793, 0.830) & 0.037 (0.026, 0.048) & 0.692 \\
UKBDRS (w/o \textit{APOE} $\varepsilon$4) & 0.810 (0.791, 0.828) & 0.818 (0.799, 0.836) & 0.813 (0.794, 0.831) & 0.808 (0.789, 0.826) & 0.040 (0.028, 0.053) & 0.704 \\
DRS & 0.781 (0.761, 0.800) & 0.788 (0.769, 0.807) & 0.784 (0.764, 0.803) & 0.780 (0.760, 0.799) & 0.069 (0.055, 0.084) & 0.677 \\
CAIDE & 0.682 (0.660, 0.705) & 0.688 (0.666, 0.710) & 0.685 (0.662, 0.706) & 0.682 (0.659, 0.703) & 0.167 (0.145, 0.190) & 0.631 \\
CAIDE (w/o \textit{APOE} $\varepsilon$4) & 0.648 (0.627, 0.670) & 0.655 (0.634, 0.676) & 0.652 (0.630, 0.673) & 0.648 (0.625, 0.669) & 0.201 (0.176, 0.226) & 0.637 \\
ANU-ADRI & 0.621 (0.595, 0.648) & 0.625 (0.599, 0.652) & 0.623 (0.595, 0.650) & 0.621 (0.593, 0.647) & 0.228 (0.198, 0.261) & 0.586 \\
\multicolumn{7}{l}{\textbf{15 years}: 1{,}086 dementia cases, 4{,}438 deaths without dementia}\\
ITU-DESC & 0.853 (0.843, 0.863) & 0.868 (0.859, 0.878) & 0.858 (0.848, 0.867) & 0.849 (0.839, 0.859) & -- & 0.732 \\
UKB-DRP & 0.840 (0.829, 0.850) & 0.856 (0.846, 0.866) & 0.845 (0.835, 0.854) & 0.836 (0.825, 0.846) & 0.013 (0.007, 0.020) & 0.733 \\
UKBDRS & 0.824 (0.813, 0.835) & 0.840 (0.829, 0.850) & 0.829 (0.817, 0.839) & 0.821 (0.809, 0.831) & 0.029 (0.022, 0.036) & 0.712 \\
UKBDRS (w/o \textit{APOE} $\varepsilon$4) & 0.813 (0.801, 0.824) & 0.831 (0.820, 0.841) & 0.819 (0.807, 0.829) & 0.809 (0.798, 0.820) & 0.040 (0.033, 0.048) & 0.723 \\
DRS & 0.787 (0.776, 0.799) & 0.803 (0.792, 0.814) & 0.792 (0.781, 0.803) & 0.784 (0.773, 0.795) & 0.065 (0.056, 0.074) & 0.693 \\
CAIDE & 0.695 (0.682, 0.709) & 0.708 (0.695, 0.722) & 0.700 (0.686, 0.713) & 0.693 (0.679, 0.707) & 0.156 (0.143, 0.170) & 0.650 \\
CAIDE (w/o \textit{APOE} $\varepsilon$4) & 0.648 (0.634, 0.661) & 0.663 (0.649, 0.676) & 0.654 (0.640, 0.668) & 0.646 (0.632, 0.660) & 0.203 (0.188, 0.219) & 0.655 \\
ANU-ADRI & 0.593 (0.576, 0.609) & 0.601 (0.584, 0.618) & 0.597 (0.580, 0.614) & 0.592 (0.575, 0.609) & 0.258 (0.240, 0.277) & 0.587 \\
\end{longtable}
\end{footnotesize}
 \end{landscape}

\clearpage

\subsection{Feature importance without random imputation}

\begin{figure}[htbp]
    \centering
    \includegraphics{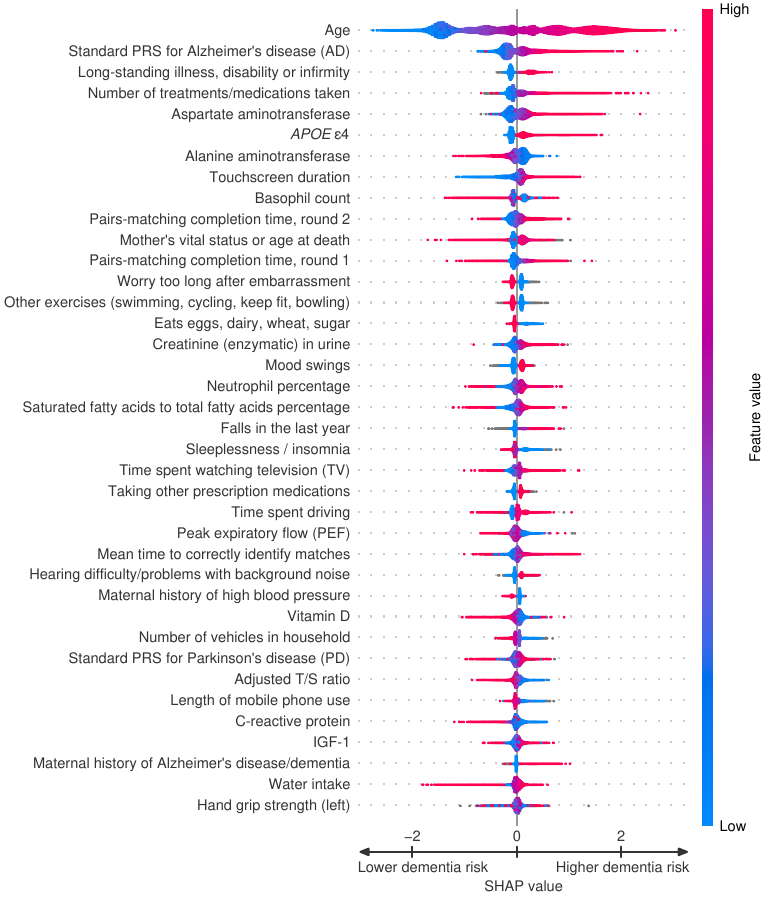}
    \caption{SHAP summary plot for ITU-DESC for all-incident dementia without random imputation. The model used the same 38 features as in the main analysis, but missing values were left in place and handled natively by CatBoost. Each point represents one participant and is positioned according to the feature's SHAP value, corresponding to its signed contribution to the model output on the log-odds scale. Points are colored by feature value. Features are ordered by mean absolute SHAP value. SHAP values were computed on the holdout test set ($n = 50{,}041$) using the model refitted on the full development set.}
    \label{fig:mean_shap_raw}
\end{figure}

We computed the SHAP summary plot in the main text on randomly imputed data, consistent with the missing-data policy used throughout model development. Because random imputation replaces each missing value with a draw independent of the participant receiving it, it also removes the predictive information carried by patterns of measurement availability. The resulting feature-importance ranking could therefore depend on the imputation policy. As a sensitivity analysis, we refitted the model using the same 38 features without imputation, letting CatBoost handle the missing values natively during tree construction (Fig.~\ref{fig:mean_shap_raw}).

The extremes of the feature-importance ranking held across missing-data policies. The same six features occupied the highest positions under both random imputation and native missing-value handling: age; the standard PRS for Alzheimer's disease; long-standing illness or disability; number of treatments and medications taken; aspartate aminotransferase; and \textit{APOE} $\varepsilon$4. Left hand grip strength and maternal history of Alzheimer's disease or dementia remained among the lowest-ranked features. Most changes occurred in the middle of the ranking and spanned only a few positions. Length of mobile phone use shifted most, falling from 25th under random imputation to 34th under native handling. The main-text interpretation therefore relies primarily on features that remained highly ranked under both missing-data policies.

\subsection{Local SHAP explanations}
\begin{figure}[htb]
    \centering
    \includegraphics{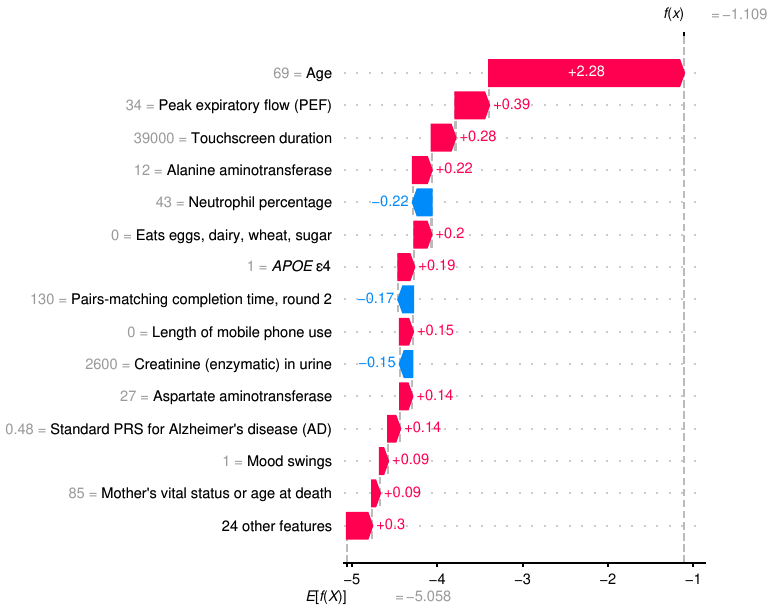}
    \caption{Local SHAP explanation for a representative true positive prediction. The waterfall plot shows how each feature moved this participant's prediction from the cohort baseline to the model output. The bottom axis is the model output. The base value $E[f(X)]=-5.058$ is the mean model output over the test set. The value $f(x)$ is this participant's model output. Each bar is one feature's SHAP contribution, labeled with its signed value; the bars sum from the base value to $f(x)$. The number to the left of each feature name is that participant's value for the feature, rounded to two significant figures to avoid displaying exact individual-level values. Features are ordered by the magnitude of their contribution. The smallest are collapsed into a single ``other features'' bar.}
    \label{fig:shap_tp}
\end{figure}

\begin{figure}[htb]
    \centering
    \includegraphics{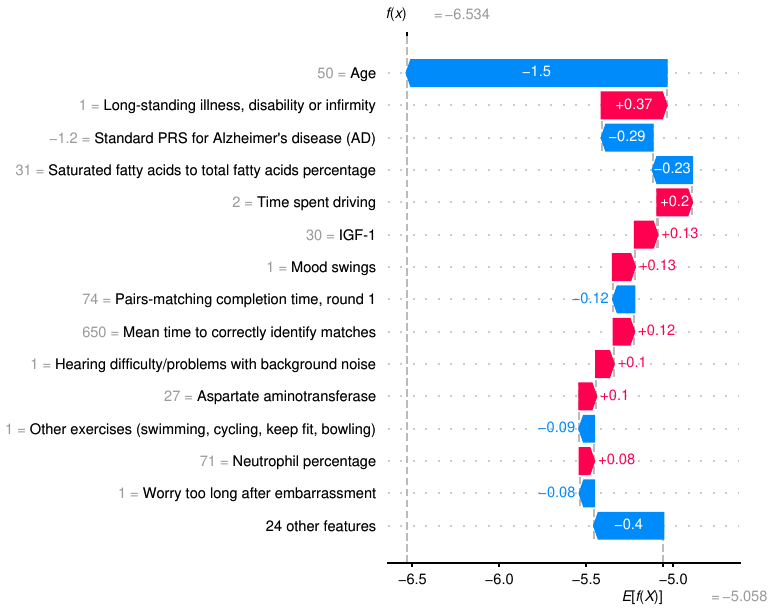}
    \caption{Local SHAP explanation for a representative true negative prediction. Axes and layout are as in Fig.~\ref{fig:shap_tp}; the bars sum from the base value $E[f(X)]$ to the participant's output $f(x)$.}
    \label{fig:shap_tn}
\end{figure}

\begin{figure}[htb]
    \centering
    \includegraphics{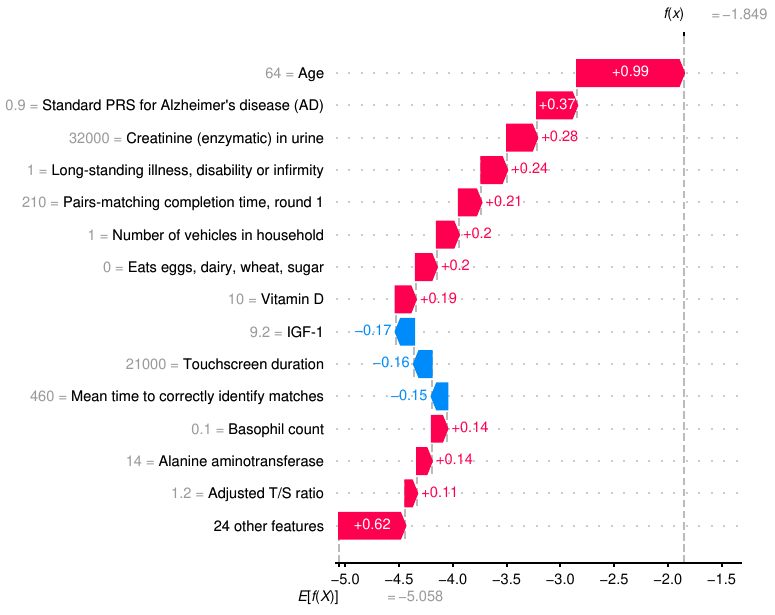}
    \caption{Local SHAP explanation for a representative false positive prediction. Axes and layout are as in Fig.~\ref{fig:shap_tp}; the bars sum from the base value $E[f(X)]$ to the participant's output $f(x)$.}
    \label{fig:shap_fp}
\end{figure}

\begin{figure}[htb]
    \centering
    \includegraphics{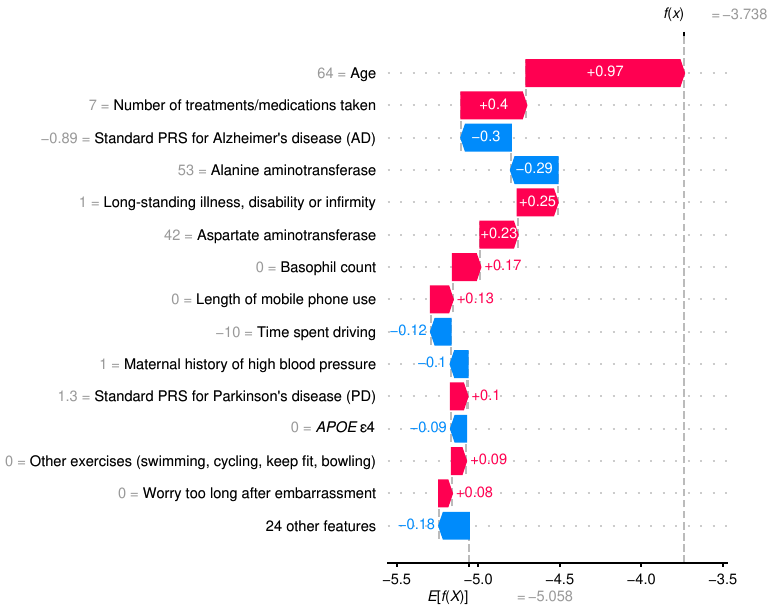}
    \caption{Local SHAP explanation for a representative false negative prediction. Axes and layout are as in Fig.~\ref{fig:shap_tp}; the bars sum from the base value $E[f(X)]$ to the participant's output $f(x)$.}
    \label{fig:shap_fn}
\end{figure}

Representative local SHAP explanations are shown for four individual predictions: a true positive (Fig.~\ref{fig:shap_tp}), true negative (Fig.~\ref{fig:shap_tn}), false positive (Fig.~\ref{fig:shap_fp}), and false negative (Fig.~\ref{fig:shap_fn}). Each plot shows how individual features shifted the participant's predicted risk relative to the model's base value. These case-level explanations complement the global mean absolute SHAP summary reported in the main text.

\clearpage

\subsection{Calibration metrics}

We report calibration at two levels of the hierarchy described by Van Calster et al.~\cite{vancalster2019_calibration}, including both summaries called calibration-in-the-large in that framework. We clipped predicted risks to $[10^{-6},\,1-10^{-6}]$ before applying any logit transformation.

\emph{Mean calibration}, the weakest level, compares the mean predicted risk with the observed incidence. We report the difference $\overline{p}-\bar{y}$ in percentage points, so positive values indicate overestimation. \emph{Weak calibration} comprises the calibration intercept and calibration slope. The calibration intercept is $\alpha$ in the logistic model $\operatorname{logit}(\bar{y})=\alpha+\operatorname{logit}(p)$, with the linear predictor entered as an offset and its coefficient therefore fixed at one. Its target value is zero; following the sign convention of Van Calster et al., positive values indicate underestimation and negative values indicate overestimation. The mean-calibration difference and calibration intercept therefore use opposite sign conventions while describing the same mean-level agreement: at $\alpha=0$, the score equation of the offset model is $\bar{y}-\overline{p}$, so the calibration intercept is zero exactly when the mean predicted risk equals the observed incidence. The calibration slope is the coefficient $\beta$ in the two-parameter logistic regression $\operatorname{logit}(\bar{y})=\alpha'+\beta\operatorname{logit}(p)$, with a target value of one. The intercept $\alpha'$ from this model is a distinct quantity that depends on $\beta$; we do not report it.

The Brier score is the mean squared difference between predicted risk and observed outcome. Because both outcomes were rare, we also report the no-skill Brier score $\bar{y}(1-\bar{y})$ for a model that predicts the observed mean incidence, and the scaled Brier score (Brier skill score), defined as $1-\mathrm{Brier}/\mathrm{Brier}_{\mathrm{null}}$. We obtained 95\% CIs by subject-level bootstrap resampling of the holdout-test predictions (2{,}000 resamples). Table~\ref{tab:calibration-metrics} reports these metrics for both all-incident outcomes.

\begin{table}[htbp]
\centering
\caption{Calibration summaries for ITU-DESC for the all-incident outcomes of all-cause dementia and Alzheimer's disease, evaluated once on the holdout test set ($n = 50{,}041$). Values are point estimates with 95\% confidence intervals, where applicable, obtained by subject-level bootstrap resampling (2{,}000 resamples). Mean calibration is the difference between mean predicted risk and observed incidence in percentage points, positive for overestimation; the calibration intercept is on the log-odds scale, positive for underestimation. Both summarize calibration-in-the-large and equal zero when mean predicted risk matches observed incidence; their opposite sign conventions are described in the text. The mean-calibration difference was calculated from unrounded values. pp, percentage points.}
\label{tab:calibration-metrics}
\begin{tabular*}{\linewidth}{@{\extracolsep{\fill}}lcc}
\toprule
& All-cause dementia & Alzheimer's disease \\
\midrule
Events / $n$ & $1{,}138$ / $50{,}041$ & $513$ / $50{,}041$ \\
\addlinespace
Calibration slope & $1.02$ ($0.97$, $1.08$) & $0.96$ ($0.91$, $1.05$) \\
Calibration intercept & $+0.001$ ($-0.06$, $0.06$) & $+0.002$ ($-0.09$, $0.09$) \\
Mean predicted risk (\%) & $2.27$ & $1.02$ \\
Observed incidence (\%) & $2.27$ & $1.03$ \\
Mean calibration (pp) & $-0.003$ & $-0.002$ \\
\addlinespace
Brier score & $0.021$ & $0.010$ \\
No-skill Brier score & $0.022$ & $0.010$ \\
Scaled Brier score & $0.068$ ($0.060$, $0.077$) & $0.035$ ($0.027$, $0.043$) \\
\bottomrule
\end{tabular*}
\end{table}